\documentclass{article}

\usepackage{microtype}
\usepackage{graphicx}
\usepackage{subfigure}
\usepackage{booktabs} 

\usepackage{hyperref}
\usepackage[dvipsnames]{xcolor}
\usepackage{color, colortbl}
\definecolor{LightCyan}{rgb}{0.88,1,1}

\usepackage{soul}
\sethlcolor{LightCyan}

\usepackage[accepted]{icml2023}

\usepackage{amsmath}
\usepackage{amssymb}
\usepackage{mathtools}
\usepackage{amsthm}

\usepackage[capitalize,noabbrev]{cleveref}

\theoremstyle{plain}
\newtheorem{theorem}{Theorem}[section]

\theoremstyle{definition}
\newtheorem{definition}[theorem]{Definition}

\theoremstyle{remark}
\newtheorem{remark}[theorem]{Remark}

\usepackage[textsize=tiny]{todonotes}

\usepackage{fancyvrb}
\usepackage{fvextra}
\usepackage{enumerate}
\usepackage{algorithm}
\usepackage{algorithmicx}
\usepackage[noend]{algpseudocode}

\usepackage{amsmath,amsfonts,bm}

\def\eqref#1{equation~\ref{#1}}

\def\1{\bm{1}}

\DeclareMathAlphabet{\mathsfit}{\encodingdefault}{\sfdefault}{m}{sl}
\SetMathAlphabet{\mathsfit}{bold}{\encodingdefault}{\sfdefault}{bx}{n}

\newcommand{\R}{\mathbb{R}}

\usepackage{amsmath, amsfonts, amssymb, amsthm, amsbsy, amscd, bm, bbm,mathrsfs}    
\usepackage{graphicx}
\usepackage{cleveref}
\usepackage[shortlabels]{enumitem}
\usepackage{stfloats} 
\usepackage{xurl}

\setlist{leftmargin=5mm}
\usepackage{titlesec}
\usepackage{wrapfig}
\usepackage{multirow}
\graphicspath{{figs/}}

\renewcommand{\a}{\bm{a}}
\newcommand{\s}{\bm{s}}

\newcommand{\g}{\bm{g}}

\renewcommand{\b}{\mathbf{b}}
\newcommand{\I}{\mathbf{I}}
\newcommand{\C}{\mathbf{C}}

\newcommand{\W}{\mathbf{W}}
\renewcommand{\s}{\bm{s}}

\newcommand{\G}{\mathbf{G}}

\usepackage{pifont}
\newcommand{\cmark}{\ding{51}}%
\newcommand{\xmark}{\ding{55}}%

\usepackage{tikz}
\newcommand*\circled[1]{\tikz[baseline=(char.base)]{\node[shape=circle,draw,inner sep=2pt] (char) {#1};}}

\icmltitlerunning{Differentially Private Optimization on Large Model at Small Cost}

\begin{document}

\twocolumn[
\icmltitle{Differentially Private Optimization on Large Model at Small Cost}



\icmlsetsymbol{equal}{*}

\begin{icmlauthorlist}
\icmlauthor{Zhiqi Bu}{yyy}
\icmlauthor{Yu-Xiang Wang}{comp}
\icmlauthor{Sheng Zha}{yyy}
\icmlauthor{George Karypis}{yyy}
\end{icmlauthorlist}

\icmlaffiliation{yyy}{Amazon Web Services}
\icmlaffiliation{comp}{University of California, Santa Barbara}

\icmlcorrespondingauthor{Zhiqi Bu}{zhiqibu@amazon.com}

\icmlkeywords{deep learning, algorithm, differential privacy, system design}

\vskip 0.3in
]



\printAffiliationsAndNotice{} 

\begin{abstract}
Differentially private (DP) optimization is the  standard paradigm to learn large neural networks that are accurate and privacy-preserving. The computational cost for DP deep learning, however, is notoriously heavy due to the per-sample gradient clipping. Existing DP implementations are $2\sim 1000\times$ more costly in time and space complexity than the standard (non-private) training. In this work, we develop a novel Book-Keeping (BK) technique that implements existing DP optimizers (thus achieving the same accuracy), with a substantial improvement on the computational cost. Specifically, BK enables DP training on large models and high dimensional data to be roughly as fast and memory-saving as the standard training, whereas previous DP algorithms can be inefficient or incapable of training due to memory error. The computational advantage of BK is supported by the complexity analysis as well as extensive experiments on vision and language tasks. Our implementation achieves state-of-the-art (SOTA) accuracy with very small extra cost: on GPT2 and at almost the same memory cost ($<1\%$ overhead), BK has 1.03$\times$ the time complexity of the standard training (0.83$\times$ training speed in practice), and 0.61$\times$ the time complexity of the most efficient DP implementation (1.36$\times$ training speed in practice). We open-source the codebase for the BK algorithm at \textbf{FastDP} library \url{https://github.com/awslabs/fast-differential-privacy}.
\end{abstract}

\section{Introduction}
Deep learning with differential privacy (DP; \cite{dwork2006calibrating}) has shown strong performance while guaranteeing rigorous protection against privacy risks, especially on large models that tend to memorize and leak the training data \cite{carlini2021extracting, haim2022reconstructing,shokri2017membership}. For example, recent advances have shed light on the success of DP GPT2 \cite{li2021large,bu2022automatic,yu2021differentially}, which achieves $64.6$ BLEU score\footnote{BLEU (\href{https://cloud.google.com/translate/automl/docs/evaluate\#interpretation}{BiLingual Evaluation Understudy}) is a metric  (0-100) for automatically evaluating translated text. BLEU $>60$ is considered as "very high quality, adequate, and fluent translations, often better than human".}
at strong privacy guarantee ($\epsilon=3$), on the text generation task using E2E restaurant review dataset. This is only marginally below the standard non-private GPT2 (BLEU score 66.8). Similarly, on computer vision tasks ($\epsilon=2$), DP vision transformers and ResNets have obtained $97.1\%/86.2\%$ accuracy on CIFAR10/100 by \cite{bu2022scalable} and over $81\%$ accuracy on ImageNet by \cite{de2022unlocking,mehta2022large}. 

However, DP training of large neural networks is well-known to be computationally burdensome in comparison to the standard training, in terms of both the training time and the memory cost. For instance, training a small recurrent neural network (0.598M parameters) experiences a $1000\times$ slowdown using DP optimizers in Tensorflow-Privacy (TF-Privacy) library in \cite{bu2021fast}, and training a small convolutional neural network (CNN, 0.605M parameters) on CIFAR10 has a $24\times$ slowdown with Tensorflow 2 and the XLA compiler \cite{subramani2021enabling}. Even with SOTA efficient implementations, large models such as RoBERTa \cite{liu2019roberta}, GPT2 \cite{radford2019language}, ResNet \cite{he2016deep}, VGG \cite{simonyan2014very}, ViT \cite{dosovitskiy2020image} and its variants, experience about $2\sim 3\times$ slowdown in Pytorch \cite{li2021large,bu2022scalable} and $2\sim  9\times$ slowdown in JAX \cite{kurakin2022toward,de2022unlocking}, with possibly $4\sim  20\times$ memory overhead \cite{bu2022scalable,li2021large,subramani2021enabling} if not running out of memory.

\begin{table*}[!htb]
\setlength\tabcolsep{1.5pt}
    \centering
    \resizebox{0.85\linewidth}{!}{
    \begin{tabular}{c|c|c|c|c|c|c}
    \multirow{2}{*}{Dataset}&\multirow{2}{*}{SOTA setting}&\multirow{2}{*}{Model}&\multirow{2}{*}{\shortstack{Time\\/Epoch}}&\multicolumn{3}{|c}{Relative Speed (same memory contraint)}\\
    &&&&to GhostClip&to Opacus&to non-DP\\\hline
     QQP& \cite{li2021large}&RoBERTa-large (355M)&70'04''&1.36$\times$&1.96$\times$&0.77$\times${\color{orange}$(0.89\times)$}
\\
E2E& \cite{li2021large}&GPT2-large (774M)&10'01''&1.36$\times$&4.41$\times$&0.83$\times${\color{orange}$(0.97\times)$}
     \\
CIFAR& \cite{bu2022scalable}&BEiT-large (304M)&6'35''&1.33$\times$&38.3$\times$&0.76$\times${\color{orange}$(0.92\times)$} 
\end{tabular}
}
\vspace{-0.2cm}
    \caption{Efficiency of BK algorithm on DP tasks using one A100 GPU (same accuracy). Note the speed is measured in wall-time (hardware speed) and in \textcolor{orange}{complexity (theoretical speed)}. More models and tasks can be found in \Cref{tab:GPT max throughput}.}
    \label{tab:SOTA throughput}
\end{table*}

The efficiency bottleneck in DP deep learning lies in the per-sample gradient clipping, which restricts the magnitude of each per-sample gradient in the mini-batch. Applying the clipping jointly with the Gaussian noise addition, one can privately release the gradient to arbitrary optimizers like SGD and Adam, and thus guarantee the privacy of the training as described in \Cref{sec:DP prelim}:
\begin{align}
&\text{private gradient: } \hat\G:=\sum\nolimits_i \g_i\cdot C(\|\g_i\|_2)+\sigma_\text{DP} \cdot\mathcal{N}(0,\I),
\nonumber\\
&\text{private optimizer (e.g. SGD):}\quad\W_{t+1}=\W_t-\eta\hat\G.
\label{eq:DP optimizers}
\end{align}
Here $\W$ is the model parameters, $\mathcal{L}_i$ is the per-sample loss, $\g_i=\frac{\partial \mathcal{L}_i}{\partial \W}$ is the per-sample gradient, $\eta$ is the learning rate, $\sigma_\text{DP}$ is the noise magnitude that defines the privacy loss, and $C(\|\g_i\|)$ or simply $C_i$ is the per-sample clipping factor. For example, in \cite{abadi2016deep}, $C_i=\min\{R/\|\g_i\|,1\}$ for some clipping threshold $R$; in \cite{bu2021convergence}, $C_i=\mathbb{I}(\|\g_i\|\leq R)$; in \cite{bu2022automatic}, $C_i=1/(\|\g_i\|+0.01)$ or $1/\|\g_i\|$ as the gradient normalization.

At high level, the DP training is a system effort consisting of multiple parts:
\begin{enumerate}[I.]
    \item optimizer: DP-SGD, DP-Adam, DP-LAMB;
    \item parameter efficiency: last layer (linear probing), LoRA, Adapter, BiTFiT;
    \item implementation: Opacus, GhostClip, Book-Keeping;
    \item platform: Pytorch, JAX, TensorFlow.
\end{enumerate}

Previous works have tackled the efficiency bottleneck with various approaches. One approach (part II) focuses on the parameter efficiency by partially training a neural network, in contrast to fully fine-tuning all model parameters, e.g. only the last output layer \cite{tramer2020differentially}, the adapter layers \cite{houlsby2019parameter,mahabadi2021compacter}, or the Low-Rank Adaptation (LoRA) \cite{hu2021lora,yu2021differentially}. For example, \cite{mehta2022large} accelerate the DP training on ImageNet \cite{deng2009imagenet} up to $30\times$ by only training the last layer of ResNet152. Noticeably, parameter efficient fine-tuning does not improve on the efficiency in terms of complexity per parameter, rather than reducing the number of parameters. Furthermore, this approach oftentimes leads to some accuracy degradation compared to DP full fine-tuning \cite{bu2020deep,mehta2022large,li2021large,yu2021differentially}.

An orthogonal approach, including this work, focuses on the computation efficiency (part III), i.e. reducing the time and space complexity through efficient implementations, without modifying the DP optimizers (part I) and thus not affecting their performance. We will elaborate on existing methods in \Cref{sec:previous arts}. Additionally, these methods can be compiled on different platforms (part IV) such as Tensorflow 2(XLA), JAX and Pytorch \cite{li2021large,subramani2021enabling,de2022unlocking,kurakin2022toward}, where remarkable speed difference has been observed in some cases, even with the same implementation. For example, \cite{subramani2021enabling} implemented DP-SGD using JAX and claimed its efficiency advantage over the same algorithm using Tensorflow or Pytorch.

\subsection{Contributions}

\begin{figure}[!htb]
    \centering
    \includegraphics[width=\linewidth]{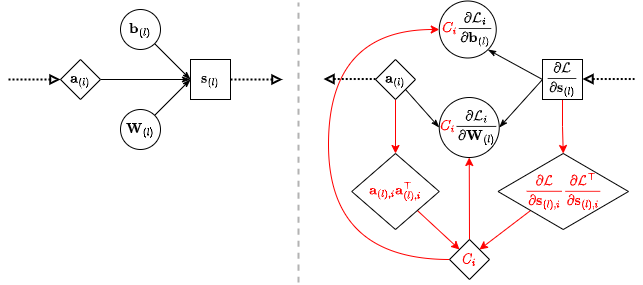}
    \vspace{-0.4cm}
    \caption{Forward pass and back-propagation of the $l$-th linear layer (standard training is in black; DP training by our book-keeping algorithm is added in \textcolor{red}{red}). Here $\a_{(l)}$ is the activation tensor, $\s_{(l)}$ is the layer output, $\W_{(l)},\b_{(l)}$ are weight and bias, $\mathcal{L}_i, \mathcal{L}$ are the per-sample loss and the summed loss. The dotted arrow is the inter-layer operation such as pooling or normalization.}
    \label{fig:MLP forward backward}
\end{figure}

\begin{enumerate}
\item \textbf{[Algorithm]} We propose the book-keeping (BK) algorithm that makes existing DP optimizers fast and memory efficient, especially comparable to non-private optimizers. We demonstrate BK via the computation graph in \Cref{fig:MLP forward backward}. The highlight is that BK \textit{only uses one back-propagation} and \textit{never instantiates per-sample gradients} $\{\frac{\partial \mathcal{L}_i}{\partial \W}\}_{i=1}^B$.

\item \textbf{[Analysis]} We analyze the complexity to show that \textit{BK has almost the same time and space complexity as non-DP training}, especially when the feature dimension is small (see \Cref{tab:detailed BK and mixed and everything}).

\item \textbf{[Extension]} We strengthen BK using a layerwise decision to mix with Opacus (see \Cref{sec:hybrid}), which proves to be efficient when the feature dimension is large (and difficult for GhostClip). We also extend BK to the parameter efficient fine-tuning such as DP LoRA and Adapter.

\item \textbf{[Codebase]} We develop a Pytorch \citep{paszke2019pytorch} codebase for our BK algorithm, leveraging the auto-differentiation technique on the computation graph and a new trick in \Cref{app:auto-differentiation}. We highlight that our codebase can automatically switch the standard training of \textit{any model} to its DP version, by adding a single piece of codes.

\item \textbf{[Experiments]} We demonstrate the amazing efficiency of BK on training large models, saving the memory up to $10\times$ and boosting the speed by $30\%\sim 5\times$ than previous DP implementations.
\end{enumerate}

\begin{table*}[!b]
\setlength\tabcolsep{2pt}
    \centering
    \resizebox{0.93\linewidth}{!}{    
    \begin{tabular}{c|c|c|c|c|c|c}
         &Non-DP&TF-privacy& Opacus&FastGradClip&GhostClip&BK (ours) \\
         Instantiating per-sample grad&\xmark&\cmark&\cmark&\cmark&\xmark&\xmark
         \\
         Storing every layer's grad&\xmark&\xmark&\textcolor{red}{\cmark}&\xmark&\xmark&\xmark
         \\
         Instantiating non-DP grad&\cmark&\cmark&\cmark&\xmark&\textcolor{red}{\cmark}&\xmark
         \\
         Number of back-propagation&1&\textcolor{red}{$B$}&1&\textcolor{red}{2}&\textcolor{red}{2}&1
         \\
         Time Complexity of Clipping &$6BTpd$&$6BTpd$&$8BTpd$&$8BTpd$&$10BTpd+O(BT^2)$&$\approx 6BTpd$
         \\
         Memory Overhead to non-DP &$0$&$0$&\textcolor{red}{$Bpd$}&\textcolor{red}{$Bpd$}&$2BT^2$&$\min\{2BT^2,Bpd\}$
      \\
Scalable to large model&\cmark&\xmark&\xmark&\xmark&\cmark&\cmark
     \\
Scalable to high-dim input&\cmark&\xmark&\cmark&\cmark&\textcolor{red}{\xmark}&\cmark
\end{tabular}
}
\vspace{-0.2cm}
    \caption{Summary of different DP implementations on a linear/convolution layer $\R^{B\times T_{(l)}\times d_{(l)}}\to\R^{B\times T_{(l)}\times p_{(l)}}$. The main bottleneck is marked in \textcolor{red}{red}.}
    \label{tab:compare different implementations}
\end{table*}

\newpage
\vspace{-0.3cm}
\subsection{Related works}
\label{sec:previous arts}
Previous arts have developed different implementations of the same DP optimizer in \Cref{eq:DP optimizers}. 
Among these implementations, the tradeoff between the time and space complexity has been constantly maneuvered. 
TF-Privacy \cite{tfprivacy} back-propagates a vectorized loss $[\mathcal{L}_1,\cdots,\mathcal{L}_B]$ to compute the per-sample gradients, each from one back-propagation, which is memory-efficient but slow. Opacus \cite{opacus} and \cite{rochette2019efficient} accelerate the training significantly using the outer product trick in \cite{goodfellow2014explaining}, though incurring heavy memory burden so as to store the per-sample gradients. This memory burden is partially alleviated in FastGradClip \cite{lee2020scaling} by sharing the space complexity in two rounds of back-propagation, hence almost doubling the time complexity. In ghost clipping \cite{goodfellow2015efficient,li2021large,bu2022scalable}, the per-sample gradients can be clipped without being instantiated, thus both time and space complexity can be further improved if the feature dimension is small. We refer interested readers to \Cref{fig:all algorithms} and \Cref{app:line by line implementations} for algorithmic details of these implementations.

\begin{figure*}[!htb]
    \centering
    \includegraphics[width=0.25\linewidth]{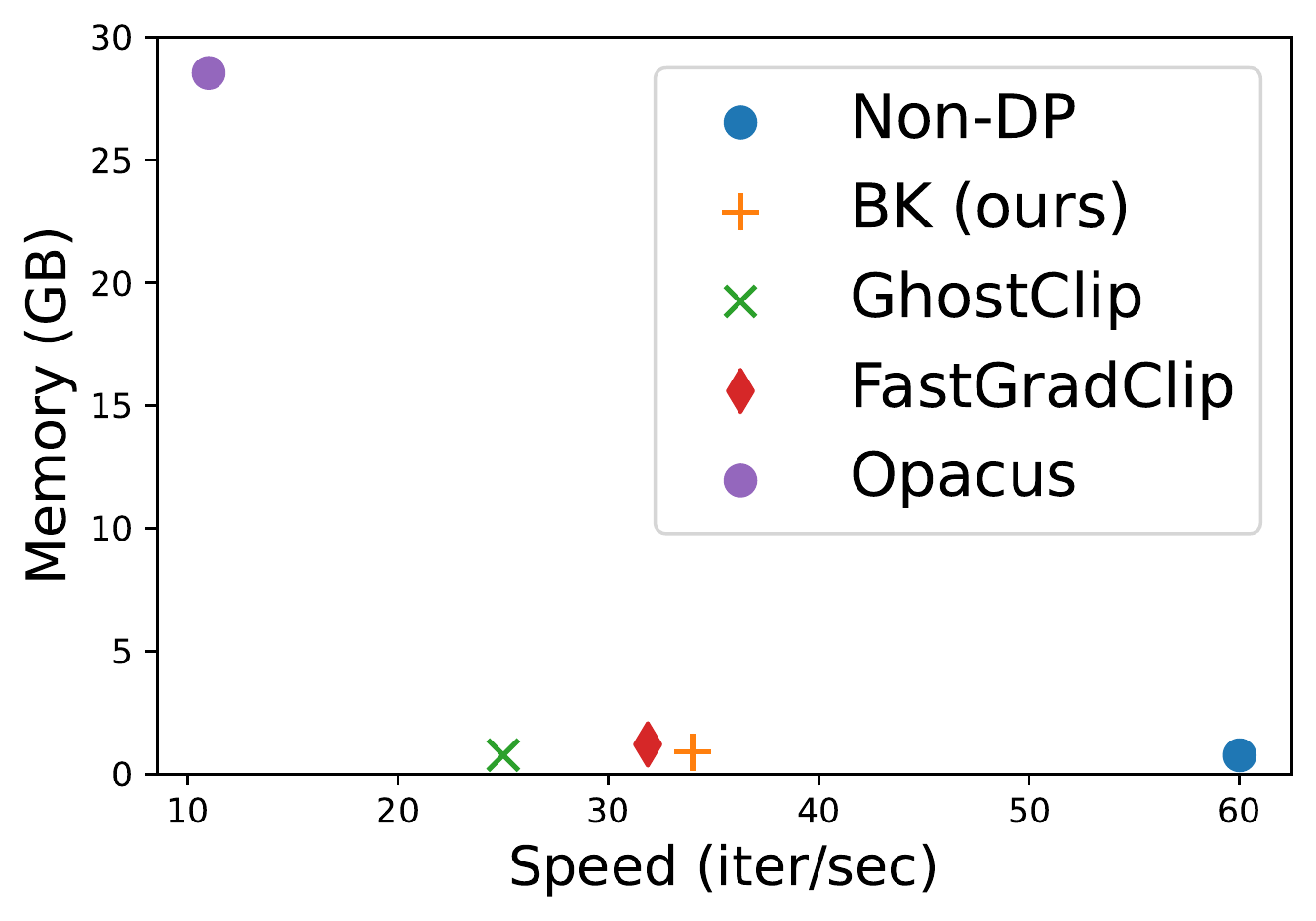}
    \includegraphics[width=0.24\linewidth]{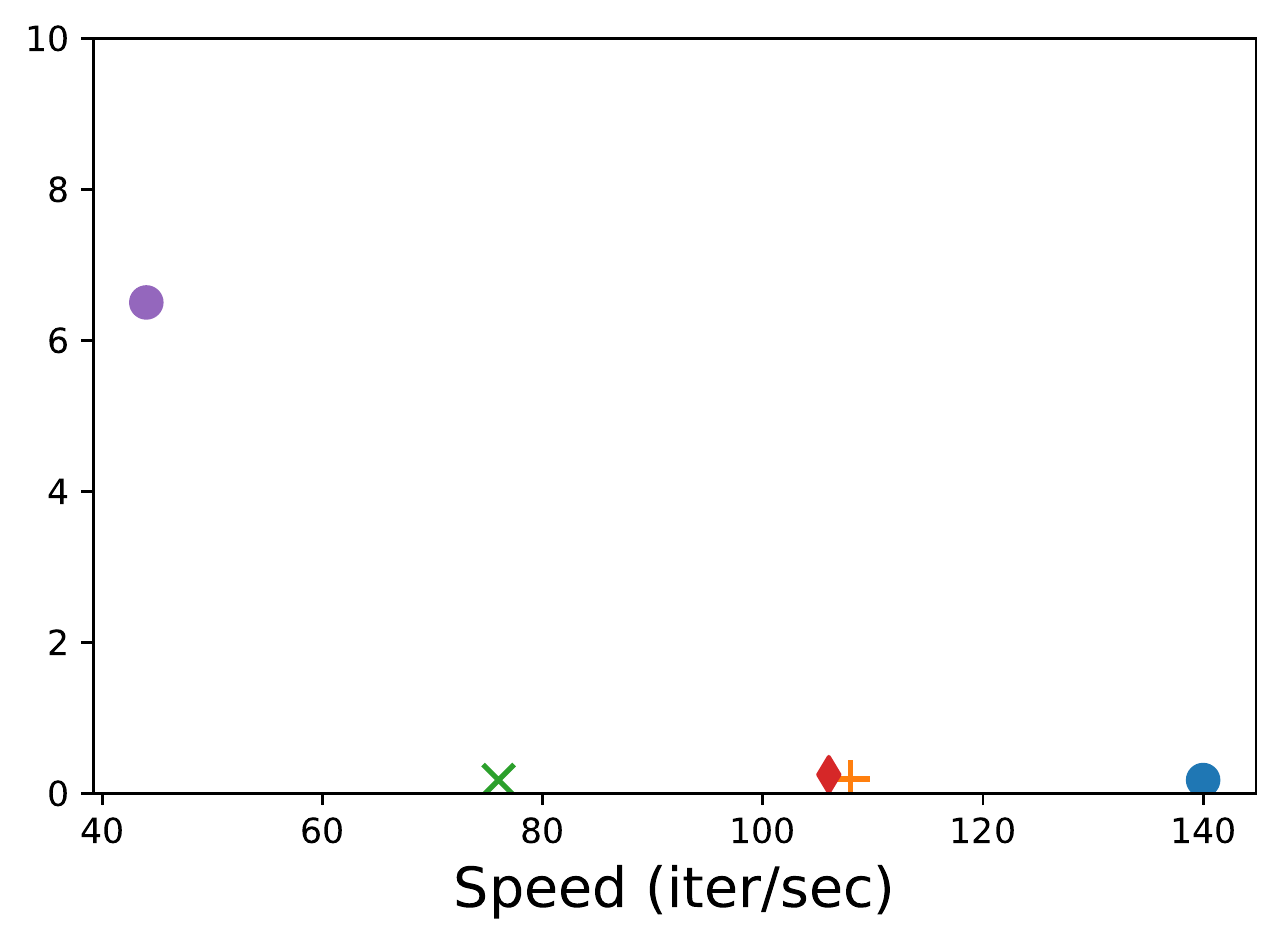}
    \includegraphics[width=0.24\linewidth]{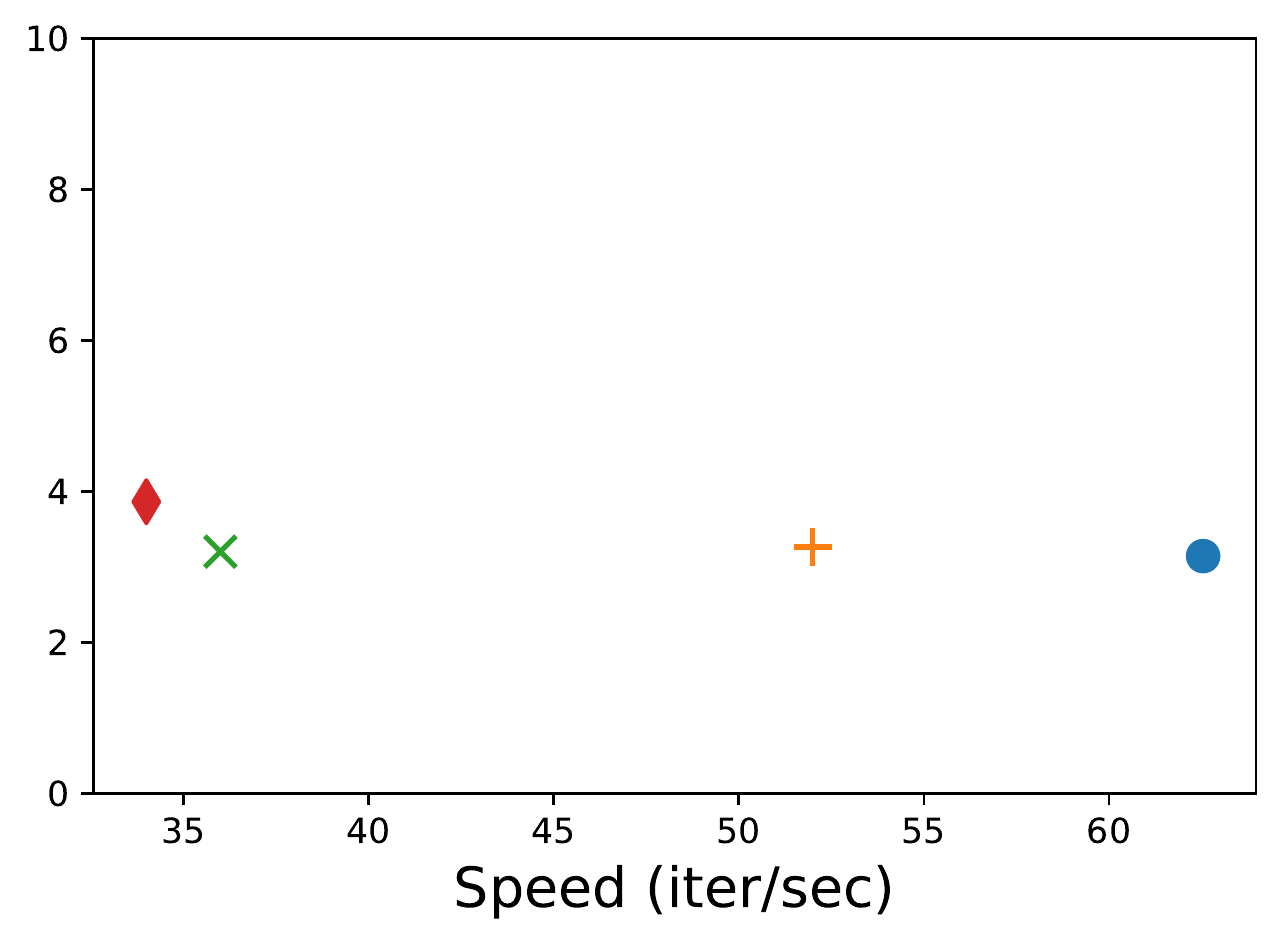}
    \includegraphics[width=0.24\linewidth]{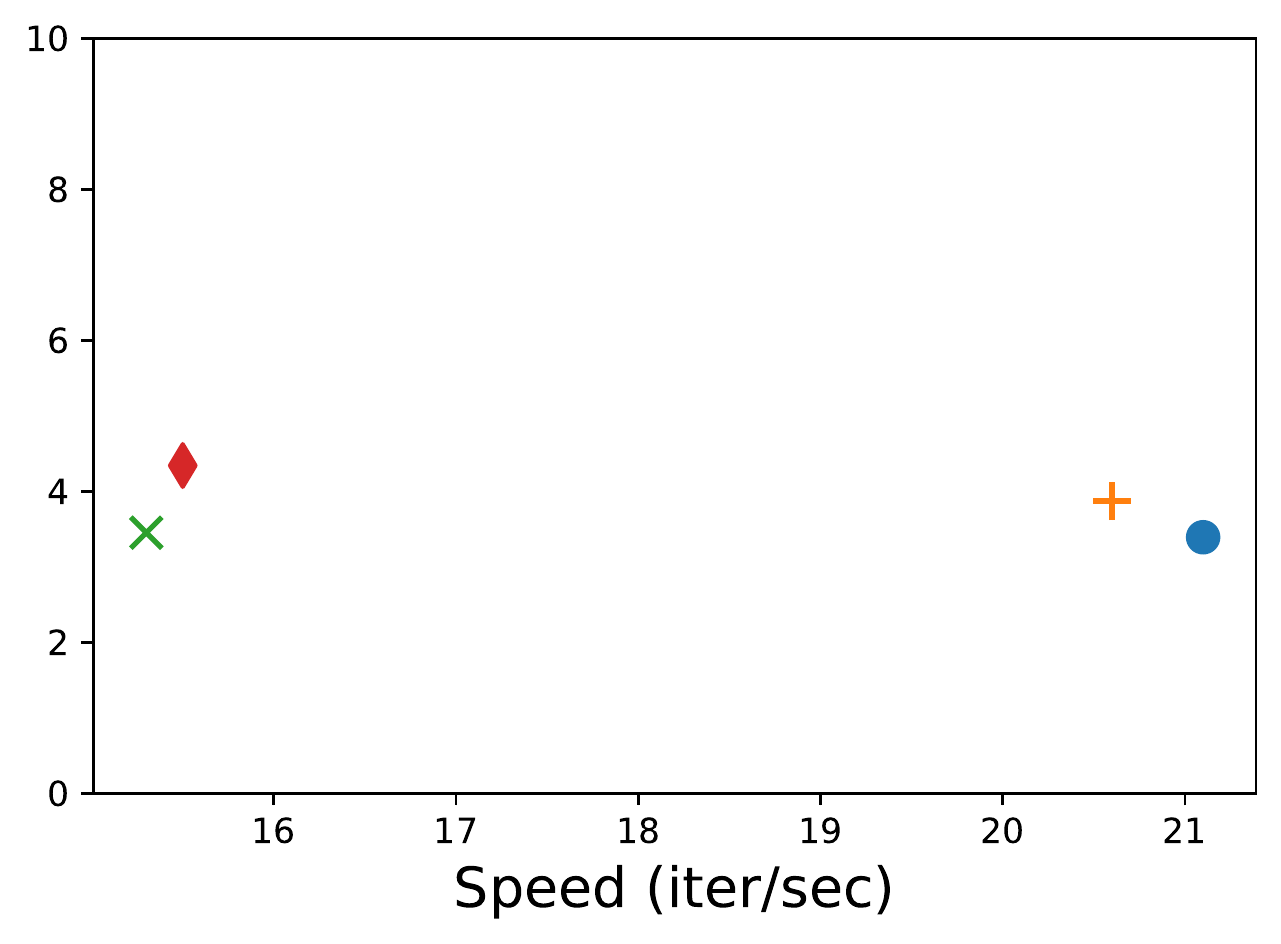}
\vspace{-0.3cm}
    \caption{Speed and memory on MLP and CIFAR100 (images are flattened into vectors). Left to right: deep network (50 layers, width 1000, 50M parameters, batch size 128), shallow network (10 layers, width 1000, 10M parameters, batch size 128), and wide network (10 layers, width 5000, 250M parameters, batch size 128 or 1024; Opacus is OOM). See more ablation study in \Cref{app:additional}.}
    \label{fig:comparison implementations MLP}
\end{figure*}

We now compare BK to different implementations in \Cref{tab:compare different implementations} and \Cref{fig:comparison implementations MLP}. In what follows, $B$ is the batch size\footnote{In this work, we report the physical batch size, which affects the efficiency but not the accuracy; the accuracy is only affected by the logical batch size, which can be implemented through the gradient accumulation of physical batch size.}, $T_{(l)}$ is the feature dimension\footnote{For non-sequential data, $T=1$; for texts, $T$ is the sequence length, which is layer-independent; for images (or videos), $T_{(l)}$ is the height$\times$width($\times$time) of hidden feature representation, which is layer-dependent.}, $d_{(l)},p_{(l)}$ are the input or output dimension of a layer.

\vspace{-0.1cm}
\subsection{Preliminaries}
\label{sec:DP prelim}
We work with the $(\epsilon,\delta)$-DP by
\cite{dwork2006calibrating}, defined in \Cref{app:prelim}, which makes it difficult for any privacy attacker to distinguish or detect an arbitrary training sample, even with full access to the model. In deep learning, DP is achieved by training on the private gradient in \Cref{eq:DP optimizers} with any optimizer such as SGD, Adam, FedAvg, etc. Essentially, the private gradient is the addition of Gaussian noise to the sum of clipped per-sample gradients, which guarantees the DP protection through the privacy accounting theorems \cite{abadi2016deep,mironov2017renyi,dong2019gaussian,zhu2021optimal,gopi2021numerical,koskela2020computing}.

\section{Book-keeping: Efficient DP training in low dimension}

The main computational bottleneck of DP training comes from the per-sample gradient clipping, or from the computation of per-sample gradient norms, to be exact. One widely used approach in Opacus, TF-privacy, and FastGradClip, is to instantiate the per-sample gradients and then deriving their norms. Straight-forward implementation of this approach on a mini-batch of per-sample losses requires $B$ rounds of back-propagation (unacceptable slowdown) or $B\times$ gradient storage (unacceptable memory burden; see Opacus in \Cref{fig:comparison implementations MLP}). Consequently, these implementations are not suitable for large model training. For instance, \cite{li2021large} shows that, when training GPT2-large (774M parameters), Opacus \cite{opacus} and JAX \cite{subramani2021enabling} cannot fit even one single sample into a 24GB GPU.

An alternative approach, termed as the ghost clipping (GhostClip), directly computes the per-sample gradient norms without computing the gradients themselves. This is made possible, unfortunately, through two rounds of back-propagation. During the first back-propagation, one uses the regular loss $\sum_i \mathcal{L}_i$ and extracts the activation tensor and the output gradient $(\a,\frac{\partial \mathcal{L}}{\partial\s})$. One can use an algebraic trick in \Cref{eq:ghost norm} to compute the per-sample gradient norms $\{\|\frac{\partial \mathcal{L}_i}{\partial \W}\|\}_i$ and the clipping factors $\{C_i\}_i$ in \Cref{eq:DP optimizers}. During the second back-propagation, one uses the reweighted loss $\sum_i C_i\mathcal{L}_i$ whose gradient is directly the weighted gradient $\sum_i C_i\g_i$, which constitutes the private gradient we need. Note that this double back-propagation roughly doubles the training time (or to be more precise, $10/6\approx 1.667\times$  when $T$ is small; but this approach loses its advantage when $T$ is large), as shown in \Cref{tab:compare different implementations}).

To make the DP training as efficient as the standard training, we propose the book-keeping technique (BK) that $\langle1\rangle$ only requires a single round of back-propagation, like Opacus and the standard training; $\langle2\rangle$ does not instantiate the per-sample gradients, like GhostClip.

\subsection{Book-keeping algorithms}
\label{sec:tricks}

\begin{figure*}[!htb]
    \centering
    \includegraphics[width=\linewidth]{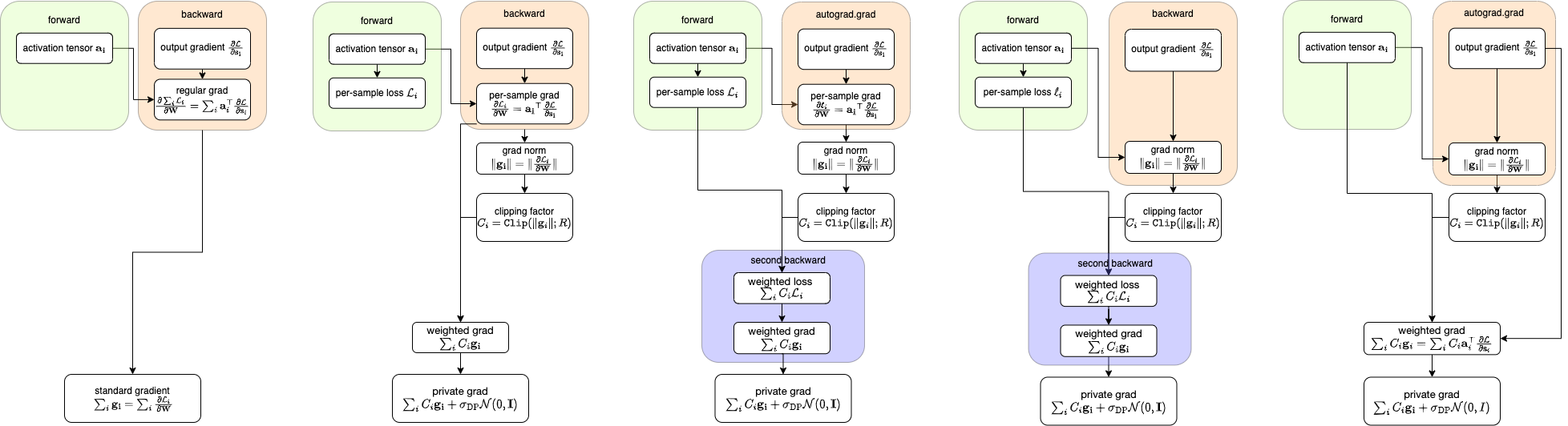}
    \vspace{-0.3cm}
    \caption{Standard (non-DP), Opacus, FastGradClip, GhostClip, and BK implementations, from left to right. Notice that BK directly computes clipped gradient like Opacus, computes the ghost norm like GhostClip, and uses auto-differentiation like FastGradClip.}
    \label{fig:all algorithms}
\end{figure*}

BK algorithms in their base forms are built on GhostClip and especially the \textit{ghost norm} trick, so as to avoid instantiating the memory costly per-sample gradients: as can be seen in \Cref{alg:dpsgd-ghostBK} and \Cref{fig:all algorithms}, $\frac{\partial \mathcal{L}_i}{\partial \W}=\a_{i}^\top\frac{\partial \mathcal{L}}{\partial \s_{i}}$ is not computed throughout the training. In comparison to GhostClip, our significant improvement is solely on the speed (see \Cref{tab:compare different implementations}) through two novel tricks: the \textit{book-keeping} and the \textit{ghost differentiation}. The entire BK algorithm is built on the understanding of computation graph in \Cref{app:prelim}. Note that these tricks also offer improved efficiency for existing implementations, to be presented in \Cref{sec:apply to others}. We now elaborate on these tricks.

\resizebox{\linewidth}{!}{
$\text{BK (base)}=\underbrace{\text{ghost norm}}_{\text{from GhostClip}}+\underbrace{\text{book-keeping}}_{\text{ours}}+\underbrace{\text{ghost differentiation}}_{\text{ours}}$
}

\vspace{-0.2cm}
\resizebox{\linewidth}{!}{
\begin{minipage}{\linewidth}
\begin{algorithm}[H]
\caption{Differentially private deep learning with BK}
\label{alg:dpsgd-ghostBK}

\textbf{ Parameters:} $l$-th layer weights $\W_{(l)}$, number of layers $L$, noise level $\sigma$, clipping threshold $R$.

\begin{algorithmic}[1]
\For{layer $l\in 1,2,\cdots,L$}
\State Get activation tensor $\{\a_{(l),i}\}$ by forward hook
\EndFor
\For{layer $l\in L,L-1,\cdots,1$}
\State Get output gradient $\{\frac{\partial \mathcal{L}}{\partial \s_{(l),i}}\}$ by backward hook
\State {Compute per-example gradient norm $\|\frac{\partial \mathcal{L}_i}{\partial\W_{(l)}}\|_F^2$ by ghost norm trick in \Cref{eq:ghost norm}}
\EndFor

\State Aggregate gradient norm across layers: $\|\frac{\partial \mathcal{L}_i}{\partial\W}\|_F^2=\sum_l\|\frac{\partial \mathcal{L}_i}{\partial\W_{(l)}}\|_F^2$
\State Compute clipping factor: $C_{i}=C(\|\frac{\partial \mathcal{L}_i}{\partial\W}\|_F;R)$
\For{layer $l\in L,L-1,\cdots,1$}
\State Compute sum of clipped gradients $\G_l=\a_{(l)}^\top\text{diag}(C_1,C_2,\cdots)\frac{\partial \mathcal{L}}{\partial \s_{(l)}}$
\State Delete $\{\a_{(l),i}\},\{\frac{\partial \mathcal{L}}{\partial \s_{(l),i}}\}$
\EndFor
\State Add Gaussian noise $\hat\G=\G+\sigma R\cdot \mathcal{N}(0, \I)$
\State Apply SGD/Adam/LAMB with the private gradient $\hat\G$
\end{algorithmic}
\end{algorithm}
\end{minipage}
}

\paragraph{[Ghost norm trick {\color{orange} $\longleftarrow$memory improvement}]}
The ghost norm trick \cite{goodfellow2015efficient} computes the gradient norm without the gradient: while the gradient is instantiated by the multiplication in \Cref{eq:ghost norm}, the gradient norm can be computed without $\a_i$ meeting $\frac{\partial\mathcal{L}}{\partial \s_i}$. This trick is applicable to generalized linear layers including the linear, the embedding \cite{li2021large}, and the convolution layers \cite{bu2022scalable}. We emphasize that these generalized linear layers represent 99.9\% of the trainable parameters in modern neural networks.

We demonstrate this trick using a simple linear layer $\s_{i}=\a_{i}\W $, where $\W\in\R^{d\times p}$ is the weight matrix, $\a\in\R^{B\times T\times d}$ is the mini-batch input of this layer (a.k.a. the activation tensor) and $\s\in\R^{B\times T \times p}$ is the output. Given that the output gradient $\frac{\partial \mathcal{L}}{\partial \s}$ is readily available in the back-propagation, for DP and standard training, one can directly derive the per-sample gradient norm
\begin{align}
\begin{split}
&\left\|\frac{\partial \mathcal{L}_i}{\partial \W}\right\|_\text{Frobenius}^2=
\text{vec}\Big(\frac{\partial \mathcal{L}}{\partial \s_i} \frac{\partial \mathcal{L}}{\partial \s_i}^\top\Big)\cdot\text{vec}\left(\a_i \a_i^\top\right)
\end{split}
\label{eq:ghost norm}
\end{align}
without actually computing $\frac{\partial \mathcal{L}_i}{\partial \W}=\a_i^\top \frac{\partial \mathcal{L}}{\partial \s_i}$. Here `vec' means flattening the $T\times T$ matrix to a vector. 
This trick is particularly efficient when $T$ is small, reducing the space complexity from $O(Bpd)$ to $O(BT^2)$ by \Cref{tab:block complexity}.

\begin{figure}[!htb]
    \centering
    \includegraphics[width=0.55\linewidth]{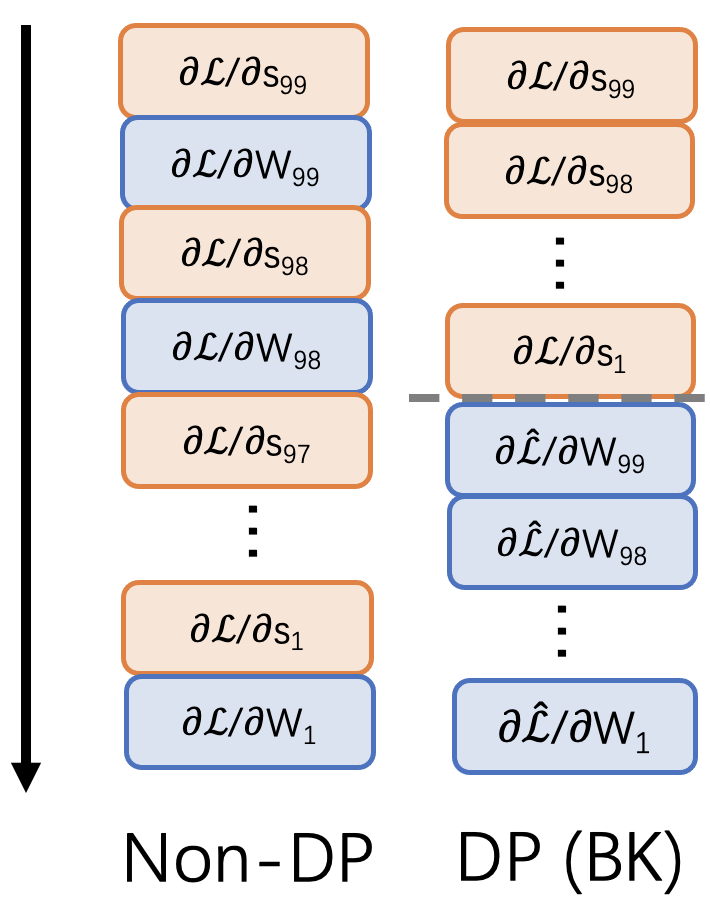}
\vspace{-0.4cm}
    \caption{Backward propagation of BK algorithm. Here $\mathcal{L}:=\sum_i \mathcal{L}_i, \hat{\mathcal{L}}:=\sum_i C_i\mathcal{L}_i$.}
    \label{fig:my_label}
\end{figure}

\paragraph{[Book-keeping trick {\color{blue} $\longleftarrow$speed improvement}]} This trick improves the time complexity by removing the second back-propagation from GhostClip. Our idea is to book-keep and re-use the output gradient $\frac{\partial \mathcal{L}}{\partial \s_{(l)}}$, which is deleted after the first back-propagation of GhostClip and must be re-computed during the second back-propagation. The difference between GhostClip and BK is clearly illustrated via a line-by-line comparison in \Cref{alg:BK v.s. GhostClip}. In fact, denoting the total number of model parameters as $M=\sum_l p_{(l)}d_{(l)}$, our trick reduces the time complexity from $10BTM+O(BT^2)$ by GhostClip to $8BTM+O(BT^2)$ according to \Cref{tab:block complexity}. In contrast to Opacus, which book-keeps the per-sample gradients $\g^{(l)}_i$ using $O(BM)=O(B\sum_l p_{(l)}d_{(l)})$ memory, we instead book-keep the output gradient with substantially cheaper $O(BT\sum_l p_{(l)})$ memory when the feature dimension $T$ is small.

\paragraph{[Ghost differentiation trick {\color{blue} $\longleftarrow$speed improvement}]} This trick improves the time complexity on the first back-propagation in GhostClip, further reducing from $8BTM+O(BT^2)$ to $6BTM+O(BT^2)$ in \Cref{tab:compare different implementations}. Our idea is to only compute the output gradient $\frac{\partial\mathcal{L}}{\partial\s_{(l)}}$ but not the (non-private) parameter gradient $\frac{\partial\mathcal{L}}{\partial\W}$. That is, we break the $4BTM$ time complexity of the full back-propagation into two sub-processes, each of $2BTM$ complexity, and remove the unnecessary one. 

To be more specific, during the back-propagation of Opacus and GhostClip, the output gradient $\frac{\partial \mathcal{L}}{\partial \s}$ and then the parameter gradient $\frac{\partial \mathcal{L}}{\partial \W}=\a^\top \frac{\partial \mathcal{L}}{\partial \s}$ are computed. However, we can stop after we obtain $\frac{\partial\mathcal{L}}{\partial \s}$: we only need the output gradient to compute the clipped parameter gradient $\frac{\partial \sum_i C_i\mathcal{L}_i}{\partial \W}$ in Line 9 of \Cref{alg:dpsgd-ghostBK}. Therefore, the ghost differentiation trick sets all parameters to not require gradients. See the technical details in \Cref{app:auto-differentiation}, including the \textit{origin parameter trick} that propagates on a computation graph even when no parameters require gradients.

\subsection{Complexity of DP implementations: a modular analysis}
In this section, we analyze the complexity of DP implementations from their opearation modules. We summarize the time and space complexity in \Cref{tab:block complexity} and give the derivation in \Cref{app:complexities}. We will refer to these modules by indices, e.g. \circled{\small 2a} for the computation of output gradient.

\begin{table*}[!b]
    \centering
    \setlength\tabcolsep{2pt}
    \resizebox{\linewidth}{!}{    
    \begin{tabular}{c|c|c|c|c|c|c}
    \hline
     \multirow{2}{*}{Module} &\multirow{2}{*}{\circled{\tiny 1}Forward pass} &\multicolumn{2}{|c|}{\circled{\tiny 2}Back-propagation}&\multirow{2}{*}{\circled{\tiny 3}Ghost norm} &\multirow{2}{*}{\shortstack[c]{\circled{\tiny 4}Per-sample grad\\ instantiation}}&\multirow{2}{*}{\shortstack[c]{\circled{\tiny 5}Weighted sum of \\per-sample grad}} \\
        &&(a)output gradient&(b)parameter gradient&&& \\\hline
        Time complexity &$2BTpd$&$2BTpd$&$2BTpd$&$2BT^2(p+d)$&$2BTpd$&$2Bpd$ \\\hline
        Space complexity&$pd+BTd$&$BT(p+d)$&$pd$&$2BT^2$&$Bpd$&0 \\\hline
    \end{tabular}
    }
    \caption{Time and space complexity of modules in DP training for one generalized linear layer.}
    \label{tab:block complexity}
\end{table*}

Now we are ready to decompose each implementation, following the flowcharts in \Cref{fig:all algorithms}. Consequently, we can easily write down the complexity of different implementations in \Cref{tab:compare different implementations}. Such a modular analysis displays the clear effects of the tricks in BK algorithm: the ghost norm trick removes the memory costly \circled{\small 4} from Opacus and FastGradClip; the book-keeping trick removes the \circled{\small 2b} in the second  back-propagation of FastGradClip and GhostClip; the ghost differentiation trick removes the \circled{\small 2b} in the first back-propagation of Opacus and GhostClip.

\begin{itemize}
\item Standard (non-DP)$=\circled{\tiny 1}+\circled{\tiny 2a}+\circled{\tiny 2b}$
\item Opacus$=\circled{\tiny 1}+\circled{\tiny 2a}+\circled{\tiny 2b}+\circled{\tiny 4}+\circled{\tiny 5}$
\item FastGradClip$=\circled{\tiny 1}+\circled{\tiny 2a}+\circled{\tiny 4}+\circled{\tiny 2a}+\circled{\tiny 2b}$
\item GhostClip$=\circled{\tiny 1}+\circled{\tiny 2a}+\circled{\tiny 2b}+\circled{\tiny 3}+\circled{\tiny 2a}+\circled{\tiny 2b}$
\item BK (ours)$=\circled{\tiny 1}+\circled{\tiny 2a}+\circled{\tiny 3}+\circled{\tiny 2b}$
\end{itemize}

\subsection{BK is optimally efficient in low dimension}\label{sec:efficient BK in low dim}
When the feature dimension $T$ is small, we claim that BK is almost as efficient as the standard non-private training, with a negligible $O(BT^2)$ time and memory overhead by \Cref{tab:compare different implementations}:
\begin{align*}
\textbf{Memory complexity: }&\text{non-DP}\approx\text{BK}\approx\text{GhostClip}
\\
&<\text{FastGradClip}\ll\text{Opacus}
\\
\textbf{Time complexity: }&\text{non-DP}\approx\text{BK}<\text{FastGradClip}
\\
&\approx\text{Opacus}<\text{GhostClip}
\end{align*}
Now, we discuss the cases where the data has low dimension and thus $T$ is small. Generally speaking, the feature dimension $T_{(l)}$ depends on both the data and the model. 

\begin{figure}[!htb]
    \centering
    \includegraphics[width=0.6\linewidth]{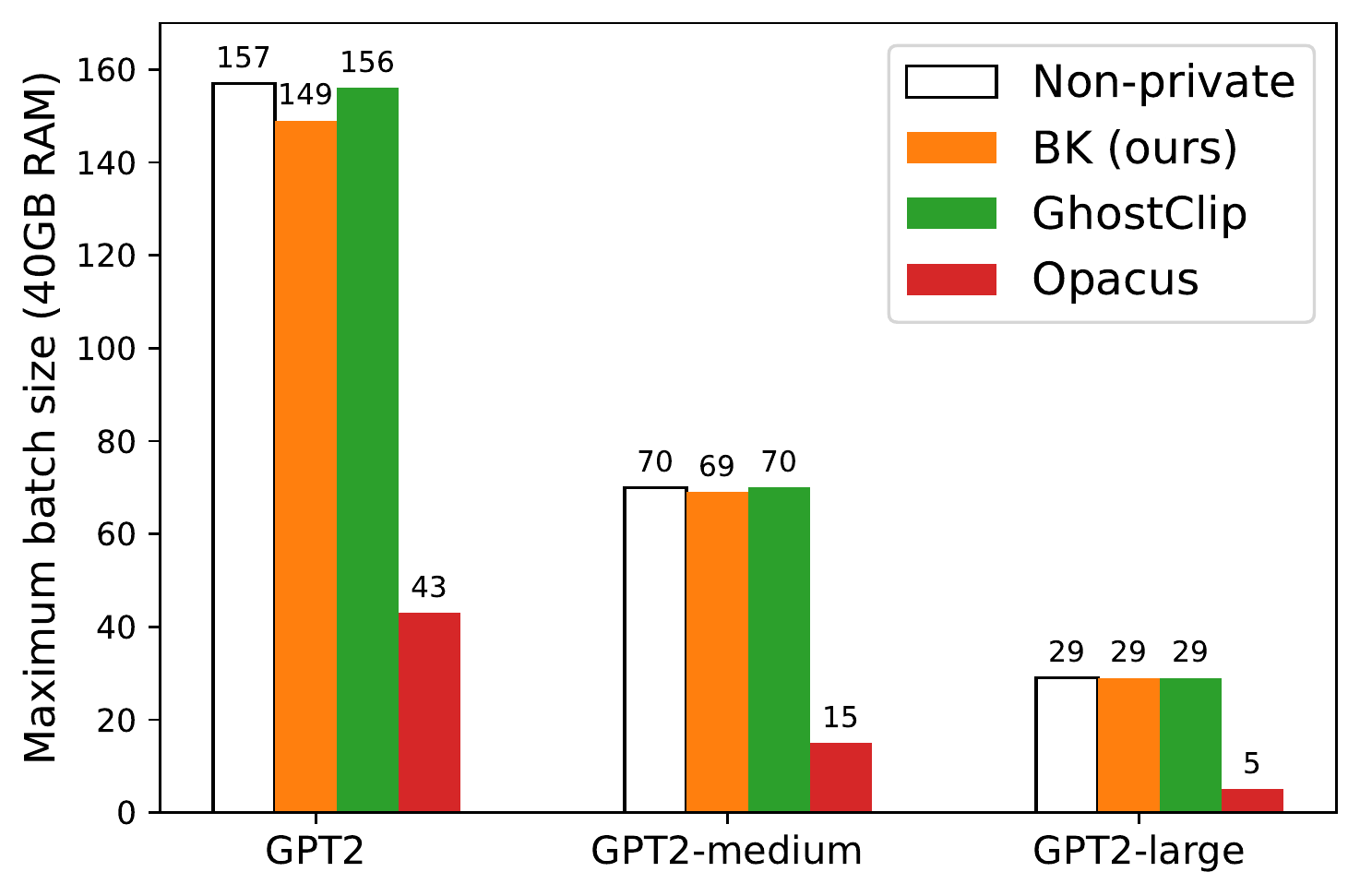}
    \includegraphics[width=0.6\linewidth]{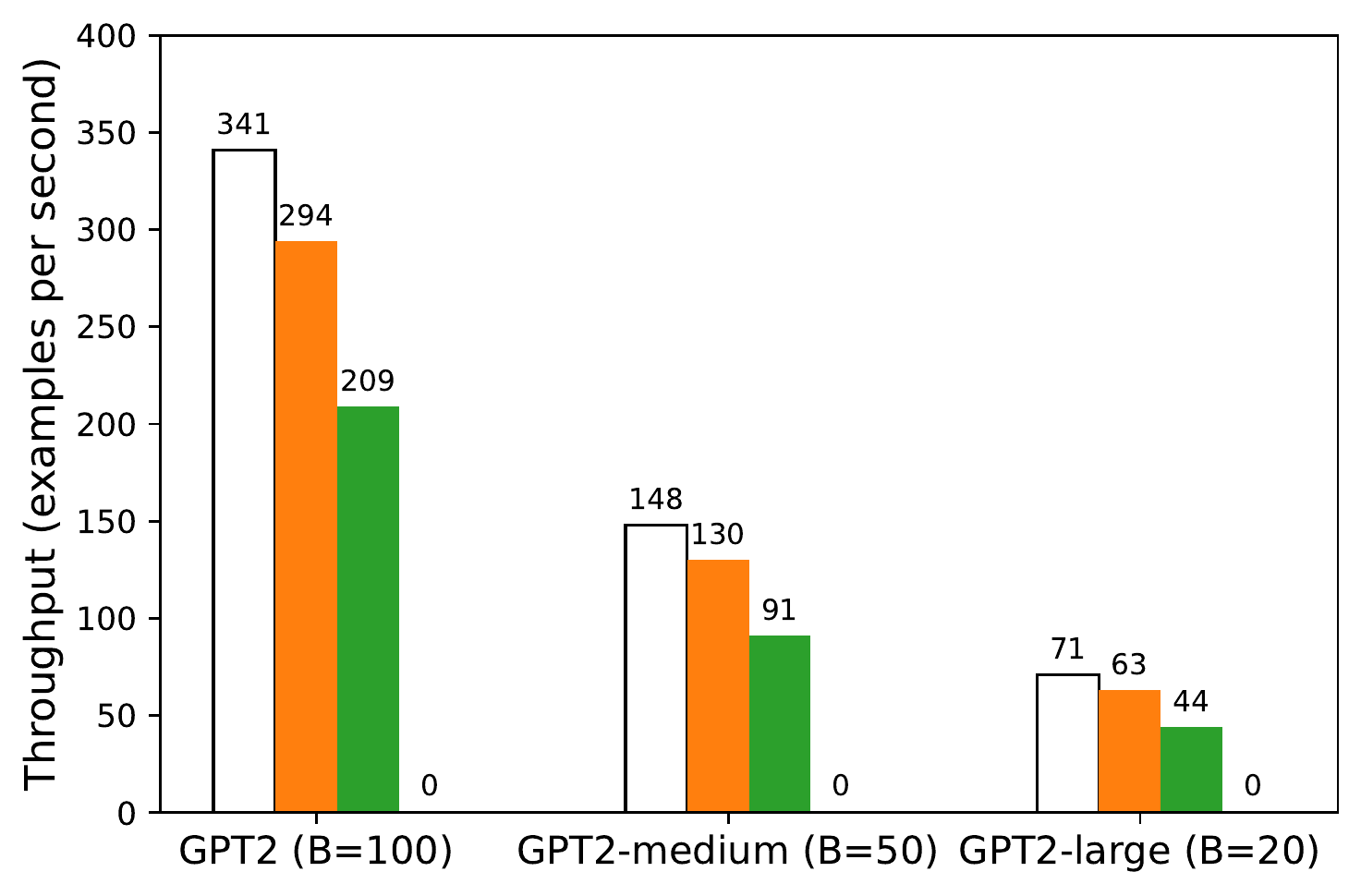}
 \includegraphics[width=0.6\linewidth]{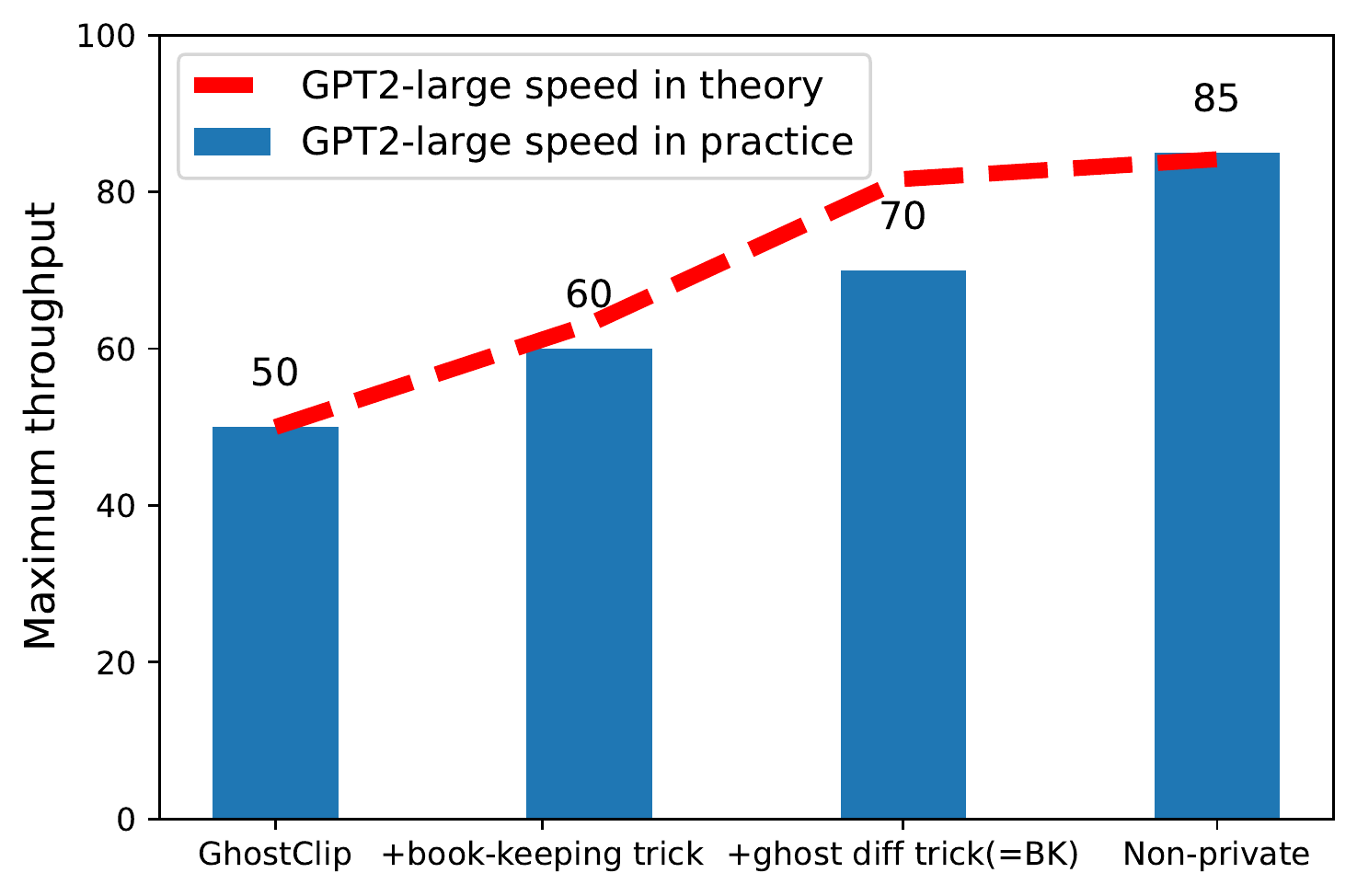}
    \\
    \includegraphics[width=0.99\linewidth]{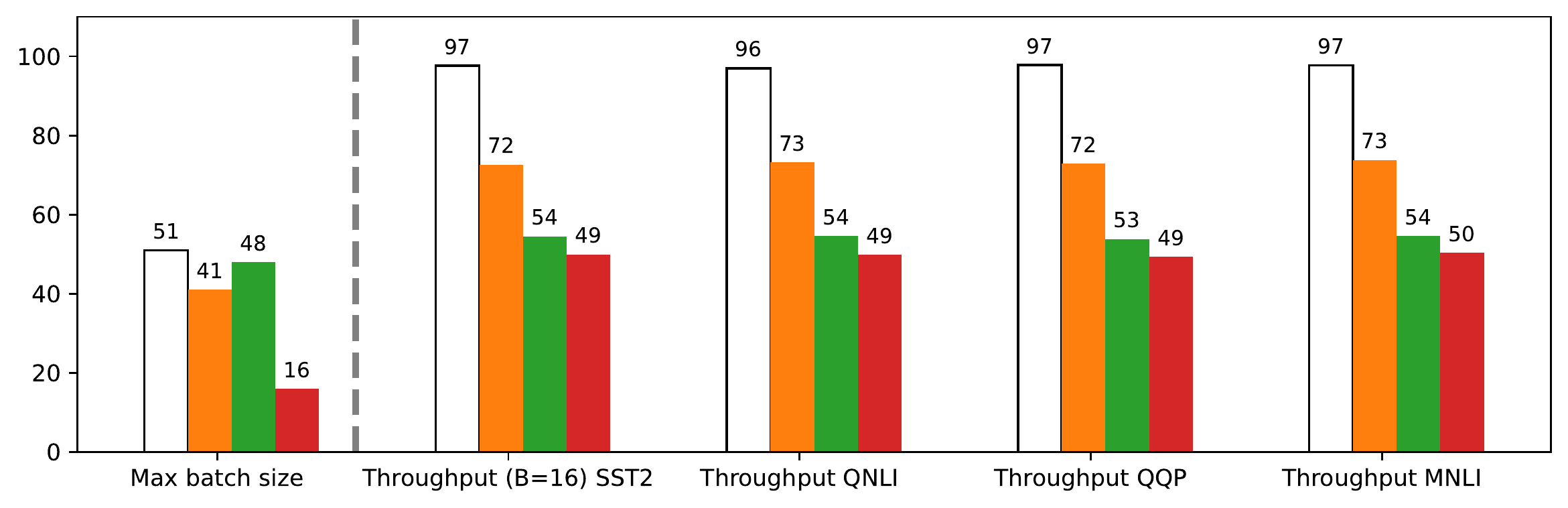}
    \vspace{-0.2cm}
    \caption{Memory and speed of different DP implementations. Upper: GPT2 on E2E dataset (fixing $B$, DP speed is $0.86\sim 0.89\times$ of non-DP). Lower: RoBERTa-large on GLUE datasets. Note here the hybrid implementations are equivalent to the base ones, because of the short sequence length.
    }
    \label{fig:NLP_memory_speed}
\end{figure}

For non-sequential input and 1D audio data, $T=1$. For sequential data such as texts ($T$ being sentence length) or time series ($T$ being time duration), $T_{(l)}$ is fixed across layers. In this case, BK is efficient on short-sequence datasets including GLUE \cite{wang2019glue} (e.g. SST2/QNLI/MNLI/QQP) and natural language generation datasets (e.g. E2E/DART), since $T^2\ll p_{(l)}d_{(l)}$. For instance, \cite{yu2021differentially,li2021large,bu2022automatic} applied GPT2 on E2E dataset, which has a sequence length $T\approx 100$ and the number of parameters $p_{(l)}d_{(l)}$ per layer is $2-4$M; \cite{yu2021differentially,li2021large} applied RoBERTa-large on GLUE datasets, which has a sequence length $T=256$ and the number of parameters per layer is $1-4$M. As illustrated in \Cref{fig:NLP_memory_speed} and \Cref{tab:SOTA throughput} (extended in \Cref{tab:GPT max throughput}), BK improves the throughput of existing implementations by $25- 388\%$ on multiple language tasks in \cite{li2021large,bu2022automatic}, with minor memory overhead compared to GhostClip and non-private training. 

However, on the convolution layers with image data, $T_{(l)}$ is the product of hidden feature sizes (c.f. Section 3 in \cite{bu2022scalable}), thus $T_{(l)}$ depends on the original image size and network architecture. For example, larger kernel size/dilation/stride in convolution layer reduces $T_{(l)}$, while larger images have larger $T_{(l)}$ at each layer. Therefore, BK (and GhostClip) may suffer on when training ResNet on ImageNet $(224\times 224)$, as we show in \Cref{fig:memory_time_vs_sequence_beit} (see also Table 7 in \cite{bu2022scalable}), although training the same network efficiently on CIFAR10/100 $(32\times 32)$.

\subsection{Applying our tricks to existing implementations}
\label{sec:apply to others}
Our tricks in \Cref{sec:tricks} can also improve other existing implementations, reducing the time complexity of GhostClip from $10BTpd+2BT^2(p+d)$ to $6BTpd+2BT^2(p+d)$, that of Opacus and FastGradClip from $8BTpd$ to $6BTpd$. We highlight that these improved implementations are leveraged to design hybrid implementation in \Cref{sec:hybrid}. In addition to DP full fine-tuning, BK is demonstrated in \Cref{app:param eff BK} to also apply to the parameter efficient fine-tuning like Adapters \cite{houlsby2019parameter} and LoRA \cite{hu2021lora}.
\begin{align*}
\begin{split}
\text{GhostClip} &= \circled{\tiny 1}+\circled{\tiny 2a}+\circled{\tiny 2b}+\circled{\tiny 3}+\circled{\tiny 2a}+\circled{\tiny 2b}
\\
&\xrightarrow[\text{\tiny book-keeping}]{\makebox[2cm]{\tiny\text{ghost differentiation}}} \circled{\tiny 1}+\circled{\tiny 2a}+\circled{\tiny 3}+\circled{\tiny 2b} \text{ (BK)}
\\
\text{Opacus} &= \circled{\tiny 1}+\circled{\tiny 2a}+\circled{\tiny 2b}+\circled{\tiny 4}+\circled{\tiny 5}
\\
&\xrightarrow{\makebox[2cm]{\tiny\text{ghost differentiation}}} \circled{\tiny 1}+\circled{\tiny 2a}+\circled{\tiny 4}+\circled{\tiny 5}
\\
\text{FastGradClip} &= \circled{\tiny 1}+\circled{\tiny 2a}+\circled{\tiny 4}+\circled{\tiny 2a}+\circled{\tiny 2b} \\
&\xrightarrow{\makebox[2cm]{\tiny\text{book-keeping}}}\circled{\tiny 1}+\circled{\tiny 2a}+\circled{\tiny 4}+\circled{\tiny 2b}
\end{split}
\end{align*}

\section{Hybrid Book-keeping: Efficient DP training in high dimension}
In previous section, we have analyzed DP implementations in the small $T$ regime, where the ghost norm-based GhostClip and BK are efficient. Nevertheless, in the large $T$ and large model regime, none of the base implementations may be efficient (see  \Cref{fig:memory_time_vs_sequence_beit}) and we turn to hybrid methods.

\subsection{Large $T$ necessitates non-ghost norm method}
\begin{figure}[!htb]
\centering
    \includegraphics[width=0.49\linewidth]{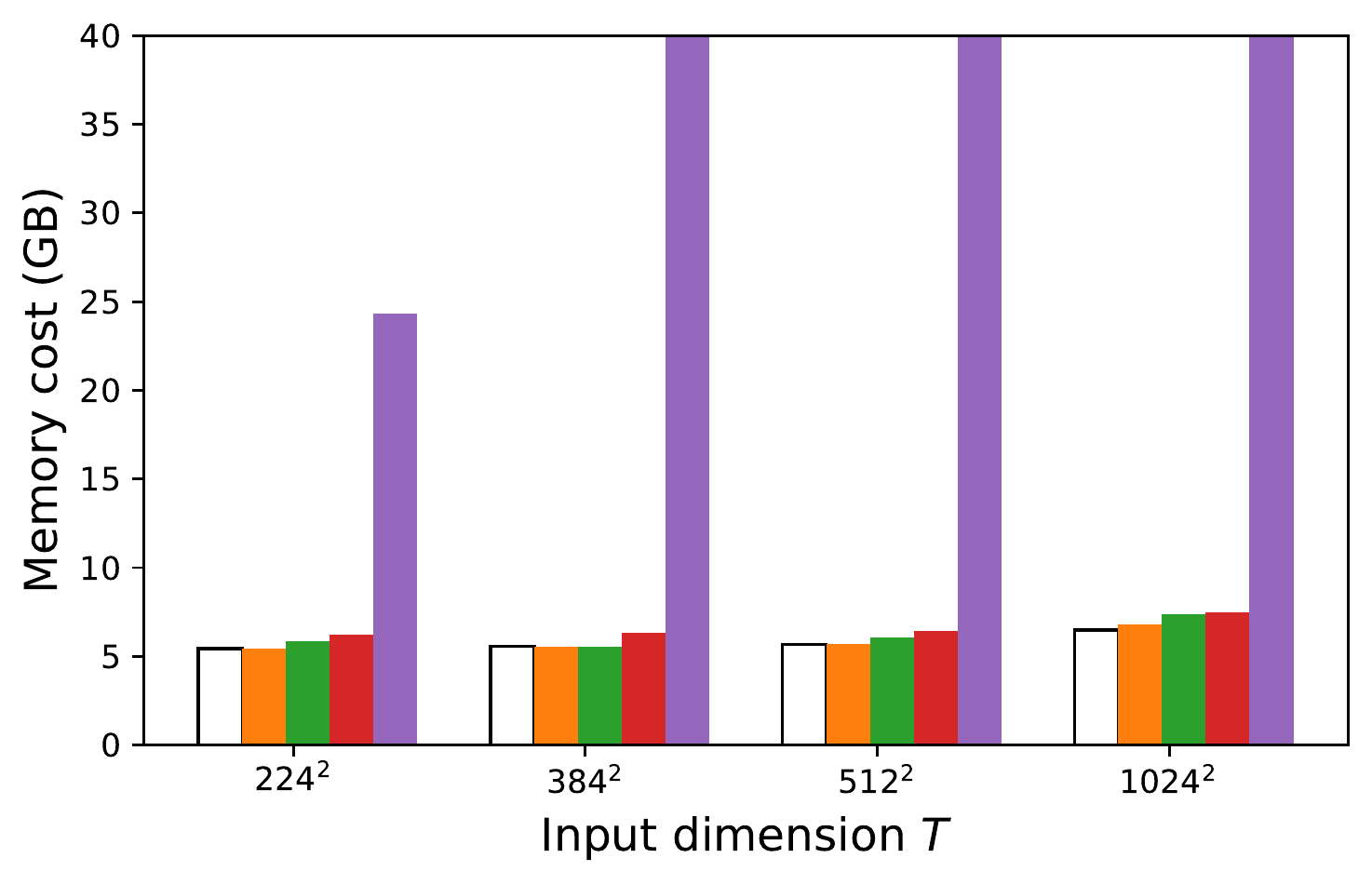}
    \includegraphics[width=0.49\linewidth]{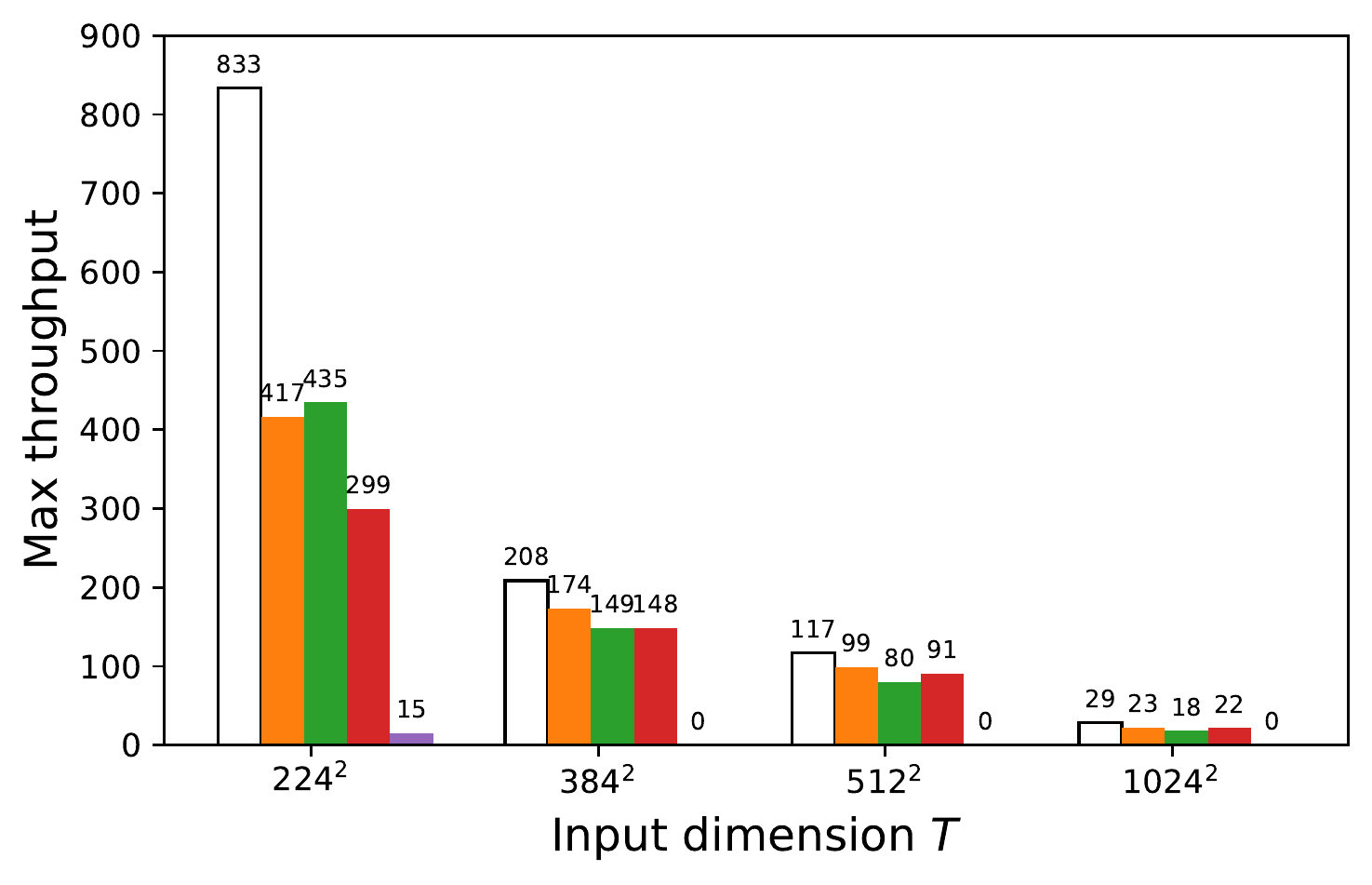}
    \\
    \includegraphics[width=0.49\linewidth]{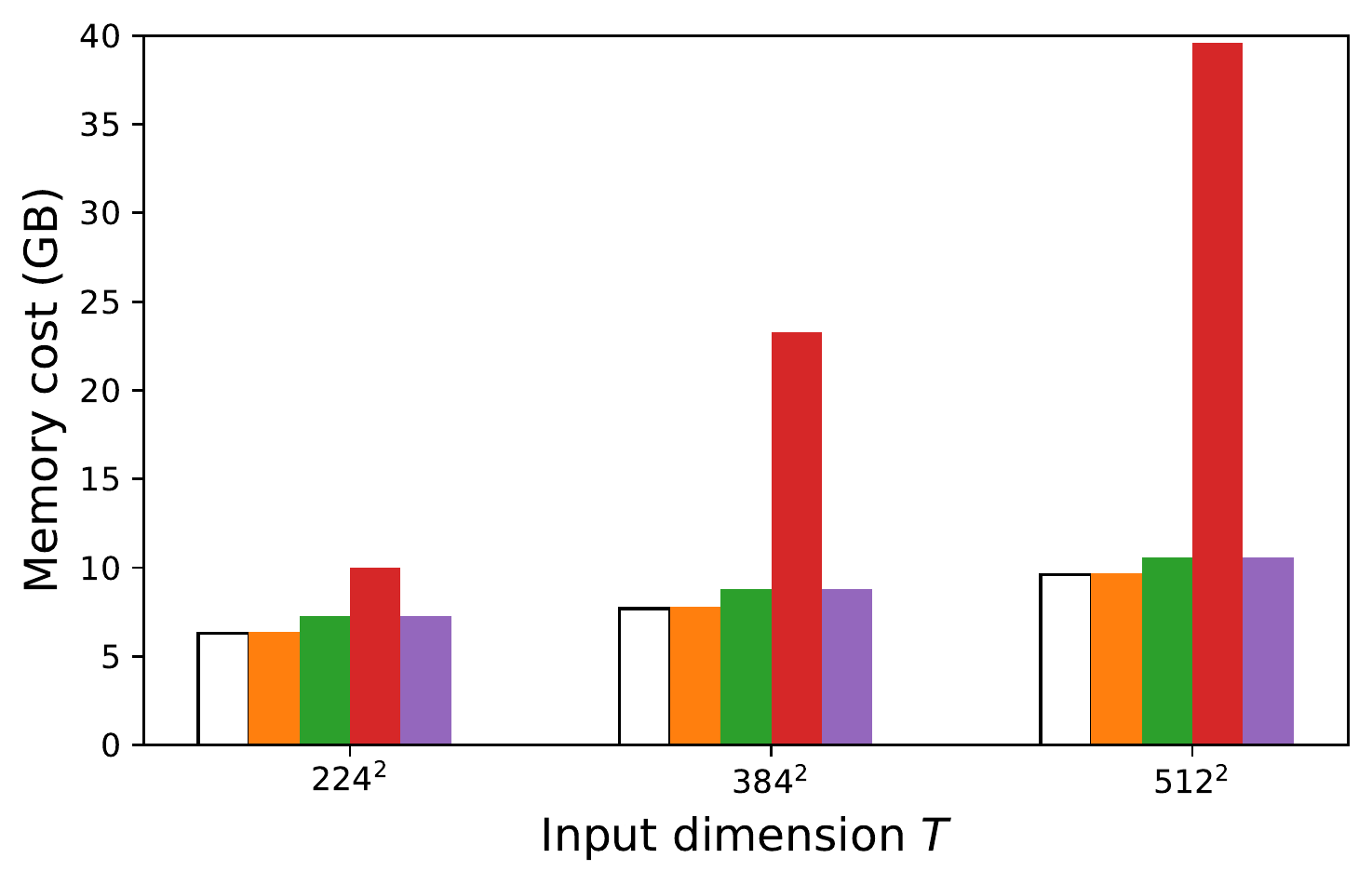}
    \includegraphics[width=0.49\linewidth]{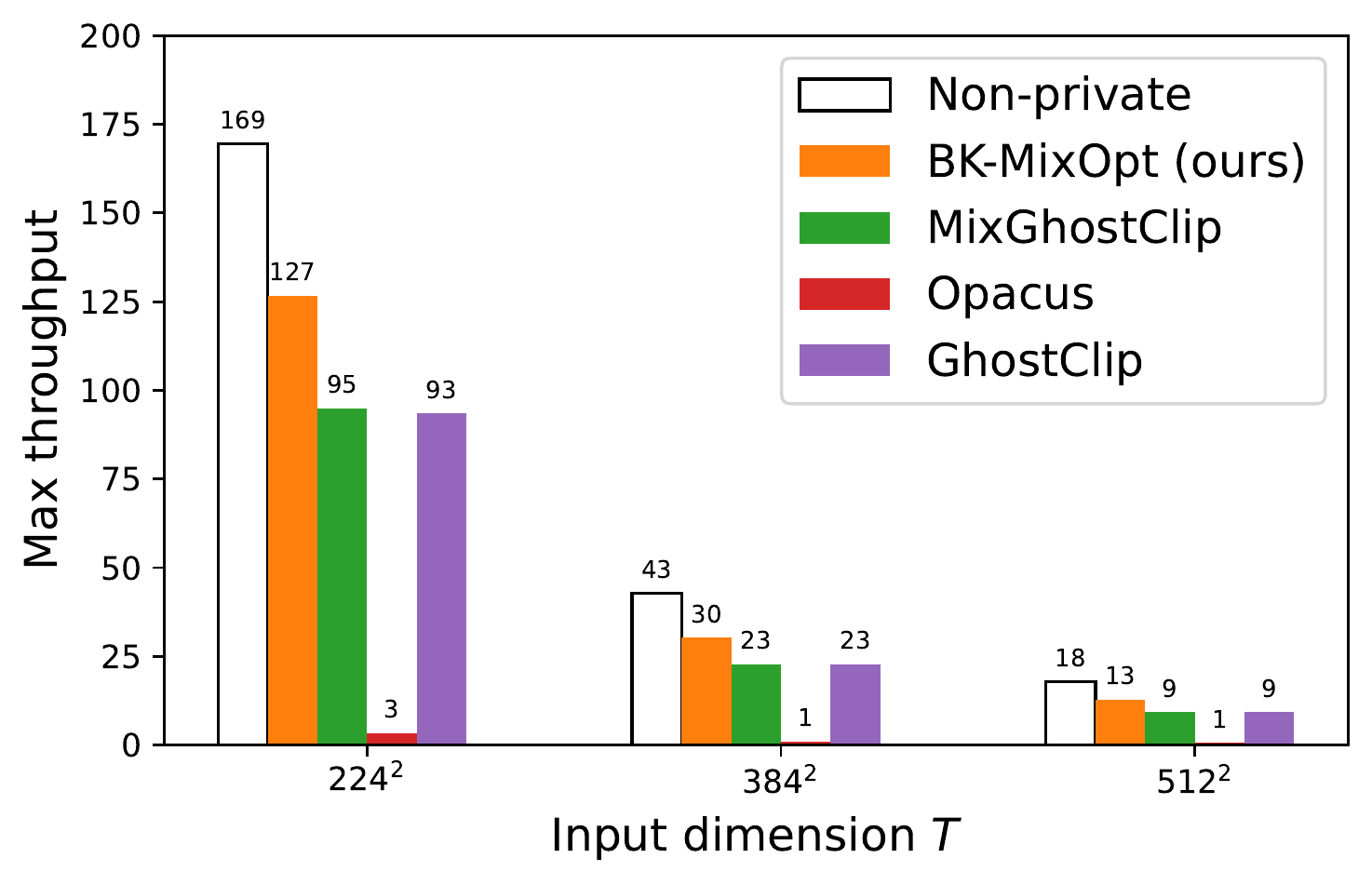}    
    \vspace{-0.2cm}
    \caption{Memory and speed by different implementations on 50000 images. Top is VGG11 (133M; \cite{simonyan2014very}). Bottom is BEiT-large (304M; \cite{bao2021beit}). Memory cost is measured with physical batch size 1. Throughput is measured with the maximum physical batch size on 40GB memory.}
    \label{fig:memory_time_vs_sequence_beit}
\end{figure}

A closer look at the space complexity in \Cref{tab:block complexity} shows that, the ghost norm trick is favored over the per-sample gradient instantiation if and only if $2T^2_{(l)}<p_{(l)}d_{(l)}$, where $p_{(l)}d_{(l)}$ is the number of parameters at one layer. When this criterion is violated for large $T$, GhostClip/BK (base) can significantly under-perform Opacus/FastGradClip, as shown in \Cref{fig:memory_time_vs_sequence_beit}, \Cref{fig:layerwise_decision} and \Cref{tab:layerwise complexity}.

Similar to \Cref{sec:efficient BK in low dim}, we discuss two cases where $T$ is large. For paragraph or document-level language tasks like WikiHop \cite{welbl2018constructing} and TriviaQA \cite{joshi2017triviaqa}, $T$ can range from $2000\sim 20000$ to train large language models, which makes $2T^2=8\sim 800$M. For example, LLAMA \cite{touvron2023llama} is trained with token length $4096\leq T\leq 8192$ and GPT-3 \cite{brown2020language} is trained with token length $T=2048$.

For image tasks, particularly on CNN, $T_{(l)}$ varies at each layer with large values on top layers, as the features are less compressed by convolution and pooling. Taking ImageNet and the first convolution layer of VGG11 as an example (see Table 3 of \cite{bu2022scalable}), $2T_{(1)}^2=5\times 10^9\gg p_{(1)}d_{(1)}=1.7\times 10^3$. Consequently, ghost norm-based implementations (i.e. GhostClip and BK) costs more than 40GB memory on ResNet18, under $B=32$, while Opacus only costs 2.5GB. This curse of dimension grows from a difficult issue on ImageNet to an impossible challenge on videos or high-resolution images, e.g. GhostClip cannot train ResNet18 with even one single CelebA-HQ image ($1024\times 1024$) using a 40GB GPU.

In short, the ghost norm trick is inefficient for large $T$ and the per-sample gradient instantiation is inefficient for large model. Hence, we must hybridize the base implementations.

\subsection{Hybrid implementations via layerwise decision}
\label{sec:hybrid}
We adopt the same layerwise decision as \cite{bu2022scalable}, known as the mixed ghost norm technique: we use the ghost norm trick on a layer if $2T_{(l)}^2<p_{(l)}d_{(l)}$, and instantiate per-sample gradients otherwise. Therefore, the space complexity of computing the per-sample gradient norm reduces to $\min\{2T_{(l)}^2,p_{(l)}d_{(l)}\}$, which is significantly cheaper than either the ghost norm or the per-sample gradient instantiation in high dimension, as depicted in \Cref{tab:layerwise decision} and \Cref{fig:layerwise_decision}. Consequently, over all layers, the space complexity is lower than both constituting methods, e.g. saving more than $10\times$ memory for the per-sample gradient clipping on ResNet18 (see more models in \Cref{tab:layerwise complexity}).

\begin{table*}[!htb]
\vspace{-0.2cm}
\setlength\tabcolsep{2pt}
    \centering
    \resizebox{0.95\linewidth}{!}{
\begin{tabular}{|c|c||c|c||c|c||c|c|}
 \hline
  &Output size& \multicolumn{6}{|c|}{Space complexity}\\ \hline
    &\multirow{3}{*}{\shortstack{$H_\text{out}\times W_\text{out}$}}
    & \multicolumn{2}{|c|}{18-layer}& \multicolumn{2}{|c|}{34-layer}
    & \multicolumn{2}{|c|}{50-layer}
    \\\cline{3-8}
    &&Ghost norm & \shortstack{Per-sample grad\\ instantiation}&Ghost norm & \shortstack{Per-sample grad\\ instantiation}&Ghost norm & \shortstack{Per-sample grad\\ instantiation} \\ 
    && \multirow{2}{*}{$2T_{(l)}^2=2H_\text{out}^2W_\text{out}^2$}& \multirow{2}{*}{$p_{(l)} d_{(l)}= $ \# params}& \multirow{2}{*}{$2T_{(l)}^2$}& \multirow{2}{*}{$p_{(l)} d_{(l)}$}& \multirow{2}{*}{$2T_{(l)}^2$}& \multirow{2}{*}{$p_{(l)} d_{(l)}$}\\
    &&&&&&&\\\hline\hline
 conv1&$112\times 112$&$3.1\times 10^8$ & $\bm{9.4\times 10^3}$
 &$3.1\times 10^8$ & $\bm{9.4\times 10^3}$
 &$3.1\times 10^8$ & $\bm{9.4\times 10^3}$ \\ \hline

 \multirow{3}{*}{conv2\_x}&\multirow{3}{*}{$56\times 56$}
 &\multirow{3}{*}{$[2.0\times 10^7] \times 4$} & \multirow{3}{*}{$\bm{[3.7\times 10^4]\times 4}$}
 &\multirow{3}{*}{$[2.0\times 10^7] \times 6$} & \multirow{3}{*}{$\bm{[3.7\times 10^4]\times 6}$}
 &\multirow{3}{*}{$[2.0\times 10^7] \times 9$} & \multirow{3}{*}{\shortstack{
        $\bm{[4.1\times 10^3]\times 1}$
      \\
      $\bm{[3.7\times 10^4]\times 3}$
      \\
      $\bm{[1.6\times 10^4]\times 5}$
     }}\\ 
 &&&&&&& \\ 
 &&&&&&& \\ 
 &&&&&&& \\ \hline

 \multirow{3}{*}{conv3\_x}&\multirow{3}{*}{$28\times 28$}&\multirow{3}{*}{$[1.2\times 10^6]\times 4$}& \multirow{3}{*}{\shortstack{$\bm{[7.4\times 10^4]\times 1}$\\$\bm{[1.5\times 10^5]\times 3}$}}
 &\multirow{3}{*}{$[1.2\times 10^6]\times 8$} & \multirow{3}{*}{\shortstack{$\bm{[7.4\times 10^4]\times 1}$\\$\bm{[1.5\times 10^5]\times 7}$}}
 &\multirow{3}{*}{\shortstack{${[2.0\times 10^7]\times 1}$\\${[1.2\times 10^6]\times 11}$}} & \multirow{3}{*}{\shortstack{
        $\bm{[3.3\times 10^4]\times 1}$
      \\
      $\bm{[6.6\times 10^4]\times 7}$
      \\
      $\bm{[1.5\times 10^5]\times 4}$
     }}
 \\ 
  &&&&&&& \\ 
 &&&&&&& \\ 
 &&&&&&& \\ \hline

 \multirow{3}{*}{conv4\_x}&\multirow{3}{*}{$14\times 14$}&\multirow{3}{*}{$\bm{[7.7\times 10^4]\times 4}$}& \multirow{3}{*}{\shortstack{${[2.9\times 10^5]\times 1}$\\${[5.9\times 10^5]\times 3}$}}
 &\multirow{3}{*}{$\bm{[7.7\times 10^4]\times 12}$} & \multirow{3}{*}{\shortstack{${[2.6\times 10^5]\times 1}$\\${[5.9\times 10^5]\times 11}$}}
 &\multirow{3}{*}{\shortstack{${[1.2\times 10^6]\times 1}$\\$\bm{[7.7\times 10^4]\times 17}$}} & \multirow{3}{*}{\shortstack{
        $\bm{[1.3\times 10^5]\times 1}$
      \\
      ${[2.6\times 10^5]\times 11}$
      \\
      ${[5.9\times 10^5]\times 6}$
     }}
\\ 
 &&&&&&& \\ 
 &&&&&&& \\ 
 &&&&&&& \\ \hline
 
 \multirow{3}{*}{conv5\_x}&\multirow{3}{*}{$7\times 7$}&\multirow{3}{*}{$\bm{[4.8\times 10^3]\times 4}$}& \multirow{3}{*}{\shortstack{${[1.2\times 10^6]\times 1}$\\${[2.4\times 10^6]\times 3}$}}
 &\multirow{3}{*}{$\bm{[4.8\times 10^3]\times 6}$} & \multirow{3}{*}{\shortstack{${[1.2\times 10^6]\times 1}$\\${[2.4\times 10^6]\times 5}$}}
 &\multirow{3}{*}{$\bm{[4.8\times 10^3]\times 9}$} & \multirow{3}{*}{\shortstack{
        $[5.2\times 10^5]\times 1$
      \\
      ${[1.0\times 10^6]\times 5}$
      \\
      ${[2.4\times 10^6]\times 3}$
     }}
\\ 
  &&&&&&& \\ 
 &&&&&&& \\ 
 &&&&&&& \\ \hline
 
linear&$1\times 1$&$\bm{2}$ & $5.1\times 10^5$&$\bm{2}$ & $5.1\times 10^5$&$\bm{2}$ & $2.0\times 10^6$\\ \hline\hline
\multicolumn{2}{|c||}{Total complexity}&399M&11.5M&444M&21.6M&528M&22.7M\\\hline
\multicolumn{2}{|c||}{Complexity by}&\multicolumn{2}{|c||}{\multirow{2}{*}{1.0M}}&\multicolumn{2}{|c||}{\multirow{2}{*}{2.3M}}&\multicolumn{2}{|c|}{\multirow{2}{*}{2.8M}}\\
\multicolumn{2}{|c||}{ mixed ghost norm}&\multicolumn{2}{|c||}{}&\multicolumn{2}{|c||}{}&\multicolumn{2}{|c|}{}\\\hline
    \end{tabular}
    }
    \vspace{-0.3cm}
    \caption{Space complexity of the per-sample gradient clipping (not the entire DP algorithm) for $B=1$ on ImageNet $224\times 224$. Layerwise decision of hybrid BK algorithms is highlighted in \textbf{bold}.}
    \label{tab:layerwise decision}
    \vspace{-0.3cm}
\end{table*}

In contrast to the mixed ghost clipping (MixGhostClip) in \cite{bu2022scalable}, which hybridizes FastGradClip and GhostClip, we boost the efficiency by hybridizing our BK with the improved FastGradClip/Opacus in \Cref{sec:apply to others}. 
We propose BK-MixOpt (and BK-MixGhostClip as an intermediate product only for comparison) and use MixGhostClip as a reference point,

\begin{table*}[!htb]
\setlength\tabcolsep{2pt}
\vspace{0.3cm}
\centering
    \resizebox{0.9\linewidth}{!}{%
    \begin{tabular}{c|c|c||c|c}
        Method&Type&Modification to previous variant&Time complexity&Space complexity
        \\\hline\hline
        Non-DP&&&$6BTpd$&$pd+3BTd+BTp$
        \\\hline\hline
        Opacus&\multirow{6}{*}{base}&Instantiate per-sample gradient&$8BTpd$&$+Bpd$
        \\\cline{1-1}\cline{3-5}
        \multirow{2}{*}{FastGradClip}&&\multirow{2}{*}{\shortstack{Not store per-sample gradient \\using a second back-propagation}}&\multirow{2}{*}{$8BTpd$}&\multirow{2}{*}{$+Bpd$}
        \\        &&&&
        \\\cline{1-1}\cline{3-5}
        \multirow{2}{*}{GhostClip}&&\multirow{2}{*}{\shortstack{Not instantiate per-sample gradient \\using ghost norm trick}}&\multirow{2}{*}{$10BTpd+2BT^2(p+d)$}&\multirow{2}{*}{$+2BT^2$}
        \\        &&&&
        \\\cline{1-1}\cline{3-5}
        BK (ours)&&Simplify the two back-propagations&$6BTpd+2BT^2(p+d)$&$+2BT^2$
\\\hline
MixGhostClip&\multirow{3}{*}{hybrid}&\multirow{2}{*}{Mix ways to compute grad norm}&$8BTpd+\langle2BTpd,2BT^2(p+d)\rangle$&$+\min\{2BT^2,Bpd\}$
\\\cline{1-1}\cline{4-5}
BK-MixGhostClip&&&$6BTpd+\langle 2BTpd,2BT^2(p+d)\rangle$&$+\min\{2BT^2,Bpd\}$
\\\cline{1-1}\cline{3-5}
BK-MixOpt&&Mix ways to compute weighted grad&$6BTpd+\langle0,2BT^2(p+d)\rangle$&$+\min\{2BT^2,Bpd\}$
\\\hline\hline
    \end{tabular}
    }
    \vspace{-0.3cm}
    \caption{Complexity of DP implementations on one layer. Here $\langle\rangle$ means between two values. The exact time complexity of BK-MixOpt is $6BTpd+2BT^2(p+d)\cdot\mathbb{I}\{2T^2<pd\}\approx 6BTpd$. The space complexity of DP algorithms is in addition to that of non-DP one.}
    \label{tab:detailed BK and mixed and everything}
    \vspace{-0.05cm}
\end{table*}

\vspace{0.1cm}
\begin{itemize}
\item MixGhostClip $= \circled{\tiny 1}+\circled{\tiny 2a}+\circled{\tiny 2b}+\min\left\{\circled{\tiny 3},\circled{\tiny 4}\right\}+\circled{\tiny 2a}+\circled{\tiny 2b}\approx \min\{\text{GhostClip, FastGradClip}\}$,
\item BK-MixGhostClip $= \circled{\tiny 1}+\circled{\tiny 2a}+\min\left\{\circled{\tiny 3},\circled{\tiny 4}\right\}+\circled{\tiny 2b} = \min\{\text{BK, improved FastGradClip in \Cref{sec:apply to others}}\}$,
\item BK-MixOpt $= \circled{\tiny 1}+\circled{\tiny 2a}+\min\left\{\circled{\tiny 3}+\circled{\tiny 2b},\circled{\tiny 4}+\circled{\tiny 5}\right\} = \min\{\text{BK, improved Opacus in \Cref{sec:apply to others}}\}$.
\end{itemize}

The hybrid BK algorithms are presented in \Cref{alg:BKAL-mixed}. We summarize the layerwise complexity in \Cref{tab:detailed BK and mixed and everything}, from which we derive the overall complexity in \Cref{tab:complexity overhead} and observe that BK has almost the same complexity as non-DP training. Note that in low dimension, the mixed ghost norm is equivalent to the ghost norm, hence MixGhostClip/BK-MixOpt is equivalent to GhostClip/BK, respectively.

\subsection{Effect of model architecture \& feature dimension on hybridization}

We dive deeper to understand when the hybridization favors the ghost or non-ghost norm tricks.

\begin{figure*}[!htb]
    \centering
    \includegraphics[width=0.245\linewidth]{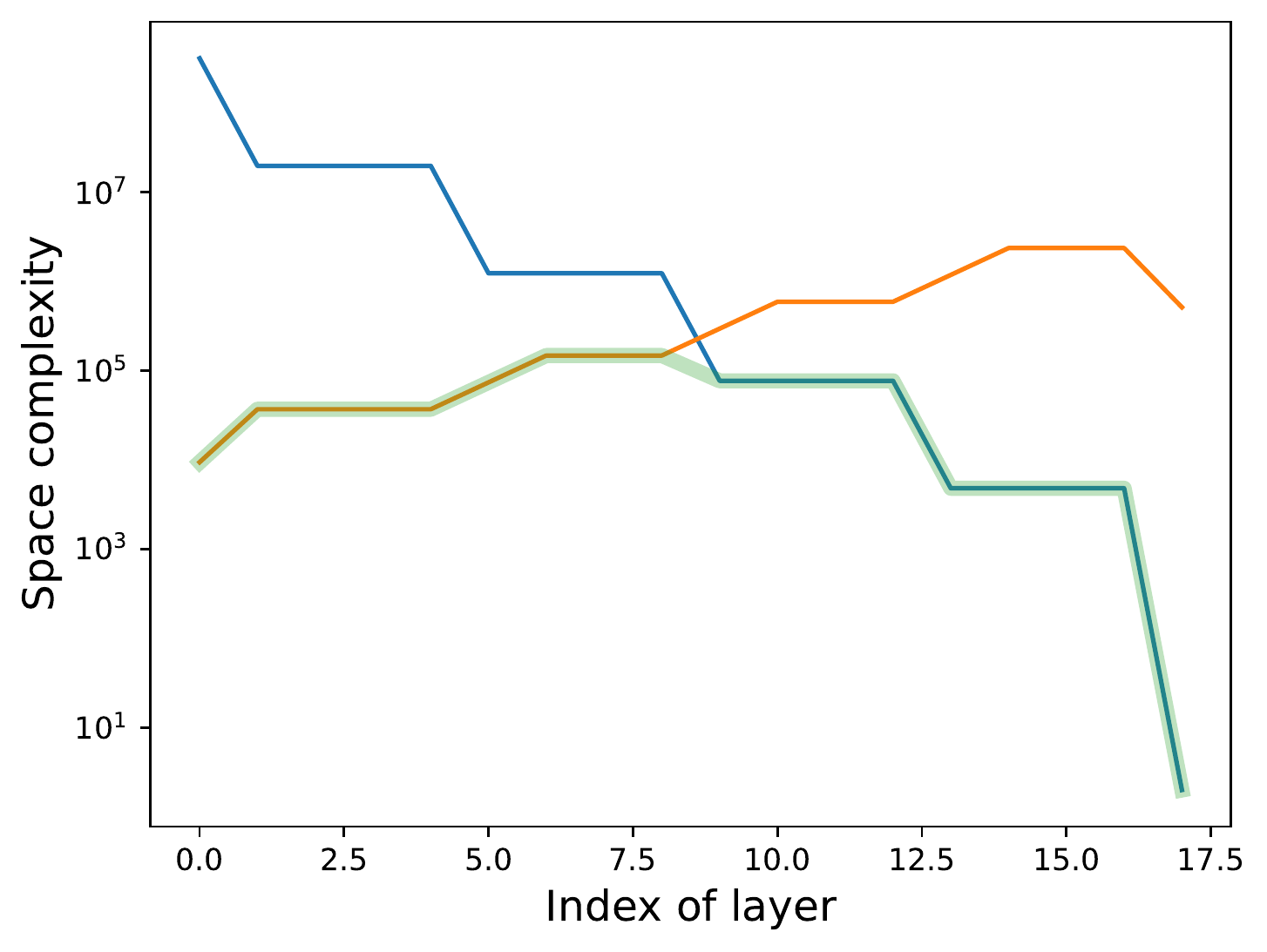}
    \includegraphics[width=0.245\linewidth]{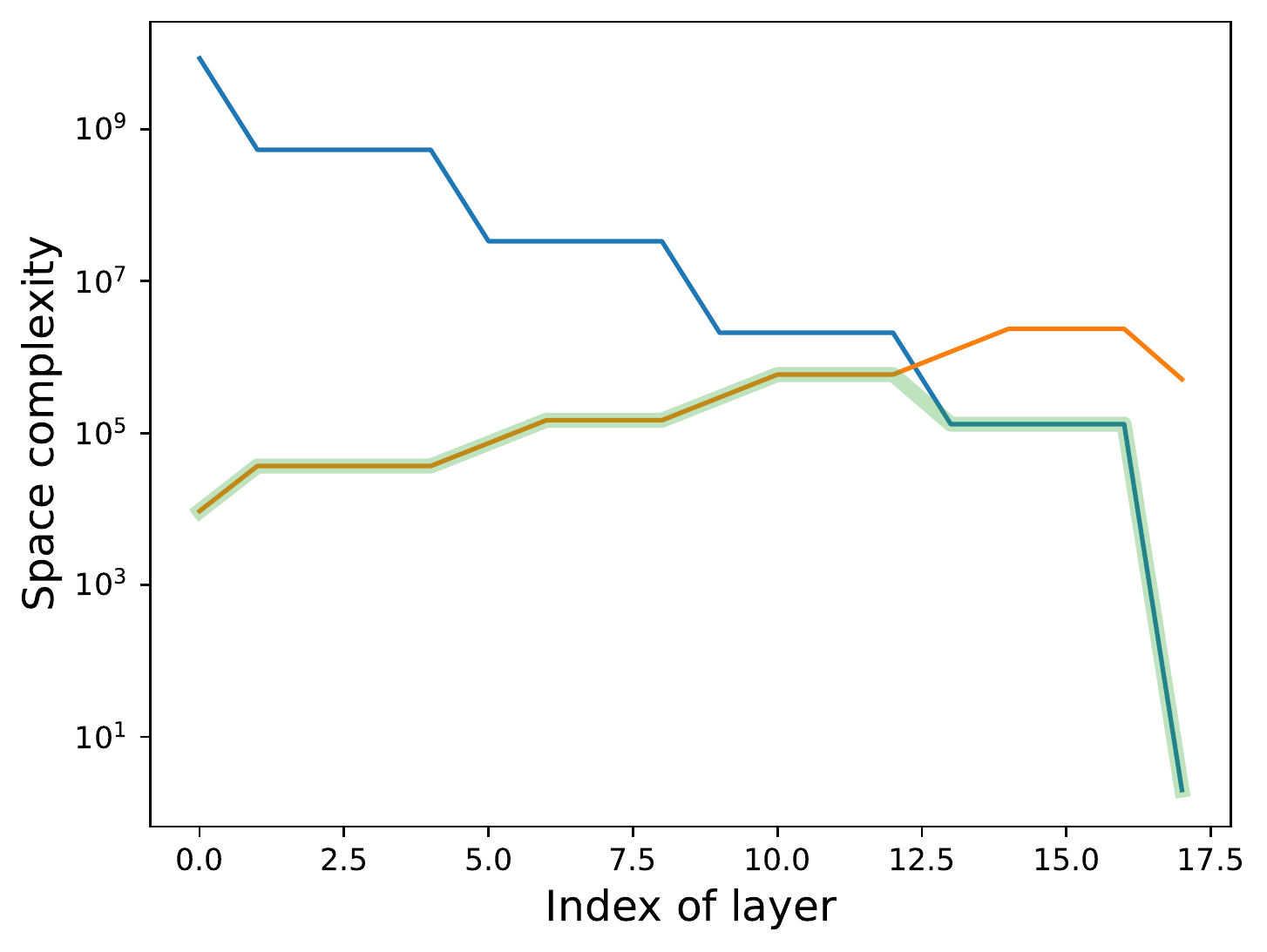}
    \includegraphics[width=0.245\linewidth]{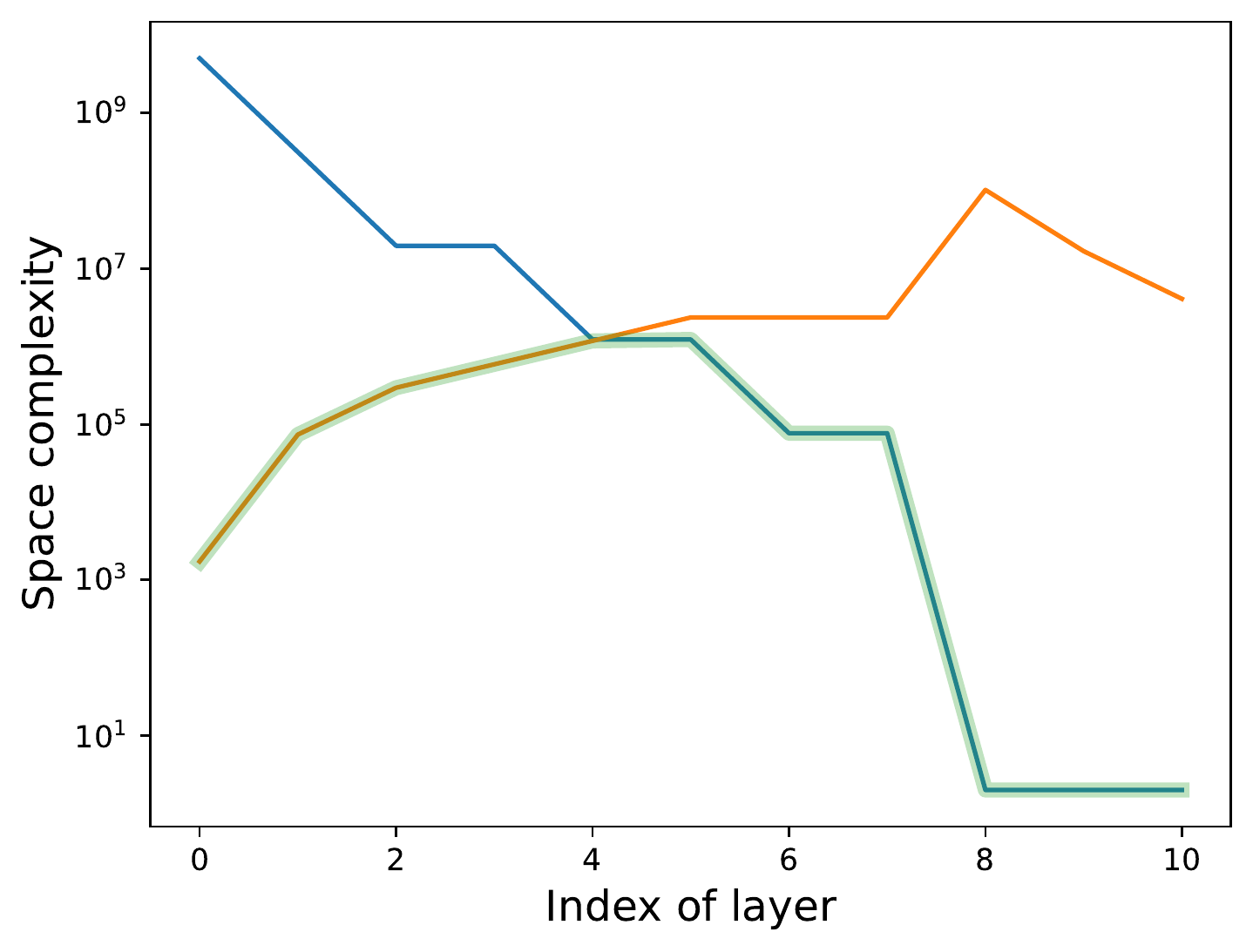}
    \includegraphics[width=0.245\linewidth]{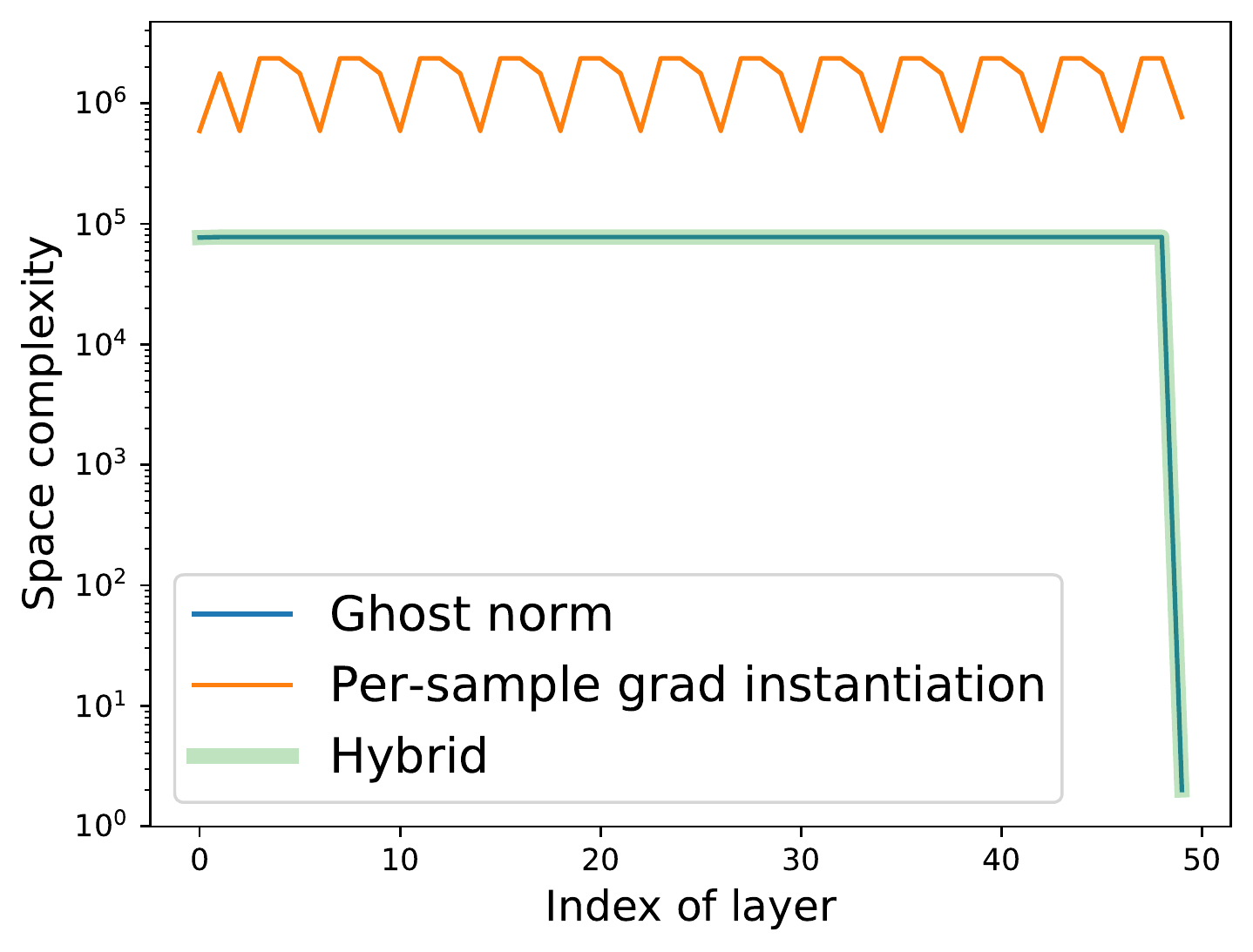}
    \vspace{-0.2cm}
    \caption{Layerwise space complexity of computing the per-sample gradient norm. Left to right: ResNet18 ($224\times 224$), ResNet18 ($512\times 512$), VGG11 ($224\times 224$), and ViT-base ($224\times 224$).}
    \label{fig:layerwise_decision}
\end{figure*}

From a model architecture viewpoint, transformers such as ViT, RoBERTa, GPT tend to prefer the ghost norm: for moderate-sequence text data and moderate-dimension image data, hybrid BK algorithms are close or equivalent to the base BK algorithm (see right-most plot in \Cref{fig:layerwise_decision}). However, CNN prefers the per-sample gradient instantiation at top layers, and there exists a depth threshold below which the ghost norm is more efficient. Hence the hybridization is necessary to take advantages of both worlds. 

From the feature dimension viewpoint, larger input enlarges this depth threshold, e.g. from the 9-th layer of ResNet18 to the 17-th layer in \Cref{fig:layerwise_decision}, when the image size increases from $224\times 224$ to $512\times 512$. We visualize this pattern on various models in \Cref{app:effect of hybrid}. In particular, we observe in \Cref{tab:complexity overhead} that when $T$ is large, both per-sample gradient instantiation (Opacus) and ghost norm trick (GhostClip) are significantly dominated by our BK algorithms.

\section{Instructions to use the codebase}

In this section, we demonstrate how to modify a standard training script to its DP variants\footnote{That is, our codebase can easily adapt to any per-sample gradient clipping function and privacy accouting methods.} by \textcolor{ForestGreen}{one piece of code}. 

\begin{Verbatim}[fontsize=\small,breaklines=true,commandchars=\\\{\},frame=single]
from fastDP import PrivacyEngine
import torch.functional as F

optimizer = torch.optim.Adam(model.parameters())

\textcolor{ForestGreen}{privacy_engine = PrivacyEngine(
    model,epochs,
    batch_size,sample_size,
    target_epsilon,target_delta)
}
\textcolor{ForestGreen}{privacy_engine.attach(optimizer)}

logits = model(data)
loss = F.cross_entropy(logits, labels)
loss.backward()
optimizer.step()
optimizer.zero_grad()
\end{Verbatim}

We highlight that our codebase automatically modifies the training for any network architecture and any optimizer. Additionally, it is designed to work compatibly with large-scale training techniques, such as the gradient accumulation, the parameter-efficient fine-tuning (e.g. LoRA and BiTFiT \cite{bu2022differentially}), and the parallel distributed learning (e.g. ZeRO \cite{bu2023zero}).

\section{Discussion}
In this work, we propose the Book-Keeping (BK) algorithms to effciently implement DP optimizers using three tricks: ghost norm, book-keeping, and ghost differentiation. Our BK reduces the time and space complexity of DP training to the similar level of the standard training. Specially, we develop hybrid BK to overcome the computational challenge of training large models with high-dimensional data, and we extend BK to parameter efficient fine-tuning such as LoRA and Adapter.

One limitation of this work is that BK (and GhostClip) only applies to the weights, not the biases, and only on the generalized linear layers, i.e. the embedding, the linear, and the convolution layers. However, this is a minor concern because the weights in the generalized linear layers constitute 99.9\% of the trainable parameters (see \Cref{tab:applicable to linear layer}). 

Implementation-wise, our codebase is automatic (allowing any model to be DP optimized) and future-proof (allowing any training setting, including the user-level DP and the distributed learning). However, although BK is theoretically as fast as the standard training for small $T$, we observe some gap between the theoretical complexity and the hardware throughput in practice. This gap is mainly due to the mechanism of Pytorch hooks which can be possibly optimized by customizing the CUDA kernel or using the symbolic programming. We expect this gap to be closed by future research.

\bibliography{reference}
\bibliographystyle{icml2023}

\clearpage
\onecolumn
\appendix
\section{Background}
\label{app:prelim}

\subsection{Differential privacy}
We formally introduce the differential privacy (DP).
\begin{definition}[\cite{dwork2006calibrating}]
A randomized algorithm $M$ is $ (\varepsilon, \delta)$-differentially private (DP) if for any two neighboring\footnote{$S^\prime$ is a neighbor of $S$ if one can obtain $S^\prime$ by adding or removing one data point from $S$.} datasets $S,S^{\prime}$, and for any event $E$,
\begin{align}
 \mathbb{P}[M(S) \in E] \leqslant \mathrm{e}^{\varepsilon} \mathbb{P}\left[M\left(S^{\prime}\right) \in E\right]+\delta.
\end{align}
\end{definition}

Clearly, stronger DP (smaller $\epsilon,\delta$) indicates the higher difficulty for privacy attackers to extract information from the training data.

DP can be achieved by adding Gaussian noise to a bounded-sensitivity function (see Theorem A.1 of \cite{dwork2014algorithmic}). In deep learning, this function is the sum of per-sample gradients $\sum \g_i$ and the bounded sensitivity is $R$ (that is guaranteed through the gradient clipping after which the per-sample gradient norm is at most $R$). Note that the Gaussian noise magnitude is proportional to the sensitivity: $\sigma_\text{DP}=\sigma R$ in \Cref{eq:DP optimizers}, and $\epsilon(\delta)$ only depends on $\sigma$, not on $R$. The derivation from $(\sigma,T,p)$ in \Cref{alg:dpsgd-ghostBK} to $\epsilon$ can be done through various methods in \Cref{sec:DP prelim}.

\subsection{Computation graph}
We elaborate on the computation graph presented in \Cref{fig:MLP forward backward}. For DP and the standard training, the forward pass is the same: we pass through the layers
$$\a_{(1)}\to\s_{(1)}\to\a_{(2)}\to\s_{(2)}\to\cdots\a_{(L)}\to\s_{(L)}$$

For the backward propagation, there are two sub-processes: 
\begin{enumerate}
    \item the computation of \textbf{output gradient} \textit{for all} layers,
$$\frac{\partial \mathcal{L}}{\partial \s_{(1)}}\leftarrow\cdots\leftarrow\frac{\partial \mathcal{L}}{\partial \s_{(l)}}=\frac{\partial \mathcal{L}}{\partial \s_{(l+1)}}\W_{(l+1)}\circ \text{ReLU}'(\s_{(l)})\leftarrow\cdots\leftarrow\frac{\partial \mathcal{L}}{\partial \s_{(L)}},$$
i.e. the output gradient meets with the weight $\W$; 
\item the computation of \textbf{parameter gradient} \textit{only for trainable} parameters,
$$\frac{\partial \mathcal{L}}{\partial \W_{(l)}}=\frac{\partial \mathcal{L}}{\partial \s_{(l)}}^\top\frac{\partial \s_{(l)}}{\partial \W_{(l)}}=\frac{\partial \mathcal{L}}{\partial \s_{(l)}}^\top\a_{(l)},$$
i.e. the output gradient meets with the activation tensor $\a$.
\end{enumerate}

Note that foward pass, output gradient, and parameter gradient have the same time complexity of $2BTM$ ($B$ being the batch size, $T$ being the feature dimension, e.g. the sequence length in texts, and $M$ being the model size).

For example, GhostClip \cite{li2021large} and MixGhostClip \cite{bu2022scalable}, which use one forward pass and double backward propagation, have a time complexity of $10BTM+O(BT^2)$, while the standard training which uses one forward pass and a single backward propagation has a time complexity of $6BTM$.

\section{Complexity analysis for one layer}
\label{app:complexities}

Let us consider a layer without bias term for simplicity:
\begin{align}
\s=\a\W  
\label{eq:linear layer mat multi}
\end{align}
where $\s\in \R^{B\times T\times p}$ is the output or the pre-activation, $\a\in \R^{B\times T\times d}$ is the input or the post-activation of previous layer, and $\W\in\R^{d\times p}$ is the weight matrix. In a linear layer, $d$ is the input dimension of the hidden feature, $p$ is the output dimension of the hidden feature, and $T$ is the sequence length (or 1 if the data are non-sequential). In a convolution layer, $d$ is the product of the input channels and kernel sizes, $p$ is the output channels, $T$ is the height times width of the hidden representation.

We now break down the time and space complexities for each operation in the training. Notice that we focus on major complexities, e.g. ignoring cubic terms like $BTp$ when higher order terms like $BTpd$ or $BT^2p$ exist.

\subsection{Forward pass}
The complexity of forward pass is incurred by the standard matrix multiplication $\s=\a\W$. Since
$\a\in\R^{B\times T\times d}$ and $\W\in\R^{d\times p}$, the time complexity is $2BTpd$ and the space complexity is $BTp+pd$.

\subsection{Back-propagation: output gradient}
The complexity to compute the output gradient is incurred by the chain rule: for a single sample,
$$\frac{\partial \mathcal{L}}{\partial\s_{(l-1),i}}=\underbrace{\frac{\partial \mathcal{L}}{\partial\s_{(l),i}}}_{\R^{T\times p}}\underbrace{\W_{(l)}^\top}_{\R^{p\times d}}\circ \underbrace{\phi'(\s_{(l-1),i})}_{\R^{T\times d}}$$
where $\phi$ is the non-linear activation function. We compute the matrix multiplication $\frac{\partial \mathcal{L}}{\partial\s_{(l),i}}\W_{(l)}$ first, with time complexity $2BTpd$ and space complexity $pd+BTd+BTp$. Then the elementwise product uses time complexity $2BTd$ and space complexity $BTd$.

\subsection{Back-propagation: parameter gradient}
This module could represent different operations in different DP implementations. In the first back-propagation of GhostClip and the only back-propagation of Opacus, it computes $\frac{\partial\mathcal{L}}{\partial\W}=\frac{\partial\sum_i\mathcal{L}_i}{\partial\W}$; in the second back-propagation of Ghost/FastGradClip/BK, it computes the clipped gradient $\frac{\partial\sum_i C_i\mathcal{L}_i}{\partial\W}$. Regardless of the cases, the operation always takes the same format as
$$\frac{\partial\mathcal{L}}{\partial\W}={\underbrace{\a}_{\R^{B\times T\times d}}}^\top\underbrace{\frac{\partial\mathcal{L}}{\partial\s}}_{\R^{B\times T\times p}}.$$
In contrast to the per-sample gradient instantiation, this operation is a tensor multiplication instead of many matrix multiplication, and the output is a single pair of gradient $\R^{d\times p}$ instead of many per-sample gradients.

This tensor multiplication has time complexity $2BTpd$ and space complexity $pd$ unless all per-sample gradients are stored.

\subsection{Ghost norm}
Ghost norm is the operation taking $\a_i$ and $\frac{\partial\mathcal{L}}{\partial\s_i}$ as the input and outputs the per-sample gradient norm. According to \Cref{eq:ghost norm} and Appendix C.3 of \cite{bu2022automatic}, this operation computes $\a_i\a_i^\top$ and $\frac{\partial\mathcal{L}}{\partial \s_i}\frac{\partial\mathcal{L}}{\partial \s_i}^\top$, taking the time complexity $2BT^2d$ and $2BT^2p$ respectively, and the space complexity $BT^2$ for each variable. Hence total time complexity is $2BT^2(p+d)$ and total space complexity is $2BT^2$.

\subsubsection{Per-sample gradient instantiation}
Alternatively, one can instantiate the per-sample gradients and then compute their norms. This is different than the computation of parameter gradient in the back-propagation: that computation is an efficient tensor multiplication while this operation consists of $B$ matrix multiplication.
$$\frac{\partial\mathcal{L}_i}{\partial\W}=\underbrace{\a_i}_{\R^{T\times d}}\underbrace{\frac{\partial\mathcal{L}}{\partial\s_i}^\top}_{\R^{T\times p}} \text{ for }i\in [B].$$
This operation has time complexity $2BTpd$ and space complexity $Bpd$ to store all per-sample gradients. Computing the norms is cheap enough to be neglected.

\subsection{Weighted sum of per-sample gradient}
This operation simply takes per-sample clipping factor $C_i\in\R$ and $\frac{\partial\mathcal{L}_i}{\partial\W}\in\R^{B\times d\times p}$ as the input, and outputs the clipped gradient $\R^{d\times p}$ as a weighted sum $\sum_i C_i\frac{\partial\mathcal{L}_i}{\partial\W}$. The time complexity is $2Bpd$ and the space complexity is 0 since the summation happens in place.

In contrast to double back-propagation, which indirectly derives the clipped gradient by differentiating the  reweighted loss $\sum_i C_i\mathcal{L}_i$ at a cost of $O(BTpd)$, this operation directly computes the clipped gradient under almost no time complexity. Noticeably, this is only possible if per-sample gradients are readily instantiated and stored.

\section{Line-by-line comparison between different implementations}
\label{app:line by line implementations}

\subsection{BK v.s. GhostClip}
\begin{algorithm}[H]
\label{alg:BK v.s. GhostClip}
	\caption{DP optimizer with \colorbox{pink}{BK} or \colorbox{lime}{GhostClip}}
\textbf{ Parameters:} $l$-th layer weights $\W_{(l)}$, number of layers $L$, noise level $\sigma$.
\begin{algorithmic}[1]
\State{\textcolor{gray}{\textit{\#  forward pass}}}
\For{layer $l\in 1,2,\cdots,L$}
\State Get $\{\a_{(l),i}\}$
\EndFor

\State{\textcolor{gray}{\textit{\#  backward propagation with loss $\mathcal{L}=\sum_i\mathcal{L}_i$}}}
\For{layer $l\in L,L-1,\cdots,1$}
\State Get output gradient $\{\frac{\partial \mathcal{L}}{\partial \s_{(l),i}}\}$
\State
Compute per-sample gradient norm: 
$\|\frac{\partial \mathcal{L}_i}{\partial\W_{(l)}}\|_F^2=\text{vec}(\frac{\partial \mathcal{L}}{\partial \s_{(l),i}}^\top \frac{\partial \mathcal{L}}{\partial \s_{(l),i}})\cdot\text{vec}(\a_{(l),i}^\top\a_{(l),i})$
\State\colorbox{lime}{Compute non-private gradient: $\frac{\partial \mathcal{L}}{\partial \W_{(l)}}=\a_{(l)}^\top\frac{\partial \mathcal{L}}{\partial \s_{(l)}}$}
\EndFor

\State Aggregate gradient norm across all layers: $\|\frac{\partial \mathcal{L}_i}{\partial\W}\|_F^2=\sum_l\|\frac{\partial \mathcal{L}_i}{\partial\W_{(l)}}\|_F^2$
\State Compute clipping factor: $C_{i}=C(\|\frac{\partial \mathcal{L}_i}{\partial\W}\|_F;R)$

\For{layer $l\in L,L-1,\cdots,1$}
\State \colorbox{pink}{Compute sum of clipped gradients $\G_l=\a_{(l)}^\top\text{diag}(\C)\frac{\partial \mathcal{L}}{\partial \s_{(l)}}$}
\State{\colorbox{lime}{\textcolor{gray}{\textit{\# 2nd backward propagation with loss $\mathcal{L}=\sum_i C_i\mathcal{L}_i$}}}}
\State \colorbox{lime}{Get output gradient $\{\frac{\partial \sum C_i\mathcal{L}_i}{\partial \s_{(l),i}}\}$}
\State \colorbox{lime}{Compute sum of clipped gradients $\G_l=\a_{(l)}^\top\frac{\partial \sum C_i\mathcal{L}_i}{\partial \s_{(l)}}$}
\State Delete $\{\a_{(l),i}\},$\colorbox{pink}{$\{\frac{\partial \mathcal{L}}{\partial \s_{(l),i}}\}$},\colorbox{lime}{$\{\frac{\partial \sum C_i\mathcal{L}_i}{\partial \s_{(l),i}}\}$}
\EndFor
\State Add Gaussian noise $\hat\G=\G+\sigma R\cdot \mathcal{N}(0, \I)$
\State Apply SGD/Adam/LAMB with the private gradient $\hat\G$ on $\W$
\end{algorithmic}
\end{algorithm}

\clearpage
\subsection{BK v.s. Opacus}
\scalebox{0.95}{
\begin{minipage}{\linewidth}
\begin{algorithm}[H]
\caption{DP optimizer with \colorbox{pink}{BK} or \colorbox{lime}{Opacus}}
	\label{alg:BKAL-opacus}
    \textbf{Parameters:} $l$-th layer's weights $\W_{(l),t}$, number of layers $L$, noise scale $\sigma$.
\begin{algorithmic}[1]
\For{layer $l\in 1,2,\cdots,L$}
\State Get $\{\a_{(l),i}\}$
\EndFor

\For{layer $l\in L,L-1,\cdots,1$}
\State Get output gradient $\{\frac{\partial \mathcal{L}}{\partial \s_{(l),i}}\}$
\State \colorbox{pink}{Compute per-sample gradient norm: 
$\|\frac{\partial \mathcal{L}_i}{\partial\W_{(l)}}\|_F^2=\text{vec}(\frac{\partial \mathcal{L}}{\partial \s_{(l),i}}^\top \frac{\partial \mathcal{L}}{\partial \s_{(l),i}})\cdot\text{vec}(\a_{(l),i}^\top\a_{(l),i})$}
\State\colorbox{lime}{Compute non-private gradient: $\frac{\partial \mathcal{L}}{\partial \W_{(l)}}=\a_{(l)}^\top\frac{\partial \mathcal{L}}{\partial \s_{(l)}}$}
\State\colorbox{lime}{Compute per-sample gradients: $\frac{\partial \mathcal{L}_i}{\partial \W_{(l)}}=\a_{(l),i}^\top\frac{\partial \mathcal{L}}{\partial \s_{(l),i}}$ and gradient norms
$\|\frac{\partial \mathcal{L}_i}{\partial\W_{(l)}}\|_F^2$}
\State \colorbox{lime}{Delete $\{\a_{(l),i}\},\{\frac{\partial \mathcal{L}}{\partial \s_{(l),i}}\}$}
\EndFor

\State Aggregate gradient norm across all layers: $\|\frac{\partial \mathcal{L}_i}{\partial\W}\|_F^2=\sum_l\|\frac{\partial \mathcal{L}_i}{\partial\W_{(l)}}\|_F^2$
\State Compute clipping factor: $C_{i}=C(\|\frac{\partial \mathcal{L}_i}{\partial\W}\|_F;R)$

\For{layer $l\in L,L-1,\cdots,1$}
\State \colorbox{pink}{Compute sum of clipped gradients $\G_l=\a_{(l)}^\top\text{diag}(\C)\frac{\partial \mathcal{L}}{\partial \s_{(l)}}$}
\State \colorbox{lime}{Compute sum of clipped gradients $\G_l=\sum C_i\frac{\partial \mathcal{L}_i}{\partial \W_{(l)}}$}
\State \colorbox{pink}{Delete $\{\a_{(l),i}\},\{\frac{\partial \mathcal{L}}{\partial \s_{(l),i}}\}$},\colorbox{lime}{$\{\frac{\partial \mathcal{L}}{\partial \W_{(l)}}\}$}
\EndFor
\State Add Gaussian noise $\hat\G=\G+\sigma R\cdot \mathcal{N}(0, \I)$
\State Apply SGD/Adam/LAMB with the private gradient $\hat\G$ on $\W$
\end{algorithmic}
\end{algorithm}
\end{minipage}
}

\subsection{BK v.s. standard (non-DP)}
\scalebox{0.95}{
\begin{minipage}{\linewidth}
\begin{algorithm}[H]
	\caption{DP optimizer with \colorbox{pink}{BK} or \colorbox{lime}{Standard} optimizer}
	\label{alg:BKAL-opacus}
    \textbf{Parameters:} $l$-th layer's weights $\W_{(l),t}$, number of layers $L$, noise scale $\sigma$.
\begin{algorithmic}[1]
\For{layer $l\in 1,2,\cdots,L$}
\State Get $\{\a_{(l),i}\}$
\EndFor

\For{layer $l\in L,L-1,\cdots,1$}
\State Get output gradient $\{\frac{\partial \mathcal{L}}{\partial \s_{(l),i}}\}$
\State \colorbox{pink}{Compute per-sample gradient norm: 
$\|\frac{\partial \mathcal{L}_i}{\partial\W_{(l)}}\|_F^2=\text{vec}(\frac{\partial \mathcal{L}}{\partial \s_{(l),i}}^\top \frac{\partial \mathcal{L}}{\partial \s_{(l),i}})\cdot\text{vec}(\a_{(l),i}^\top\a_{(l),i})$}
\State\colorbox{lime}{Compute non-private gradient: $\frac{\partial \mathcal{L}}{\partial \W_{(l)}}=\a_{(l)}^\top\frac{\partial \mathcal{L}}{\partial \s_{(l)}}$}
\State \colorbox{lime}{Delete $\{\a_{(l),i}\},\{\frac{\partial \mathcal{L}}{\partial \s_{(l),i}}\}$}
\EndFor

\State \colorbox{pink}{Aggregate gradient norm across all layers: $\|\frac{\partial \mathcal{L}_i}{\partial\W}\|_F^2=\sum_l\|\frac{\partial \mathcal{L}_i}{\partial\W_{(l)}}\|_F^2$}
\State \colorbox{pink}{Compute clipping factor: $C_{i}=C(\|\frac{\partial \mathcal{L}_i}{\partial\W}\|_F;R)$}

\For{layer $l\in L,L-1,\cdots,1$}
\State \colorbox{pink}{Compute sum of clipped gradients $\G_l=\a_{(l)}^\top\text{diag}(\C)\frac{\partial \mathcal{L}}{\partial \s_{(l)}}$}
\State \colorbox{pink}{Delete $\{\a_{(l),i}\},\{\frac{\partial \mathcal{L}}{\partial \s_{(l),i}}\}$}
\EndFor
\State \colorbox{pink}{Add Gaussian noise $\hat\G=\G+\sigma R\cdot \mathcal{N}(0, \I)$}
\State Apply SGD/Adam/LAMB with \colorbox{pink}{$\hat\G$} or \colorbox{lime}{$\G$} on $\W$
\end{algorithmic}
\end{algorithm}
\end{minipage}
}

\subsection{BK (base) v.s. hybrid BK}
\begin{minipage}{\textwidth}
\begin{algorithm}[H]
\caption{DP optimizer with BK, BK-\colorbox{lime}{MixGhostClip} or BK-\colorbox{cyan}{MixOpt}}\label{alg:BKAL-mixed}
\textbf{Parameters:} $l$-th layer's weights $\W_{(l)}$, number of layers $L$, noise scale $\sigma$.
\begin{algorithmic}[1]
\State{\textcolor{gray}{\textit{\#  forward pass}}}
\For{layer $l\in 1,2,\cdots,L$}
\State Get $\{\a_{(l),i}\}$
\EndFor

\State{\textcolor{gray}{\textit{\#  backward propagation with loss $\mathcal{L}=\sum_i\mathcal{L}_i$}}}
\For{layer $l\in L,L-1,\cdots,1$}
\State Get output gradient $\{\frac{\partial \mathcal{L}}{\partial \s_{(l),i}}\}$
\If{(\colorbox{lime}{MixGhostClip} or \colorbox{cyan}{MixOpt}) and $2T_{(l)}^2>p_{(l)}d_{(l)}$}
\State {Compute per-sample gradients: $\frac{\partial \mathcal{L}_i}{\partial \W_{(l)}}=\a_{(l),i}^\top\frac{\partial \mathcal{L}}{\partial \s_{(l),i}}$ and gradient norms $\|\frac{\partial \mathcal{L}_i}{\partial\W_{(l)}}\|_F^2$}
\Else
\State {Compute per-sample gradient norm: 
$\|\frac{\partial \mathcal{L}_i}{\partial\W_{(l)}}\|_F^2=\text{vec}(\frac{\partial \mathcal{L}}{\partial \s_{(l),i}}^\top \frac{\partial \mathcal{L}}{\partial \s_{(l),i}})\cdot\text{vec}(\a_{(l),i}^\top\a_{(l),i})$}
\EndIf
\EndFor

\State Aggregate gradient norm across all layers: $\|\frac{\partial \mathcal{L}_i}{\partial\W}\|_F^2=\sum_l\|\frac{\partial \mathcal{L}_i}{\partial\W_{(l)}}\|_F^2$
\State Compute clipping factor: $C_{i}=C(\|\frac{\partial \mathcal{L}_i}{\partial\W}\|_F;R)$

\For{layer $l\in L,L-1,\cdots,1$}
\If{\colorbox{cyan}{MixOpt} and $2T_{(l)}^2>p_{(l)}d_{(l)}$}
\State Compute weighted gradients $\G_l=\sum C_i\frac{\partial \mathcal{L}_i}{\partial \W_{(l)}}$
\Else
\State {Compute sum of clipped gradients $\G_l=\a_{(l)}^\top\text{diag}(\C)\frac{\partial \mathcal{L}}{\partial \s_{(l)}}$}
\EndIf
\State {Delete $\{\a_{(l),i}\},\{\frac{\partial \mathcal{L}}{\partial \s_{(l),i}}\}$},\colorbox{cyan}{$\{\frac{\partial \mathcal{L}_i}{\partial \W_{(l)}}\}$}
\EndFor
\State Add Gaussian noise $\hat\G=\G+\sigma R\cdot \mathcal{N}(0, \I)$
\State Apply SGD/Adam/LAMB with the private gradient $\hat\G$ on $\W$
\end{algorithmic}
\end{algorithm}
\end{minipage}

\section{Codebase README}
Here we describe some designs in our codebase for BK algorithms.
\subsection{Supported layers}
\begin{itemize}
    \item Linear: Ghost norm or per-sample gradient instantiation
    \item Embedding: Ghost norm
    \item Conv1d \& Conv2d \& Conv3d: Ghost or per-sample gradient instantiation
    \item GroupNorm \& LayerNorm \& InstanceNorm: per-sample gradient instantiation
\end{itemize}

\subsection{Instruction of implementation}
\label{app:auto-differentiation}
In this section, we will discuss the specific designs and tricks for our book-keeping technique. We illustrate through Pytorch automatic differentiation package, known as \texttt{torch.autograd} or simply \texttt{autograd}\footnote{See \url{https://pytorch.org/docs/stable/autograd.html} for an official introduction.}. It has two high-level operators, \texttt{autograd.backward} (which is the major component of the commonly used \texttt{loss.backward()}) and \texttt{autograd.grad}. We denote the model parameters as \texttt{param}.

On all trainable layers, i.e. layers with at least one trainable parameter such that \texttt{param.requires\_grad=True}, the operator \texttt{autograd.backward} does three things, 1. compute the output gradient $\frac{\partial\mathcal{L}}{\partial \s}$ for this layer; 2. compute the parameter gradient $\frac{\partial\mathcal{L}}{\partial \W}$ or $\frac{\partial\mathcal{L}}{\partial \b}$; 3. store the parameter gradient to \texttt{param.grad} attribute.

In contrast, \texttt{autograd.grad} returns but does not store the parameter gradient in step 3. However, \texttt{autograd.grad} still computes the parameter gradient in step 2 (or \circled{\small{2b}}) unnecessarily.

Therefore the key idea is to only compute the output gradient without computing the parameter gradient. This goal can be achieved by 
\begin{enumerate}
    \item registering the Pytorch backward hooks, which have free access to the output gradient $\frac{\partial\mathcal{L}}{\partial \s}$, to store this output gradient for \circled{2a} (Line 9 of \Cref{alg:dpsgd-ghostBK});
    \item setting all parameters to not require gradients, through \texttt{requires\_grad=False}.
\end{enumerate} 

\subsection{Work-around: origin parameters}
Unfortunately, the back-propagation will not be executed if all parameters are set to not require gradients, since the computation graph needs to be created at least on some trainable parameters. Therefore, while the above methodology is certainly implementable through mild modification on the low level (like CUDA kernel), we provide a lightweight work-around in Pytorch.

\begin{figure}[!htb]
    \centering
    \includegraphics[width=0.8\linewidth]{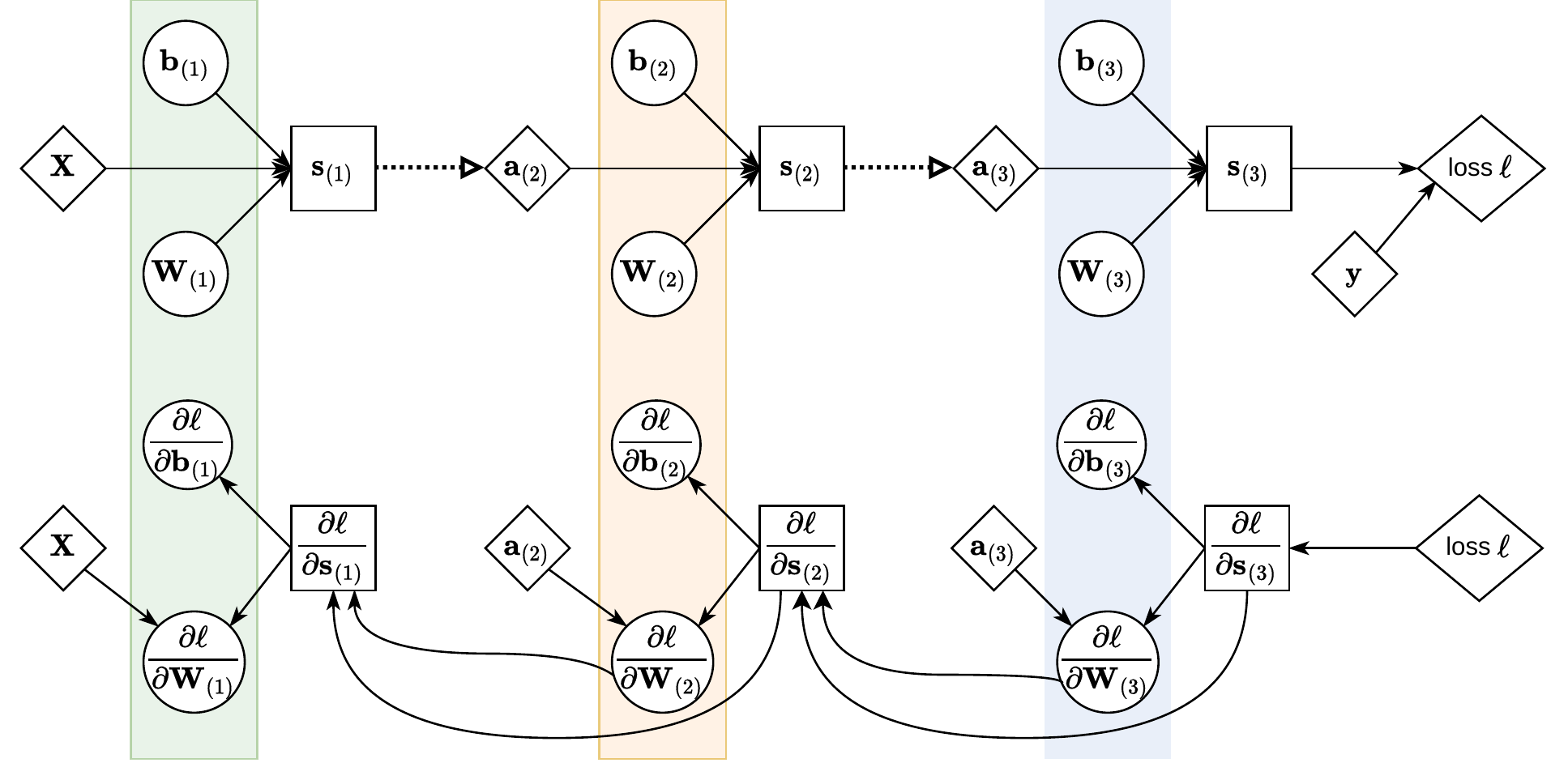}
    \caption{Forward pass (upper panel) and back-propagation (lower panel) of a 3-layer MLP.}
    \label{fig:MLP nodes}
\end{figure}

To make sure that the back-propagation indeed propagates through all trainable parameters, we set \texttt{param.requires\_grad=True} on and only on the ancestor parameter nodes of all output nodes, termed as the \textbf{origin parameters}. Specifically, we define the origin parameters as the \textit{subset of parameter nodes, whose descendant nodes cover all the output nodes}. This is visualized in \Cref{fig:MLP nodes} for a 3-layer MLP, using the same symbols as \Cref{fig:MLP forward backward}.

Here, $\s_{(i)}$ are the output nodes (in squares) from the trainable layers. The ancestor parameter nodes (in circles) of $\s_{(3)}$ are $\{\b_{(3)},\b_{(2)},\b_{(1)},\W_{(3)},\W_{(2)},\W_{(1)}\}$, those of $\s_{(2)}$ are $\{\b_{(2)},\b_{(1)},\W_{(2)},\W_{(1)}\}$, and those of $\s_{(1)}$ are $\{\b_{(1)},\W_{(1)}\}$. Therefore, subsets including but not limited to $\{\b_{(3)},\b_{(2)},\b_{(1)},\W_{(3)},\W_{(2)},\W_{(1)}\}$, $\{\b_{(1)},\W_{(1)}\}$, and $\{\b_{(1)}\}$ are qualified as the origin parameters, since their descendants cover all output nodes. In fact, the smallest subsets are $\{\b_{(1)}\}$ or $\{\W_{(1)}\}$, and either can serve as the optimal origin parameters. 

\begin{remark}
The origin parameters are usually within the embedding layer in language models and transformers, or within the first convolution layer in vision models. Since the origin parameters only constitute a small fraction of all trainable parameters (fewer than the parameters in the first layer) in deep neural networks (with hundreds of layers), the computational overhead wasted on the regular gradient of origin parameters is negligible.
\end{remark}

\begin{remark}
Since we will waste the computation of regular gradient $\frac{\partial\mathcal{L}}{\partial \texttt{origin\_parameters}}$, it is preferred to use the bias over the weight for minimum waste whenever possible. We note that sometimes the first layer contains no bias term. For example, the embedding layer by \texttt{torch.nn.Embedding} has no bias by design, and so do all convolution layers in ResNets from torchvision \cite{marcel2010torchvision}, with reasons discussed at Section 3.2 of \cite{ioffe2015batch}, which generalizes to all batch-normalized CNN if the normalization is applied before the activation function.
\end{remark}

In summary, we drive the back-propagation without computing the regular parameter gradient $\frac{\partial\sum_i\mathcal{L}_i}{\partial \W}$ (by setting \texttt{param.requires\_grad=False}), and use Pytorch backward hooks to access and store the output gradient $\frac{\partial\mathcal{L}}{\partial \s}$.

\begin{table}[!htb]
    \centering
    \resizebox{\linewidth}{!}{
    \begin{tabular}{c|c|c|c|c|c}
    &\multicolumn{2}{c|}{non-DP training} &\multicolumn{3}{|c}{DP training (Book-Keeping)}  \\\hline
    &trainable&non-trainable&trainable param&trainable param&non-trainable\\
    &param&param&(origin param)&(not origin param)&param\\\hline
    register hook&\xmark&\xmark&\cmark&\cmark&\xmark\\
    \texttt{param.requires\_grad}&\cmark&\xmark&\cmark&\xmark&\xmark
    \end{tabular}
    }
    \caption{Origin parameter trick and implementation details.}
\end{table}

\subsection{How to use BK codebase}
With a few lines of code, it is easy to use our BK codebase to change the standard training to the DP training. All you need to do is to declare a privacy engine and attach it to the optimizer.

\begin{verbatim}
from fastDP import PrivacyEngine
from transformers import AutoModel

model = AutoModel.from_pretrained('roberta-base')

optimizer = torch.optim.Adam(params=model.parameters())

privacy_engine = PrivacyEngine(
    model,batch_size=256,sample_size=50000,
    epochs=3,target_epsilon=3,clipping_mode='MixOpt')
privacy_engine.attach(optimizer)

# Same training procedure, e.g. data loading, forward pass, logits...
loss = torch.nn.functional.cross_entropy(logits, labels)
loss.backward()
optimizer.step()
optimizer.zero_grad()
\end{verbatim}

Notice that if \texttt{clipping\_mode} is set to default, then BK (base) is implemented; if \texttt{clipping\_mode=='MixGhostClip'}, then BK-MixGhostClip is implemented; if \texttt{clipping\_mode=='MixOpt'}, then BK-MixOpt is implemented.

We also allow the gradient accumulation in the same way as non-private training.

\section{Applicability of BK algorithm}
\subsection{Applying BK to full fine-tuning}
We experiment with numerous vision and language models to show the strong applicability of BK. Notice that the ghost norm trick only applies on weight parameters and in the generalized linear layers, i.e. embedding/convolutional/linear. The vision models are imported from Pytorch Image Models library \cite{rw2019timm} and the language models are imported from Hugging Face Transformers library \cite{wolf-etal-2020-transformers}\footnote{In Transformers library, layers with class name \href{https://huggingface.co/transformers/v3.1.0/internal/modeling_utils.html\#transformers.modeling_utils.Conv1D}{`Conv1D'} is actually a linear layer, different from 1D convolution \texttt{torch.nn.Conv1d}.}.

\begin{table}[!htb]
    \centering
    \resizebox{0.9\linewidth}{!}{
    \begin{tabular}{c|c|c|c|c}
         \multirow{3}{*}{Model}& \multicolumn{2}{|c|}{\multirow{2}{*}{\shortstack{\# param in\\ generalized linear layers}}}&\multirow{2}{*}{\# param in other layers}& \multirow{3}{*}{\% applicable to BK}
         \\
         &&&&
         \\
         &weight&bias&weight+bias&
         \\\hline\hline
         ResNet18&11.7M&1000&9600&99.9\%
         \\\hline
         ResNet34&21.8M&1000&17024&99.9\%
         \\\hline
         ResNet50&25.5M&1000&53120&99.8\%
         \\\hline
         ResNet101&44.4M&1000&105344&99.8\%
         \\\hline
         ResNet152&60.2M&1000&151424&99.7\%
         \\\hline
         DenseNet121&7.9M&1000&83648&98.9\%
         \\\hline
         DenseNet161&28.5M&1000&219936&99.2\%
         \\\hline
         DenseNet201&19.8M&1000&229056&98.9\%
         \\\hline
         Wide ResNet50&68.8M&1000&68224&99.9\%
         \\\hline
         Wide ResNet101&126.7M&1000&137856&99.9\%
         \\\hline
         vit\_tiny\_patch16\_224&5.6M&21928&9600&99.4\%
         \\\hline
vit\_small\_patch16\_224&21.9M&42856&19200&99.7\%
         \\\hline
vit\_base\_patch16\_224&86.3M&84712&38400&99.9\%
         \\\hline
vit\_large\_patch16\_224&303.8M&223208&100352&99.9\%
         \\\hline
crossvit\_tiny\_240&6.9M&30800&16128&99.3\%
         \\\hline
crossvit\_small\_240&26.6M&59600&32256&99.7\%
         \\\hline
crossvit\_base\_240&104.5M&117200&64512&99.8\%
         \\\hline
convnext\_small&50.1M&83656&30144&99.8\%
         \\\hline
convnext\_base&88.4M&111208&40192&99.8\%
         \\\hline
convnext\_large&197.5M&166312&60288&99.9\%
         \\\hline
deit\_tiny\_patch16\_224&5.6M&21928&9600&99.4\%
         \\\hline
deit\_small\_patch16\_224&21.9M&42856&19200&99.7\%
         \\\hline
deit\_base\_patch16\_224&86.3M&84712&38400&99.9\%
         \\\hline
beit\_base\_patch16\_224&86.3M&57064&38400&99.9\%
         \\\hline
beit\_large\_patch16\_224&303.8M&149480&100352&99.9\%
         \\\hline\hline
roberta-base&124.5M&83712&38400&99.9\%
         \\\hline
roberta-large&355.0M&222208&100352&99.9\%
         \\\hline
distilroberta-base&82.1M&42240&19968&99.9\%
         \\\hline
bert-base-uncased&109.4M&83712&38400&99.9\%
         \\\hline
bert-large-uncased&334.8M&222208&100352&99.9\%
         \\\hline
bert-base-cased&108.2M&83712&38400&99.9\%
         \\\hline
bert-large-cased&333.3M&222208&100352&99.9\%
         \\\hline
longformer-base-4096&148.5M&111360&38400&99.9\%
         \\\hline
longformer-large-4096&434.2M&295936&100352&99.9\%
         \\\hline
t5-small&60.5M&0&16384&99.9\%
         \\\hline
t5-base&222.9M&0&47616&99.98\%
         \\\hline
t5-large&737.5M&0&124928&99.98\%
         \\\hline
long-t5-local-base&222.9M&0&47616&99.98\%
         \\\hline
long-t5-local-large&750.1M&0&124928&99.98\%
         \\\hline
long-t5-tglobal-base&222.9M&0&56832&99.97\%
         \\\hline
long-t5-tglobal-large&750.1M&0&149504&99.98\%
         \\\hline
gpt2&124.3M&82944&38400&99.9\%
         \\\hline
gpt2-medium&354.5M&221184&100352&99.9\%
         \\\hline
gpt2-large&773.4M&414720&186880&99.9\%
         \\\hline
    \end{tabular}
    }
    \caption{Models and the percentage of trainable parameters in generalized linear layers.}
    \label{tab:applicable to linear layer}
\end{table}

\subsection{Applying BK to parameter efficient fine-tuning}
\label{app:param eff BK}
We demonstrate that BK (base and hybrid) can be applied to DP LoRA and DP Adapter, where the rank $r$ is usually 16-1024. For the ease of presentation, we describe the BK base, similarly to \Cref{alg:dpsgd-ghostBK}.

\paragraph{Adapter}
An adapter module is injected after a linear layer:
$$A(x)=\tau(xD)U+x$$
where $x\in\R^{B\times T\times p}, D\in\R^{p\times r},U\in\R^{r\times p}$. We decompose the module $A$ into two sub-modules:
\begin{itemize}
    \item $x\to xD:=u$, activation $x$, output grad $\frac{\partial\mathcal{L}}{\partial u}$
    \item $\tau(u)\to \tau U:=v$, activation $\tau(xD)$, output grad $\frac{\partial\mathcal{L}}{\partial v}$
\end{itemize}
Hence BK can be implemented as follows.

\begin{enumerate}
\item Get activation tensors $x$ and $\tau(xD)$ by Pytorch forward hook
\item Get output gradients $\{\frac{\partial \mathcal{L}}{\partial xD}\}$ and $\{\frac{\partial \mathcal{L}}{\partial \tau U}\}$ by Pytorch backward hook
\item Compute per-example gradient norm $\|\frac{\partial \mathcal{L}_i}{\partial D}\|_F^2$ and $\|\frac{\partial \mathcal{L}_i}{\partial U}\|_F^2$ by ghost norm trick
\item Aggregate gradient norm across all layers: $\|\frac{\partial \mathcal{L}_i}{\partial D}\|_F^2+\|\frac{\partial \mathcal{L}_i}{\partial U}\|_F^2$
\item Compute clipping factor $C_{i}$
\item Compute sum of clipped gradients $\G_D=x^\top\text{diag}(C_1,C_2,\cdots)\frac{\partial \mathcal{L}}{\partial xD}$ and $\G_U=\tau^\top\text{diag}(C_1,C_2,\cdots)\frac{\partial \mathcal{L}}{\partial \tau U}$
\item Add Gaussian noise $\hat\G_D=\G_D+\sigma R\cdot \mathcal{N}(0, \I)$ and $\hat\G_U=\G_U+\sigma R\cdot \mathcal{N}(0, \I)$
\item Apply SGD/Adam/LAMB with the private gradient $\hat\G_D$ on $D$ and $\hat\G_U$ on $U$
\end{enumerate}

Existing implementation of DP Adapter \footnote{\url{https://github.com/huseyinatahaninan/Differentially-Private-Fine-tuning-of-Language-Models/tree/main/Language-Understanding-RoBERTa/bert_adapter}} uses the per-sample gradient instantiation as in Opacus. It is not hard to see that the layerwise space overhead (in addition to forward pass and output gradient) is $2Bpr$ and the time overhead is $4BTpr$ (c.f. \Cref{tab:block complexity} \circled{4}). With the BK implementation, the space overhead is $4BT^2$ and the time overhead is $4BT^2(p + r)$ (c.f. \Cref{tab:block complexity} \circled{3}).

\paragraph{LoRA}
LoRA modifies
$$A(x)=x(W+LR)=xW+xLR$$
where $x\in\R^{B\times T\times d}, W\in\R^{d\times p}, L\in\R^{d\times r},R\in\R^{r\times p}$. We decompose the module $A$ into two sub-modules:
\begin{itemize}
    \item $x\to xL:=u$, activation $x$, output grad $\frac{\partial\mathcal{L}}{\partial u}$
    \item $u\to uR:=v$, activation $xL$, output grad $\frac{\partial\mathcal{L}}{\partial v}$
\end{itemize}
Hence BK can be implemented on each sub-module, similar to the DP Adapter.

Existing implementation of DP LoRA \footnote{\url{https://github.com/huseyinatahaninan/Differentially-Private-Fine-tuning-of-Language-Models/tree/main/Language-Understanding-RoBERTa/bert_lora}} uses the per-sample gradient instantiation as in Opacus. It is not hard to see that the layerwise space overhead (in addition to forward pass and output gradient) is $Br(p+d)$ and the time overhead is $2BTr(p+d)$ (c.f. \Cref{tab:block complexity} \circled{4}). With the BK implementation, the space overhead is $4BT^2$ and the time overhead is $2BT^2(p+d+2r)$ (c.f. \Cref{tab:block complexity} \circled{3}).

\clearpage
\section{Additional plots and tables}
\label{app:additional}
\begin{figure}[!htb]
    \centering
    \includegraphics[width=0.4\linewidth]{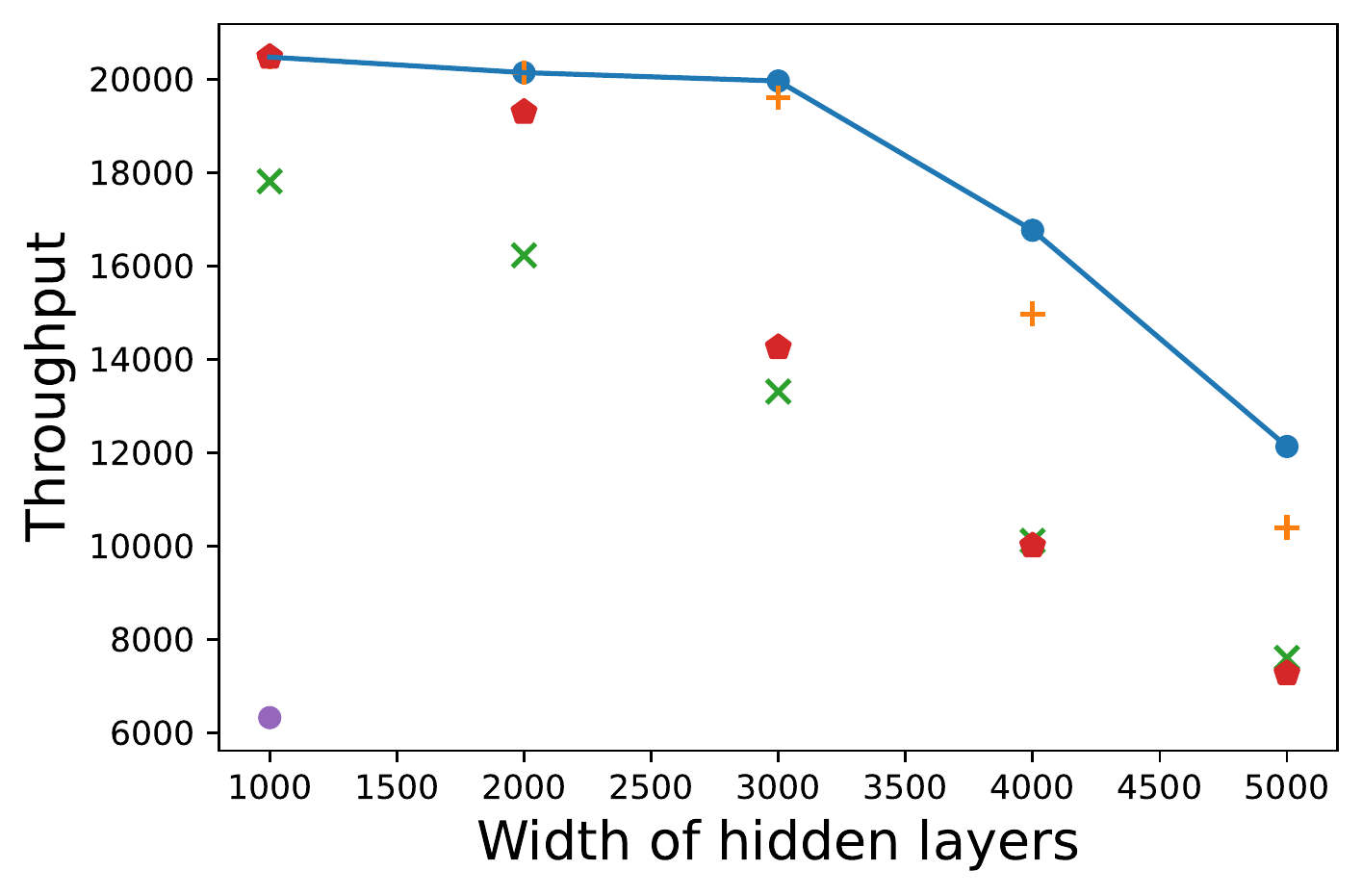}
    \includegraphics[width=0.4\linewidth]{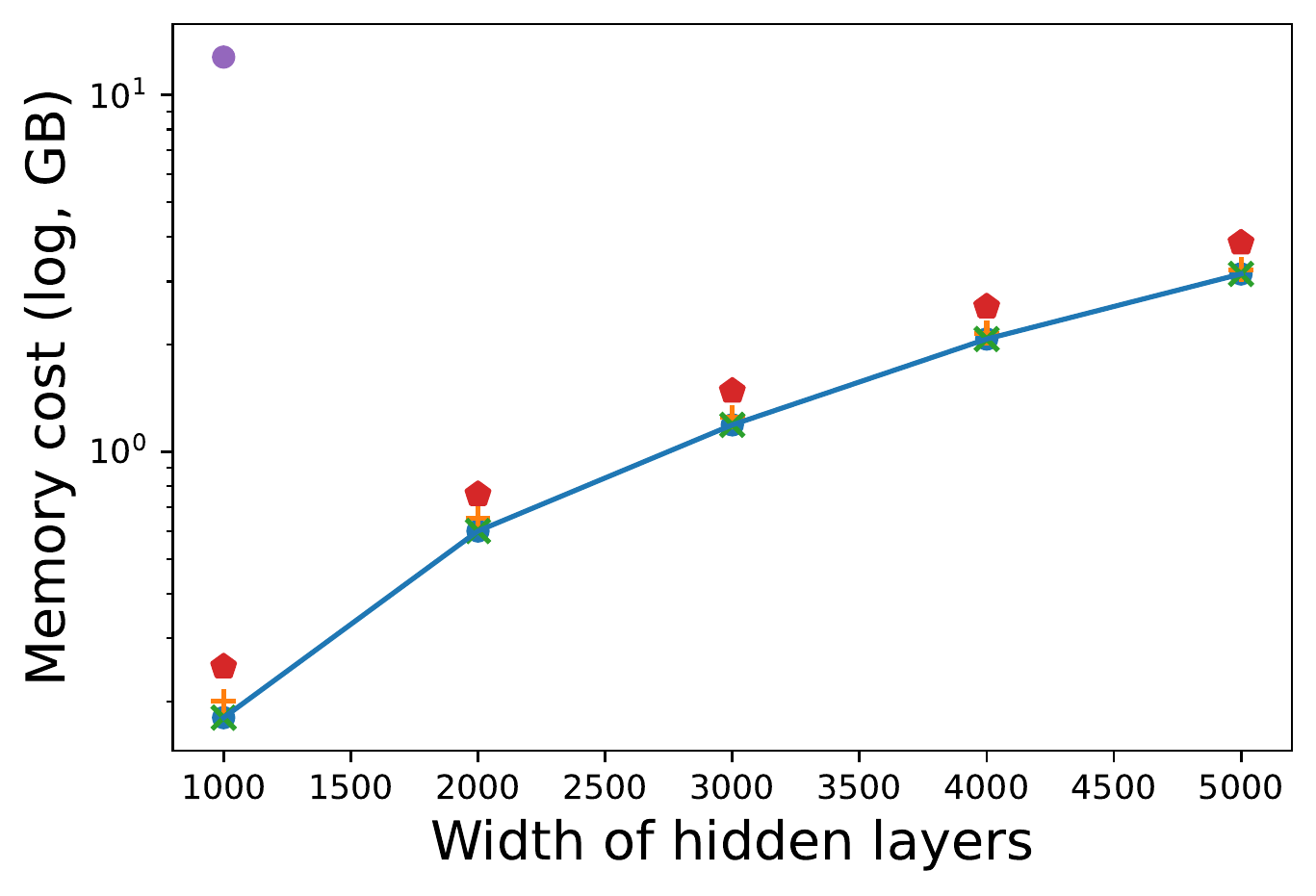}      \includegraphics[width=0.4\linewidth]{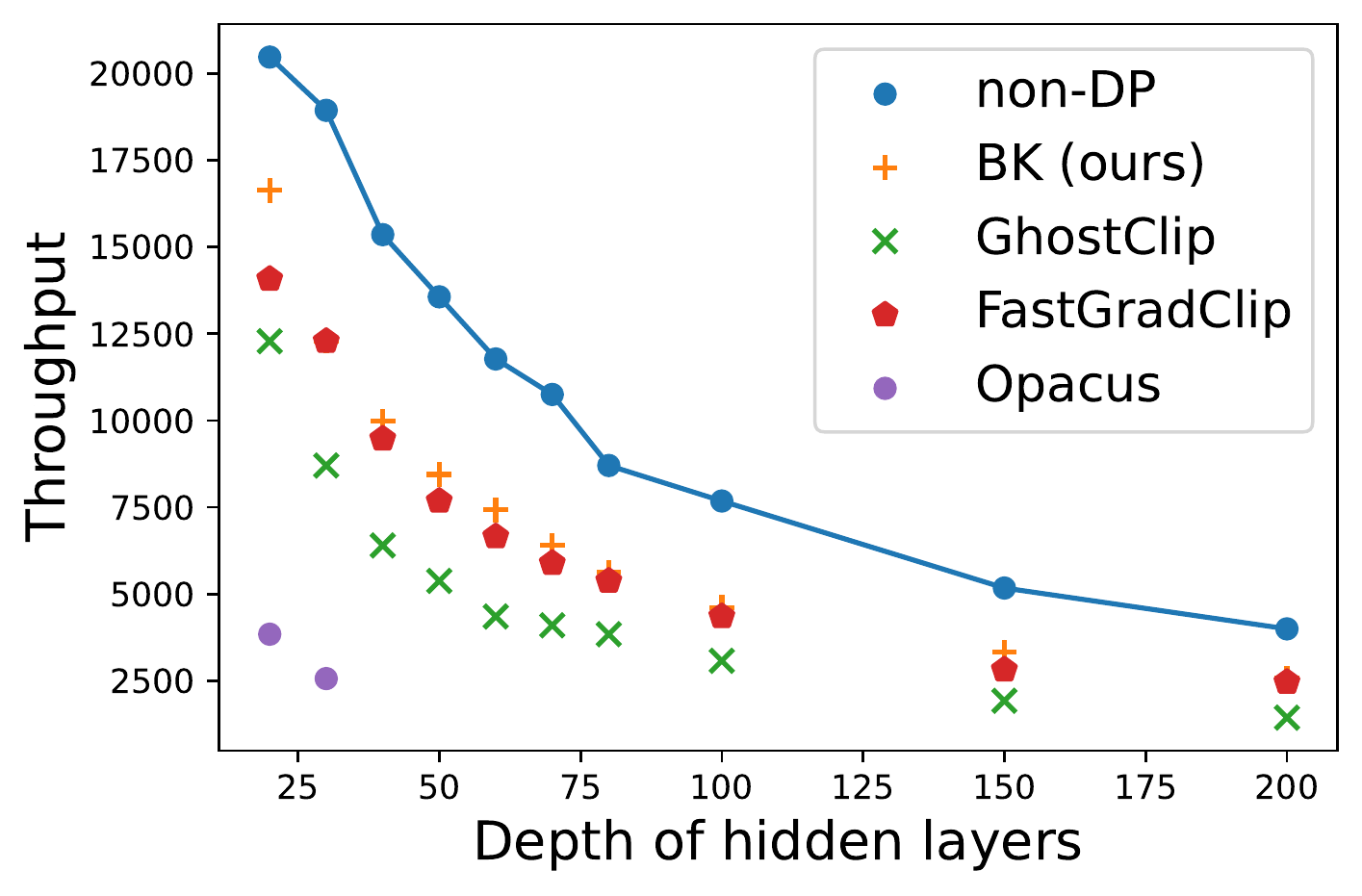}
    \includegraphics[width=0.4\linewidth]{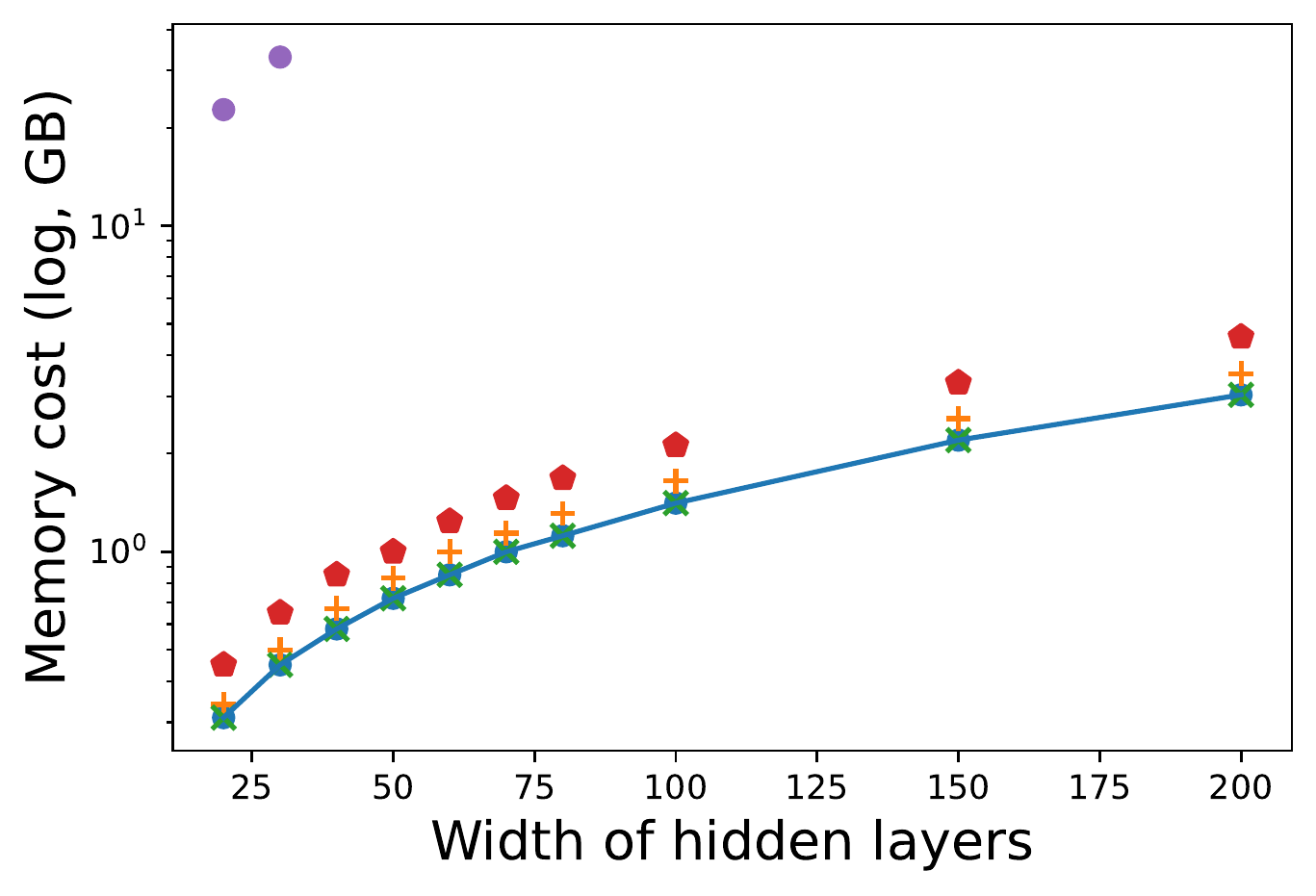}  
        \includegraphics[width=0.4\linewidth]{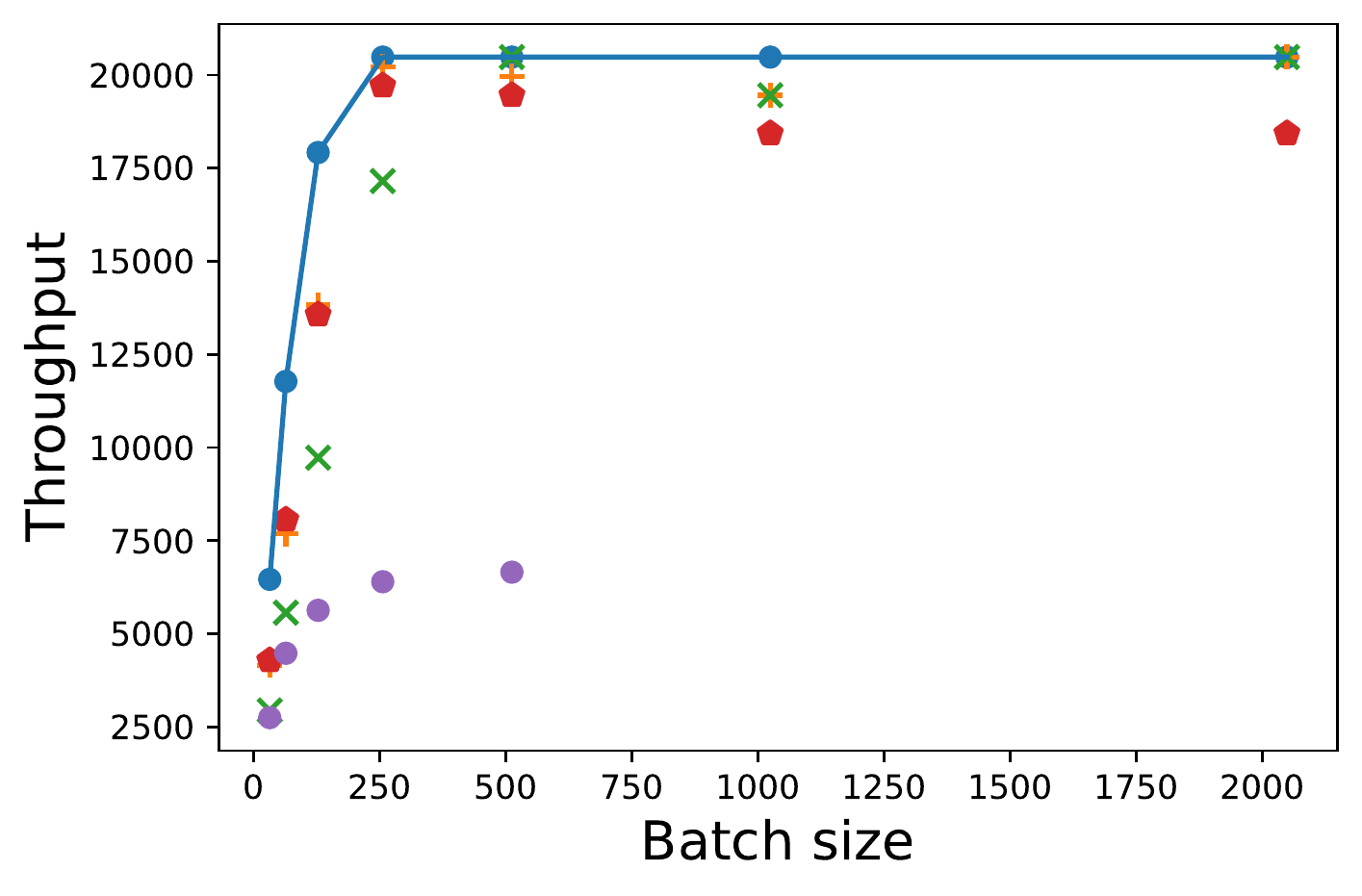}
    \includegraphics[width=0.4\linewidth]{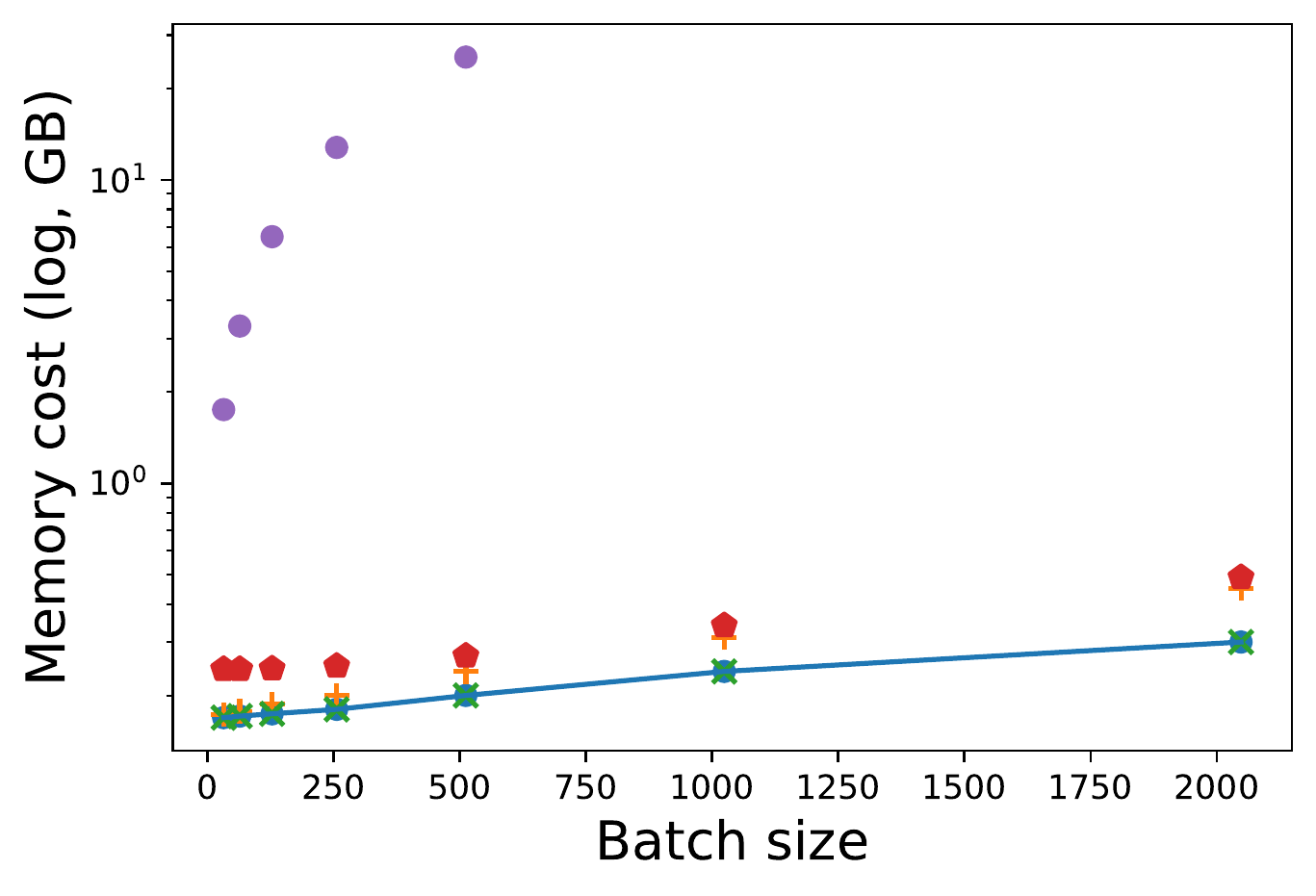}  
    \caption{Ablation study of MLP on CIFAR10/CIFAR100 (images are flattened into vectors). Default model: 10 layers, width 1000, batch size 256.}
    \label{fig:MLP_efficiencies}
\end{figure}

\begin{table}[!htb]
    \centering
    \resizebox{\linewidth}{!}{
    \begin{tabular}{c|c|c|c|c|c}
         &BK&Non-DP& GhostClip&Opacus
         \\\hline
         \multirow{3}{*}{Time complexity}&$6B\sum_l T_{(l)}p_{(l)}d_{(l)}$&\multirow{3}{*}{$6B\sum_l T_{(l)}p_{(l)}d_{(l)}$}&\multirow{3}{*}{\shortstack{$10B\sum_l T_{(l)}p_{(l)}d_{(l)}$\\$+2B\sum_l T_{(l)}^2(p_{(l)}+d_{(l)})$}}&\multirow{3}{*}{$8B\sum_l T_{(l)}p_{(l)}d_{(l)}$}\\
&$+2B\sum_l\Big(\mathbb{I}\{2T_{(l)}^2<p_{(l)}d_{(l)}\}$&&&
\\
&$\cdot T_{(l)}^2(p_{(l)}+d_{(l)})\Big)$&&&\\\hline
RoBERTa-base&$15.3*10^{12}$&$13.1*10^{12} (0.86\times)$&$24.1*10^{12} (1.57\times)$&$17.5*10^{12} (1.14\times)$\\\hline
RoBERTa-large&$52.3*10^{12}$&$46.5*10^{12} (0.89\times)$&$83.3*10^{12} (1.59\times)$&$62.0*10^{12} (1.18\times)$\\\hline
ViT-base&$11.2*10^{12}$&$10.1*10^{12} (0.90\times)$&$18.0*10^{12} (1.60\times)$&$13.5*10^{12} (1.20\times)$\\\hline
ViT-large&$38.8*10^{12}$&$35.8*10^{12} (0.92\times)$&$62.7*10^{12} (1.61\times)$&$47.7*10^{12} (1.23\times)$\\\hline
BEiT-large&$29.1*10^{12}$&$26.9*10^{12} (0.92\times)$&$47.1*10^{12} (1.61\times)$&$35.8*10^{12} (1.23\times)$\\\hline
GPT2-small&$7.7*10^{12}$&$7.5*10^{12} (0.96\times)$&$12.7*10^{12} (1.64\times)$&$10.0*10^{12} (1.28\times)$\\\hline
GPT2-medium&$22.1*10^{12}$&$21.4*10^{12} (0.96\times)$&$36.2*10^{12} (1.64\times)$&$28.4*10^{12} (1.29\times)$\\\hline
GPT2-large&$47.9*10^{12}$&$46.4*10^{12} (0.97\times)$&$78.8*10^{12} (1.65\times)$&$61.9*10^{12} (1.30\times)$
     \\\hline
\rowcolor{LightCyan}
 GPT2-small&$9.3*10^{13}$&$7.5*10^{13} (0.80\times)$&$15.5*10^{13} (1.66\times)$&$9.9*10^{12} (1.07\times)$\\\hline
\rowcolor{LightCyan}
GPT2-medium&$28.2*10^{13}$&$21.4*10^{13} (0.76\times)$&$43.4*10^{13} (1.54\times)$&$28.4*10^{13} (1.01\times)$\\\hline
\rowcolor{LightCyan}
GPT2-large&$59.4*10^{13}$&$46.4*10^{13} (0.79\times)$&$92.2*10^{13} (1.55\times)$&$61.9*10^{13} (1.04\times)$
     \\\hline\hline
 \multirow{3}{*}{Space complexity}&\multirow{3}{*}{\shortstack{$B\sum_l \min\{2T_{(l)}^2,p_{(l)}d_{(l)}\}$\\$+B\sum_l T_{(l)}(3d_{(l)}+p_{(l)})$}}&\multirow{3}{*}{\shortstack{$\sum_l p_{(l)}d_{(l)}$\\$+B\sum_l T_{(l)}(3d_{(l)}+p_{(l)})$}}&\multirow{3}{*}{\shortstack{$2B\sum_l T_{(l)}^2$\\$+B\sum_l T_{(l)}(3d_{(l)}+p_{(l)})$}}&\multirow{3}{*}{\shortstack{$B\sum_l p_{(l)}d_{(l)}$\\$+B\sum_l T_{(l)}(3d_{(l)}+p_{(l)})$}}\\
 &&&&\\
 &&&&\\\hline
RoBERTa-base&$5.3*10^{9}$&$4.5*10^{9} (0.84\times)$&$5.3*10^{9} (1.00\times)$&$16.7*10^{9} (3.17\times)$\\\hline
RoBERTa-large&$13.3*10^{9}$&$11.8*10^{9} (0.88\times)$&$13.3*10^{9} (1.00\times)$&$46.9*10^{9} (3.52\times)$\\\hline
ViT-base&$3.3*10^{9}$&$3.0*10^{9} (0.91\times)$&$3.3*10^{9} (1.00\times)$&$11.5*10^{9} (3.47\times)$\\\hline
ViT-large&$8.5*10^{9}$&$8.1*10^{9} (0.95\times)$&$8.5*10^{9} (1.00\times)$&$38.1*10^{9} (4.46\times)$\\\hline
BEiT-large&$6.4*10^{9}$&$6.1*10^{9} (0.95\times)$&$6.4*10^{9} (1.00\times)$&$28.6*10^{9} (4.46\times)$\\\hline
GPT2-small&$1.7*10^{9}$&$1.6*10^{9} (0.94\times)$&$1.7*10^{9} (1.00\times)$&$14.0*10^{9} (8.19\times)$\\\hline
GPT2-medium&$4.5*10^{9}$&$4.3*10^{9} (0.96\times)$&$4.5*10^{9} (1.00\times)$&$39.8*10^{9} (8.82\times)$\\\hline
GPT2-large&$8.47*10^{9}$&$8.17*10^{9} (0.97\times)$&$8.47*10^{9} (1.00\times)$&$85.5*10^{9} (10.1\times)$
\\\hline
\rowcolor{LightCyan}
GPT2-small&$2.3*10^{10}$&$1.5*10^{10} (0.68\times)$&$2.5*10^{10} (1.10\times)$&$2.8*10^{10} (1.20\times)$\\\hline
\rowcolor{LightCyan}
GPT2-medium&$5.7*10^{10}$&$4.0*10^{10} (0.70\times)$&$6.0*10^{10} (1.04\times)$&$7.6*10^{10} (1.32\times)$\\\hline
\rowcolor{LightCyan}
GPT2-large&$10.1*10^{10}$&$7.5*10^{10} (0.75\times)$&$10.5*10^{10} (1.02\times)$&$15.2*10^{10} (1.48\times)$
\\\hline
 \end{tabular}
    }
    \caption{Time (upper half) and space (lower half) complexity of implementations ($B=100$). For text classification, $T=256$ and we use BK base ($\equiv$ BK-MixOpt). For vision transformers and ImageNet, $T=224\times 224$ and we use BK-MixOpt. For text generation (GPT2, which has token length limit as 1024), we use $T=100$ in black or \hl{$1000$ in light cyan}. We mark the ratio of an algorithm's complexity to BK's inside the parenthesis. Note that neither per-sample gradient instantiation (Opacus) nor ghost norm trick (GhostClip) is satisfying when $T$ is large, and they are dominated by BK-MixOpt.}
\label{tab:complexity overhead}
\end{table}

\begin{table}[!htb]
    \centering
    \resizebox{\linewidth}{!}{
    \begin{tabular}{c|c|c|c|c|c}
    Model&Algorithm&Maximum batch size&Time/Epoch&Maximum throughput&Speedup by BK\\\hline
\multirow{4}{*}{\shortstack{RoBERTa-large\\SST-2}}& BK (ours)&41&13:03&86&---\\
& Non-private&51&9:50&114&$0.75\times$\\
& GhostClip&48&17:34&64&$1.34\times$\\
& Opacus&16&22:30&50&$1.72\times$\\\hline
\multirow{4}{*}{\shortstack{RoBERTa-large\\QNLI}}& BK (ours)&41&20:14&86&---\\
& Non-private&51&15:33&112&$0.77\times$\\
& GhostClip&48&27:45&63&$1.37\times$\\
& Opacus&16&35:03&50&$1.73\times$\\\hline
\multirow{4}{*}{\shortstack{RoBERTa-large\\QQP}}& BK (ours)&41&70:04&87&---\\
& Non-private&51&53:42&113&$0.77\times$\\
& GhostClip&48&95:09&64&$1.36\times$\\
& Opacus&16&137:00&44&$1.96\times$\\\hline
\multirow{4}{*}{\shortstack{RoBERTa-large\\MNLI}}& BK (ours)&41&77:07&85&---\\
& Non-private&51&58:02&113&$0.75\times$\\
& GhostClip&48&103:30&63&$1.34\times$\\
& Opacus&16&134:30&49&$1.75\times$\\\hline\hline

\multirow{4}{*}{GPT2}& BK (ours)&149&2:13&316&---\\
& Non-private&157&1:47&393&$0.80\times$\\
& GhostClip&156&2:54&242&$1.31\times$\\
& Opacus&43&5:03&139&$2.27\times$\\\hline
\multirow{4}{*}{GPT2-medium}& BK (ours)&69&4:58&141&---\\
& Non-private&70&4:05&172&$0.82\times$\\
& GhostClip&70&6:46&104&$1.36\times$\\
& Opacus&15&14:22&49&$2.88\times$\\\hline
\multirow{4}{*}{GPT2-large}& BK (ours)&29&10:01&70&---\\
& Non-private&29&8:16&85&$0.83\times$\\
& GhostClip&29&13:56&50&$1.36\times$\\
& Opacus&5&44:05&16&$4.41\times$\\\hline\hline

\multirow{4}{*}{BEiT-large}& BK (ours)&96&6:35&127&---\\
& Non-private&98&4:55&169&$0.76\times$\\
& GhostClip&95&8:53&93&$1.33\times$\\
& Opacus&5&4:12:00&3&$38.3\times$\\\hline
\end{tabular}
}
    \caption{Extension of \Cref{tab:SOTA throughput}. Note that CIFAR means both CIFAR10 and CIFAR100. Performance of GPT2 on E2E dataset (same setting as \cite{li2021large,bu2022automatic}).}
    \label{tab:GPT max throughput}
\end{table}

\begin{table}[!htb]
    \centering
\setlength\tabcolsep{1pt}
    \resizebox{\linewidth}{!}{
    \begin{tabular}{c|c|c|c|c|c|c}
         \multirow{2}{*}{Model}&Mixed ghost norm (MGN)& \multicolumn{2}{|c|}{Per-sample grad instantiation}&\multicolumn{2}{|c|}{Ghost norm}
         \\
         &$\sum_l\min\{2T_{(l)}^2,p_{(l)}d_{(l)}\}$&($\sum_l p_{(l)}d_{(l)}$; \# param)&Saving by MGN&($\sum_l 2T_{(l)}^2=2H_\text{out}^2 W_\text{out}^2$)&Saving by MGN
         \\\hline\hline
         ResNet18&1.0M&11.5M&11.5$\times$&399M&399$\times$
         \\\hline
         ResNet34&2.3M&21.6M&9.4$\times$&444M&194$\times$
         \\\hline
         ResNet50&2.8M&22.7M&8.0$\times$&528M&186$\times$
         \\\hline
         ResNet101&6.8M&41.7M&6.2$\times$&532M&79$\times$
         \\\hline
         ResNet152&10.9&57.3M&5.3$\times$&549M&51$\times$
         \\\hline
         DenseNet121&4.1M&7.9M&1.9$\times$&605M&147$\times$
         \\\hline
         DenseNet161&9.0M&28.5M&3.2$\times$&607M&67$\times$
         \\\hline
         DenseNet201&7.0M&19.8M&2.8$\times$&609M&87$\times$
         \\\hline
         Wide ResNet50&5.6M&66.0M&11.7$\times$&528M&93$\times$
         \\\hline
         Wide ResNet101&9.6M&124.0M&13.0$\times$&531M&56$\times$
         \\\hline
         vit\_tiny\_patch16\_224&3.3M&5.6M&1.7$\times$&3.8M&1.1$\times$
         \\\hline
vit\_small\_patch16\_224&3.8M&21.9M&5.8$\times$&13.8M&1.0$\times$
         \\\hline
vit\_base\_patch16\_224&3.8M&86.3M&22.7$\times$&3.8M&1.0$\times$
         \\\hline
vit\_large\_patch16\_224&7.5M&303.8M&40.4$\times$&7.5M&1.0$\times$
         \\\hline
crossvit\_tiny\_240&4.0M&6.9M&1.7$\times$&10.4M&2.6$\times$
         \\\hline
crossvit\_small\_240&5.9M&26.6M&4.5$\times$&10.4M&1.8$\times$
         \\\hline
crossvit\_base\_240&8.7M&104.5M&12.1$\times$&10.4M&1.2$\times$
         \\\hline
convnext\_small&12.4M&50.1M&4.0$\times$&214M&17$\times$
         \\\hline
convnext\_base&14.3M&88.4M&6.2$\times$&214M&15$\times$
         \\\hline
convnext\_large&19.8M&197.5M&10.0$\times$&214M&11$\times$
         \\\hline
deit\_tiny\_patch16\_224&3.3M&5.6M&1.7$\times$&3.8M&1.1$\times$
         \\\hline
deit\_small\_patch16\_224&3.8M&21.9M&5.8$\times$&3.8M&1.0$\times$
         \\\hline
deit\_base\_patch16\_224&3.8M&86.3M&22.7$\times$&3.8M&1.0$\times$
         \\\hline
beit\_base\_patch16\_224&2.9M&86.3M&29.8$\times$&2.9M&1.0$\times$
         \\\hline
beit\_large\_patch16\_224&5.7M&303.8M&53.3$\times$&5.7M&1.0$\times$
         \\\hline\hline
     \end{tabular}
    }
    \caption{Space complexity of computing per-sample gradient norm, on ImageNet image ($224\times 224$). The saving by the mixed ghost norm, adopted in BK-MixGhostClip and BK-MixOpt, is substantial.}
    \label{tab:layerwise complexity}
\end{table}

\clearpage
\section{Effect of hybridization: layerwise space complexity}
\label{app:effect of hybrid}
We demonstrate the effect of hybridization (i.e. mixed ghost norm \cite{bu2022scalable}) on the computation of per-sample gradient norm. We consider the moderate feature dimension and the high feature dimension, respectively. We conclude that ghost norm trick (adopted in GhostClip and BK) is favored closer to the input layer, whereas the per-sample gradient instantiation (adopted in Opacus and FastGradClip) is favored closer to the output layer.

\subsection{Effect by model achitecture ($T=224\times 224$)}
Generally speaking, CNN can benefit from hybridization, but vision transformers may not (unless the feature dimension is high, see next section for BEiT). 

\begin{figure}[!htb]
    \centering
\includegraphics[width=0.24\linewidth]{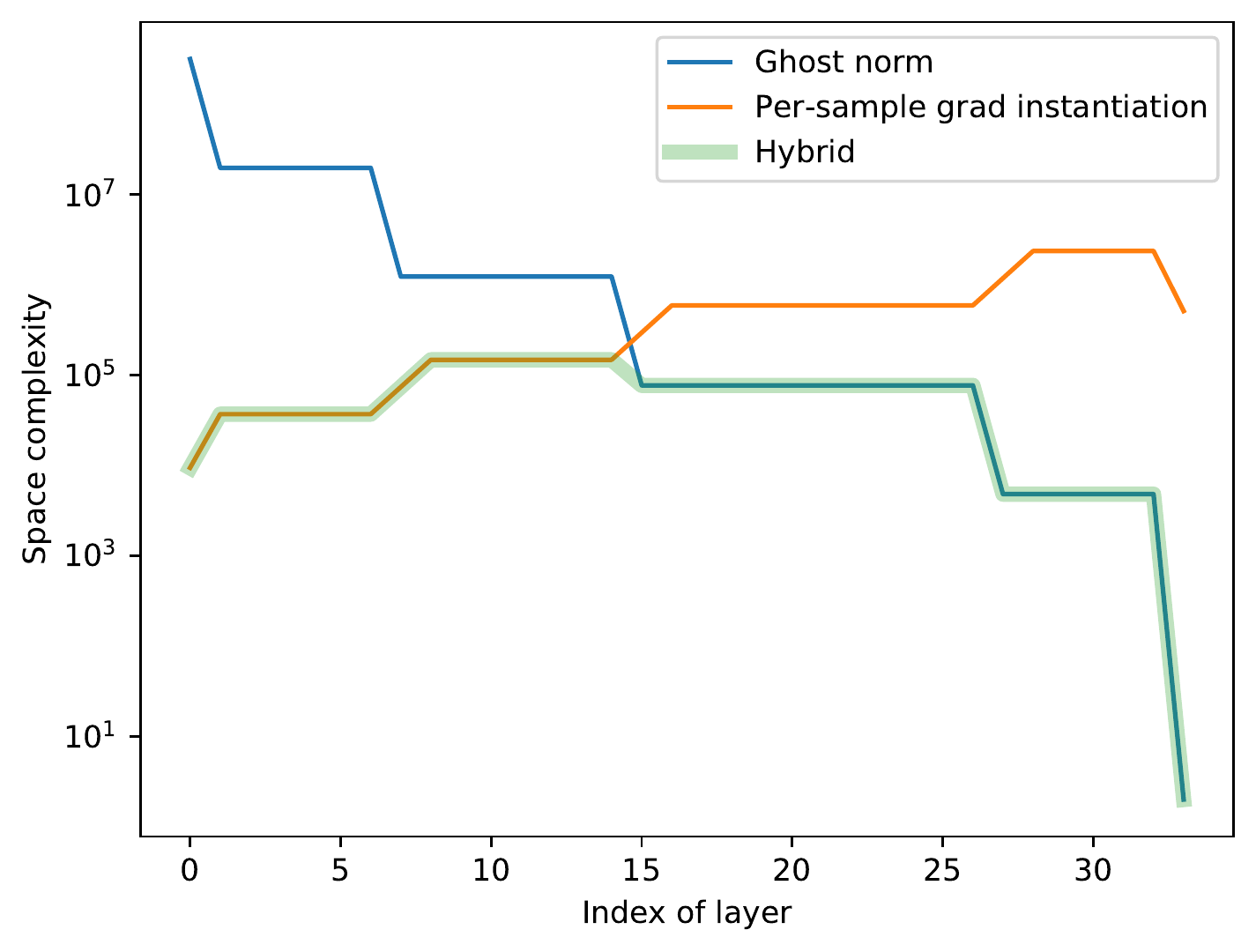}
\includegraphics[width=0.24\linewidth]{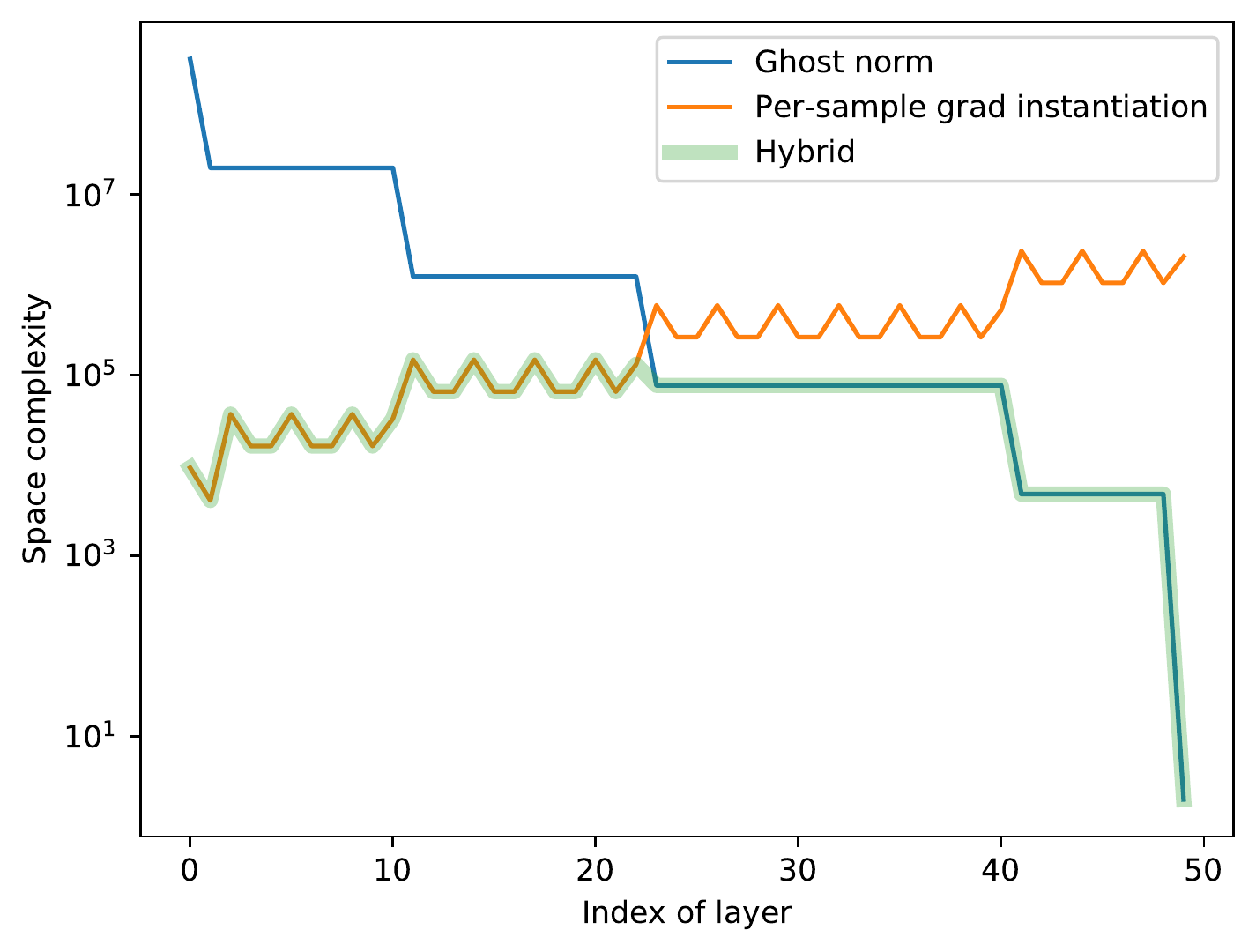}
\includegraphics[width=0.24\linewidth]{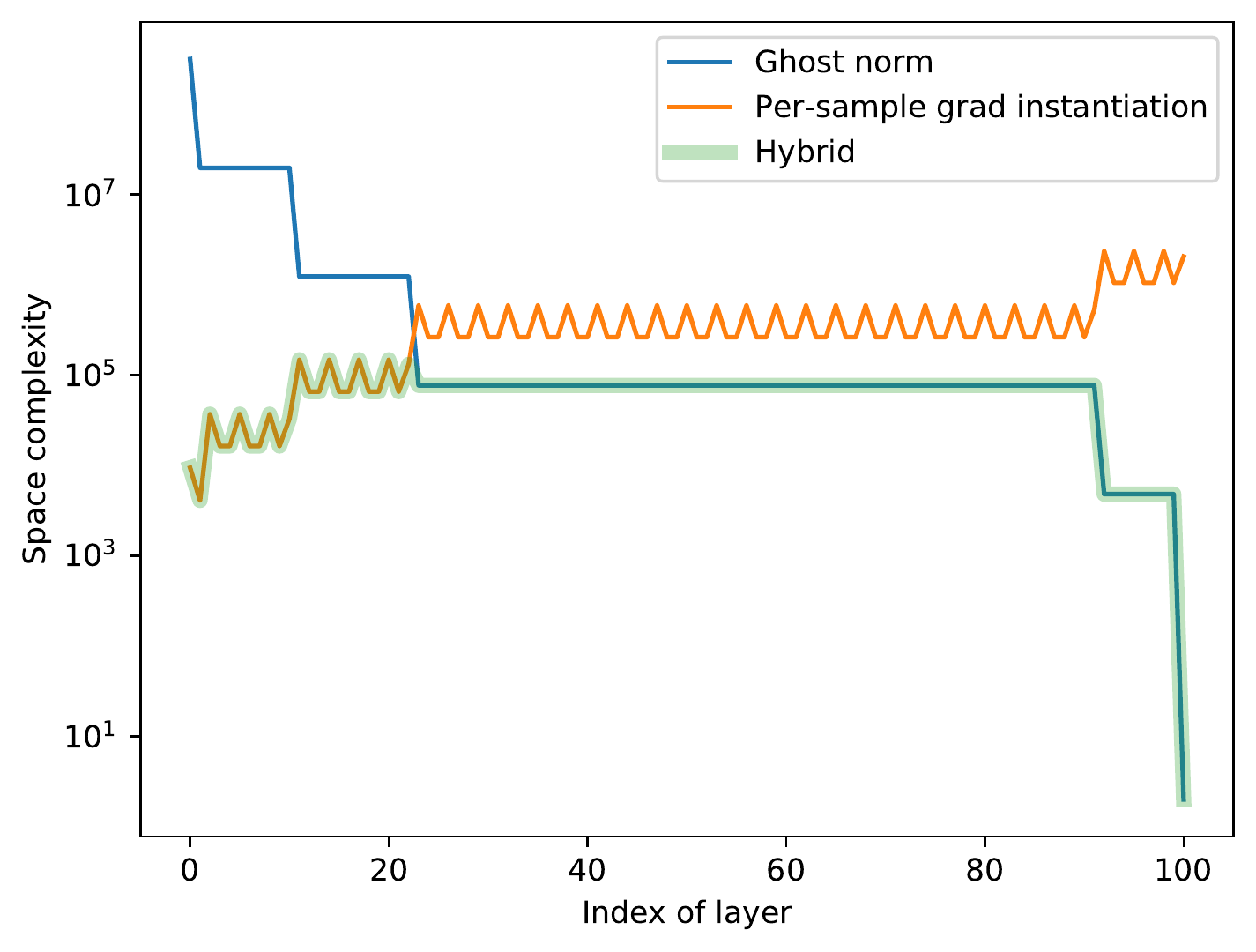}
\includegraphics[width=0.24\linewidth]{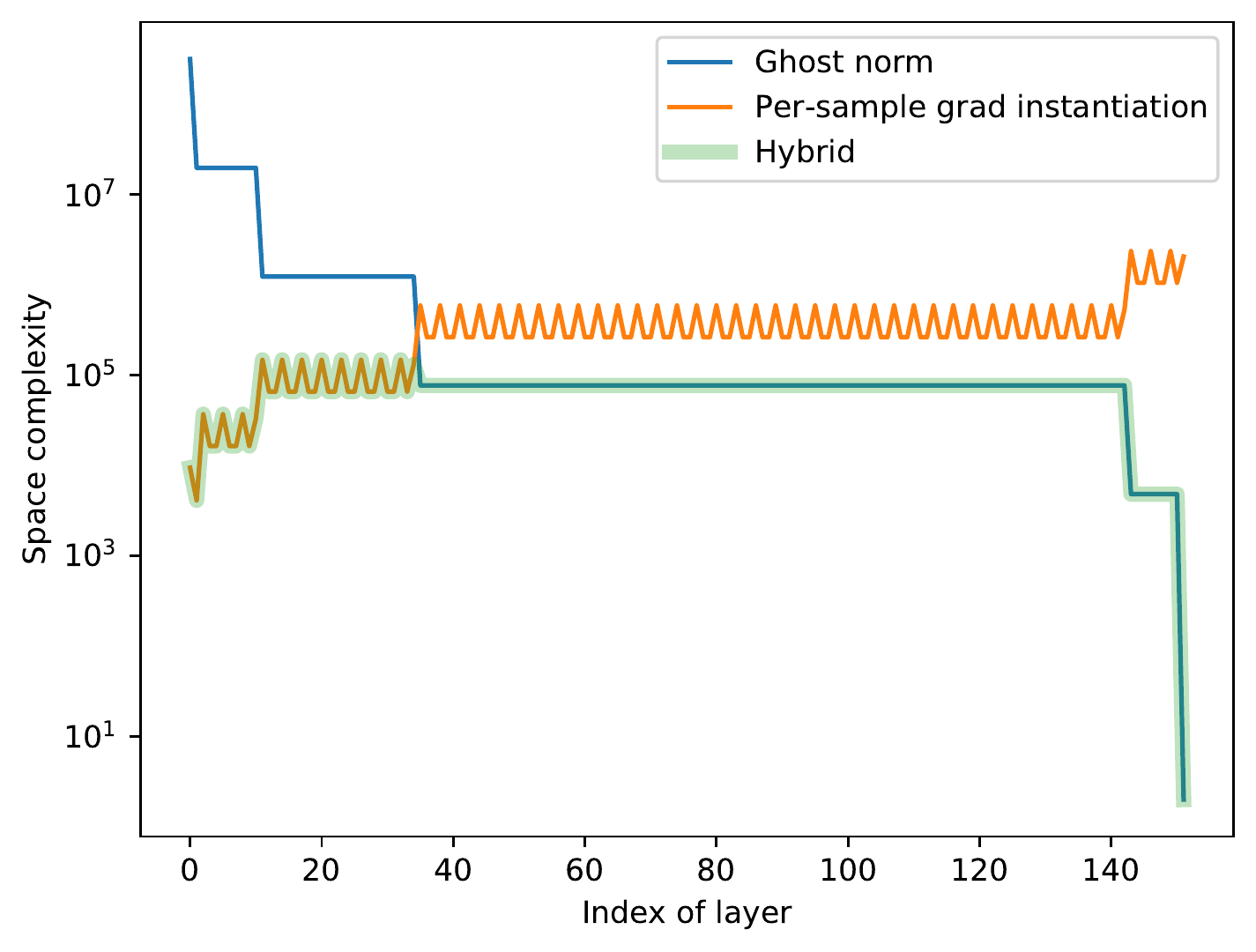}
\caption{Layerwise space complexity of computing the per-sample gradient norm. Left to right: ResNet 34/50/101/152.}
\end{figure}

\begin{figure}[!htb]
    \centering
\includegraphics[width=0.24\linewidth]{figs/vgg11_layer_complexity.pdf}
\includegraphics[width=0.24\linewidth]{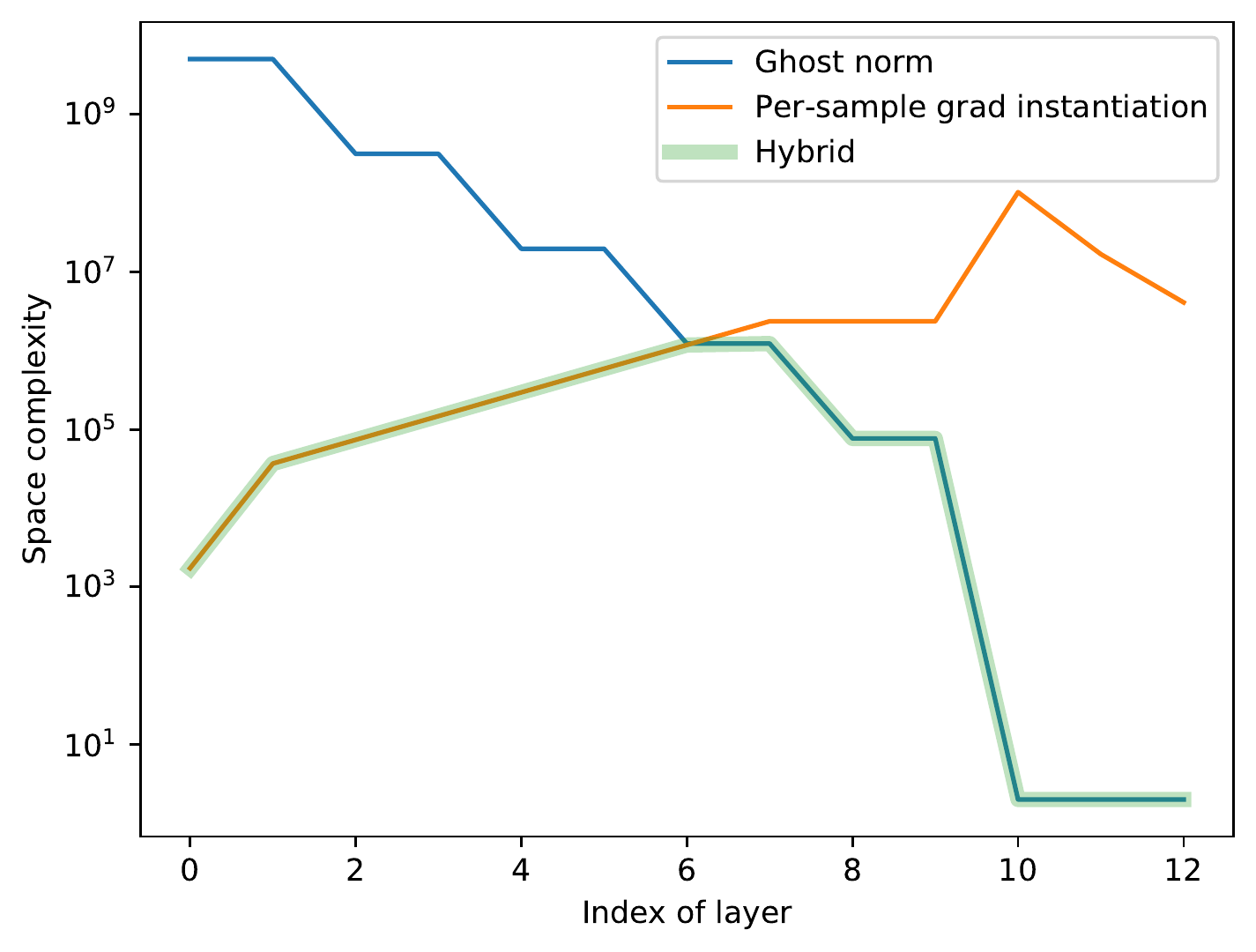}
\includegraphics[width=0.24\linewidth]{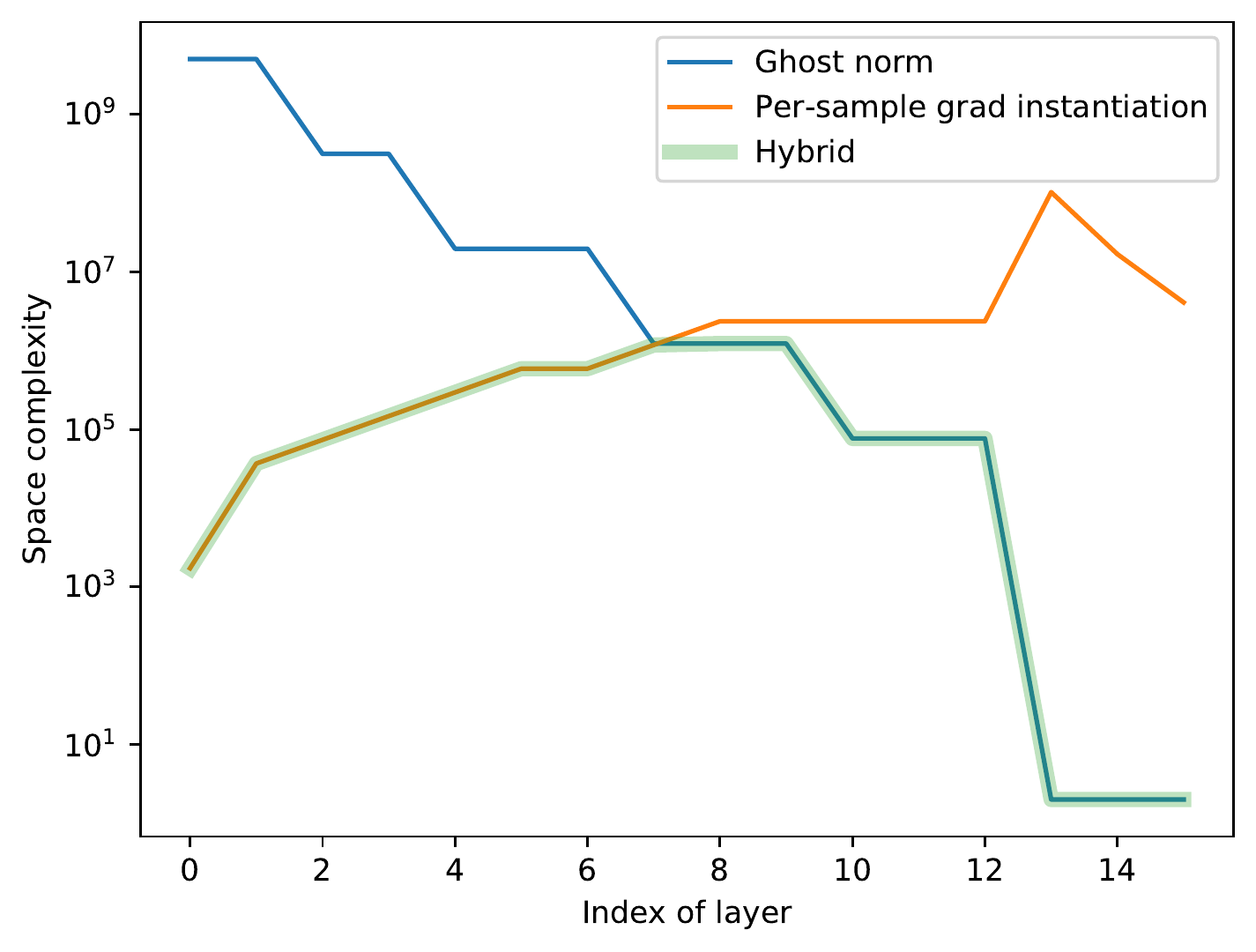}
\includegraphics[width=0.24\linewidth]{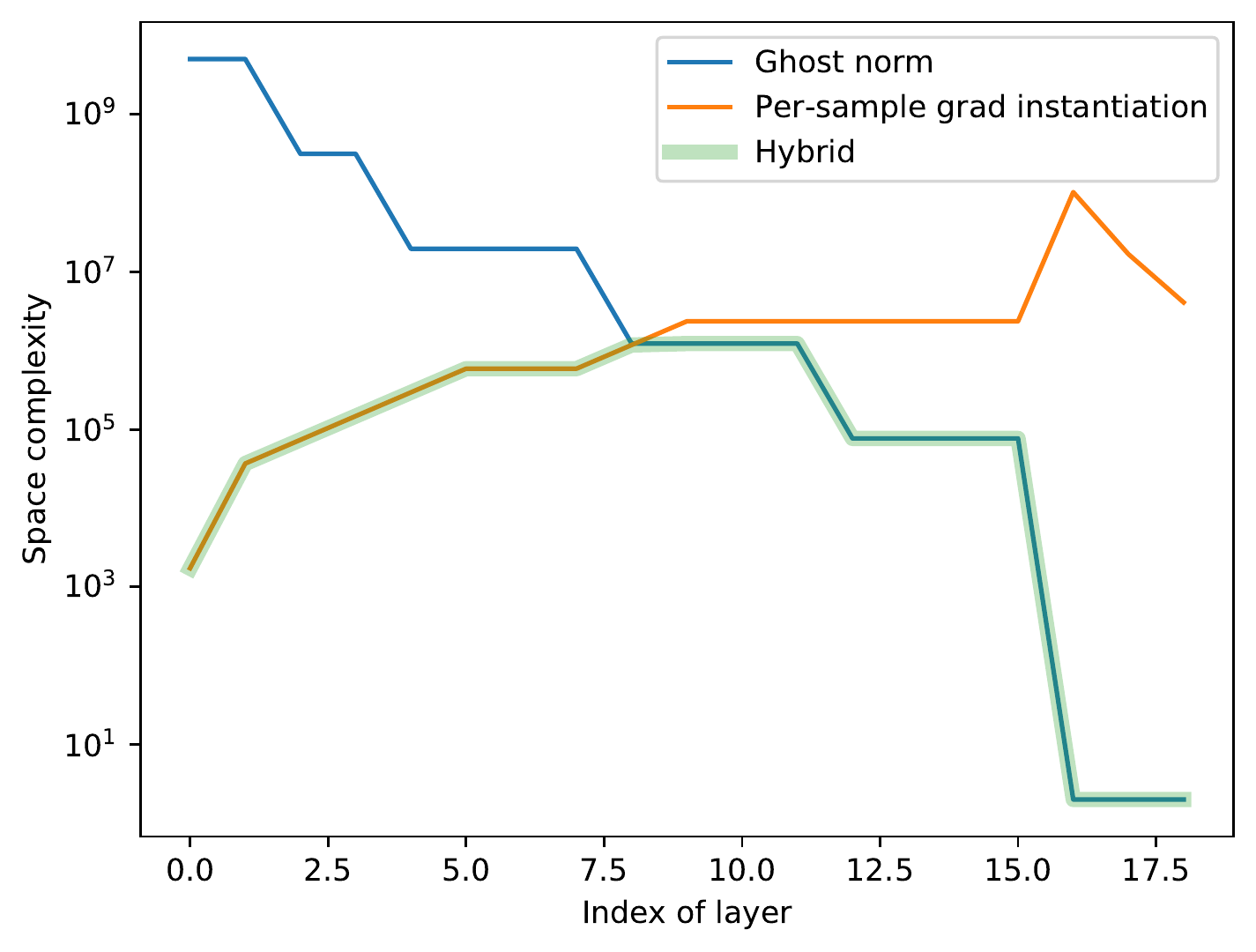}
\caption{Layerwise space complexity of computing the per-sample gradient norm. Left to right: VGG 11/13/16/19.}
\end{figure}

\begin{figure}[!htb]
    \centering
\includegraphics[width=0.3\linewidth]{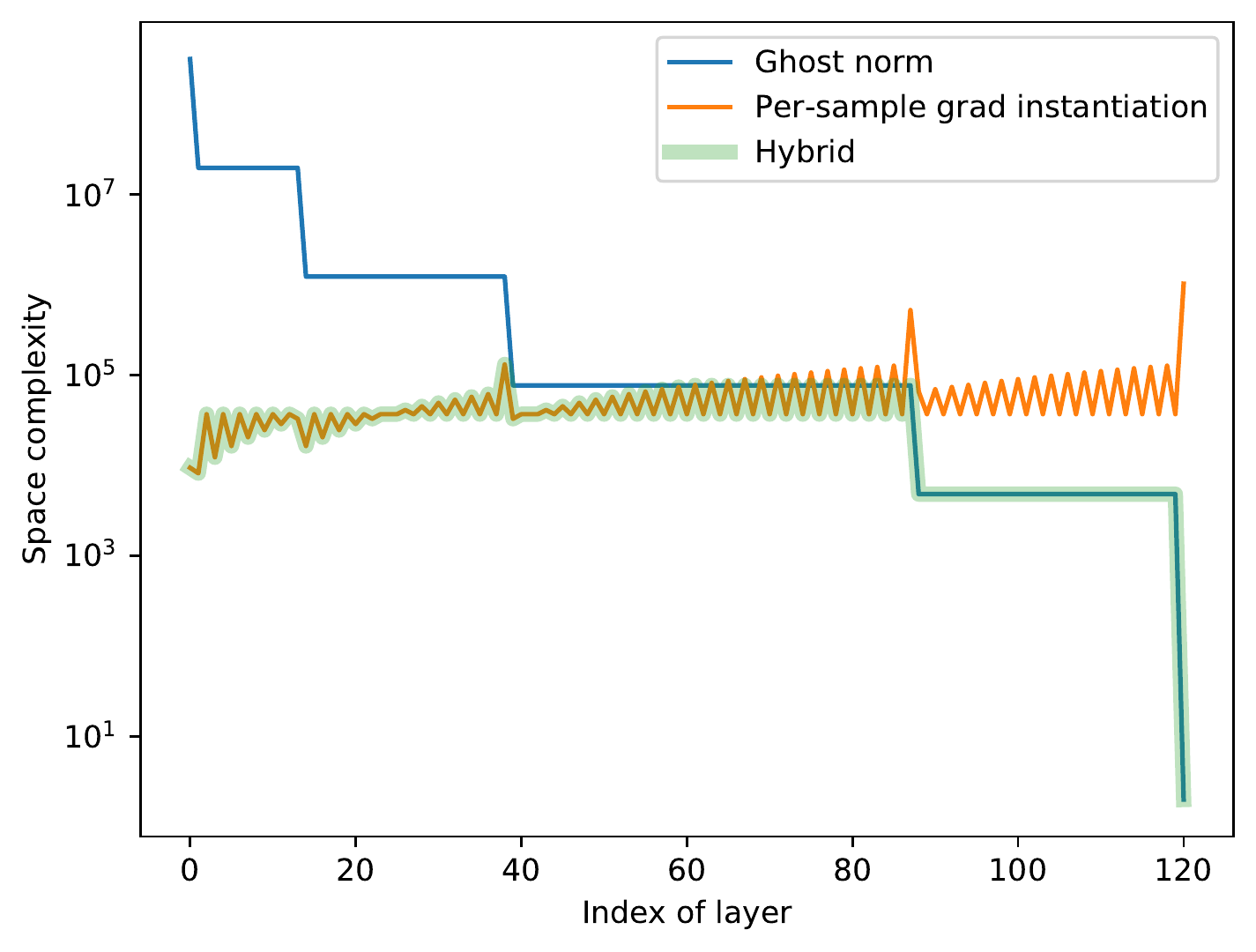}
\includegraphics[width=0.3\linewidth]{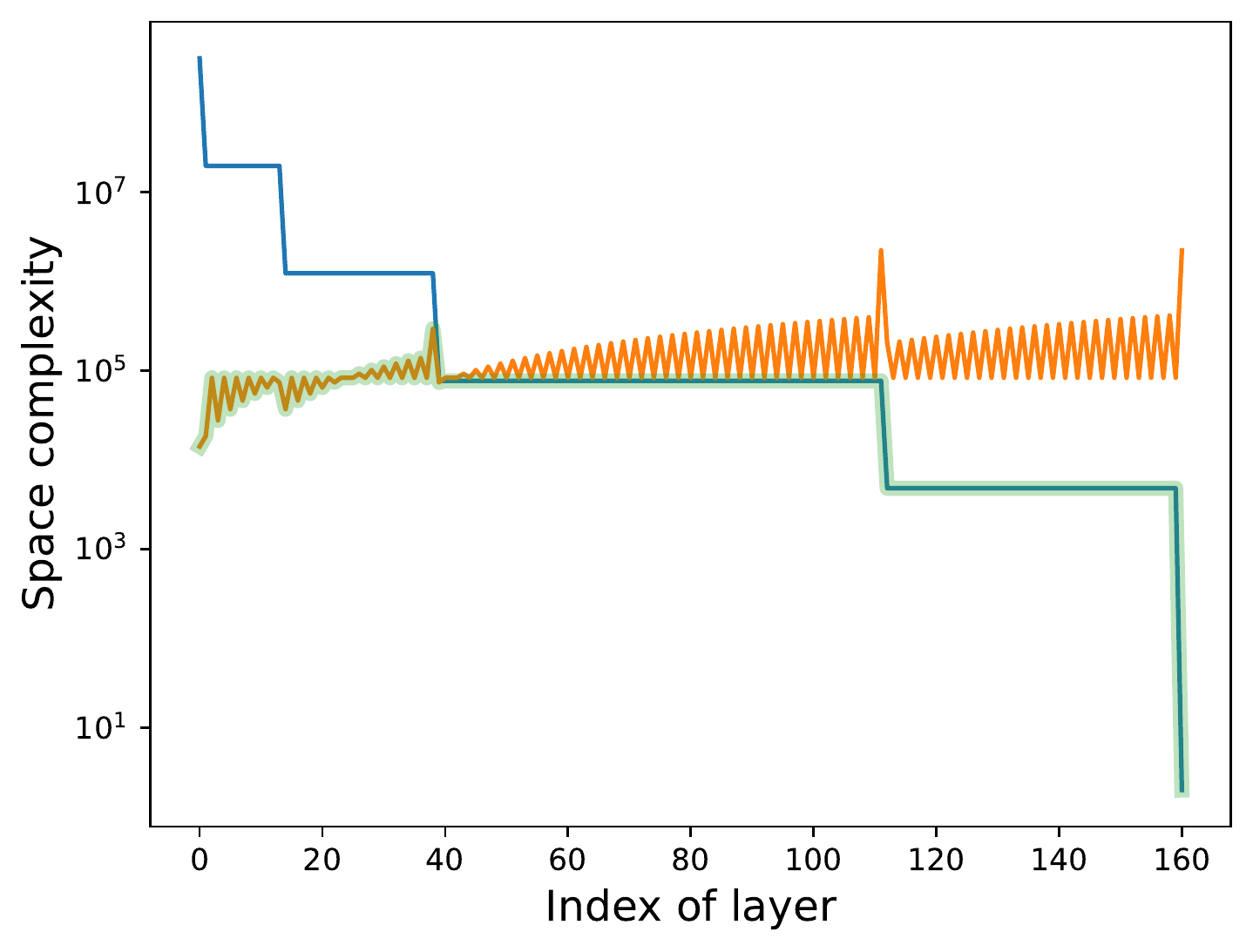}
\includegraphics[width=0.3\linewidth]{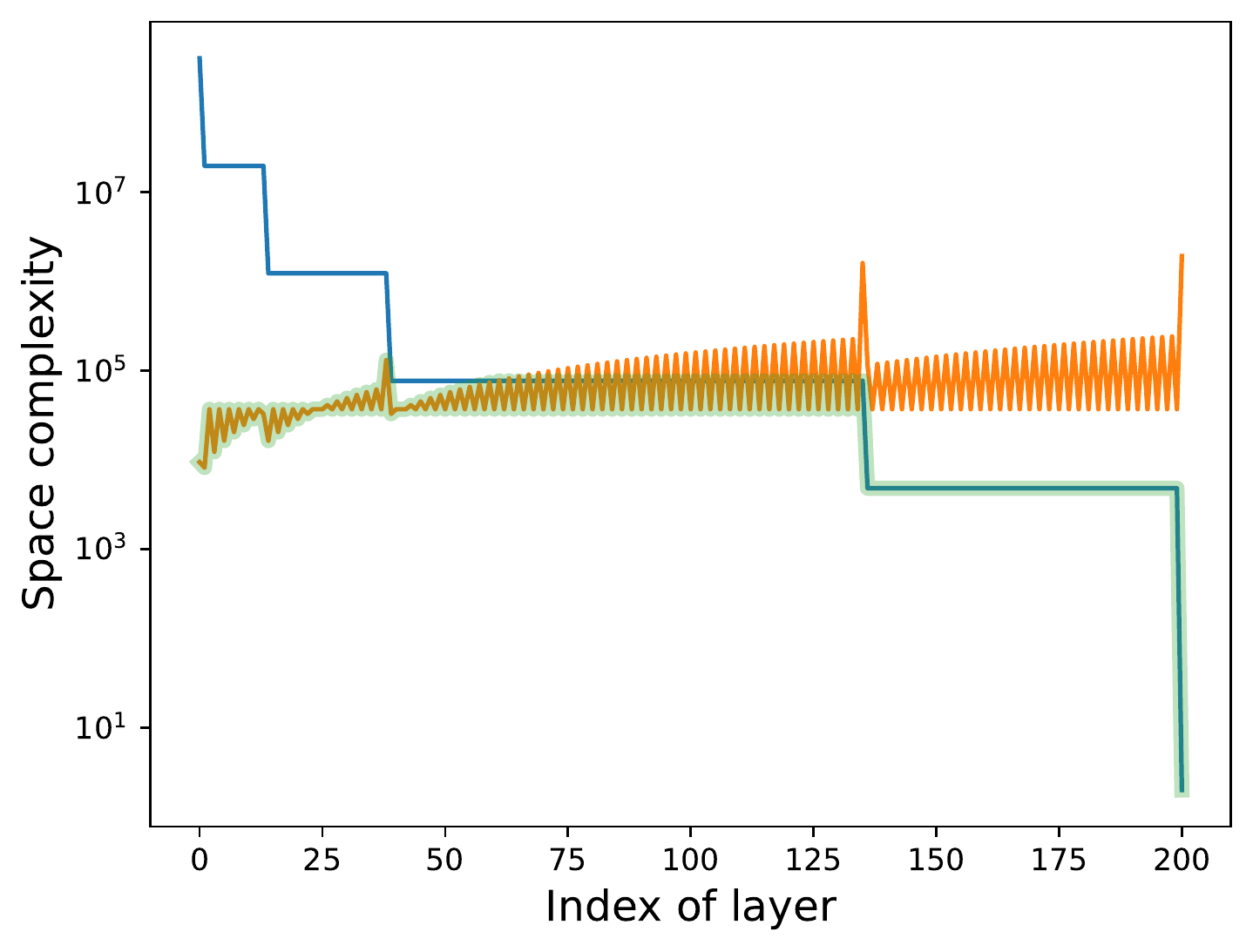}
\caption{Layerwise space complexity of computing the per-sample gradient norm. Left to right: DenseNet 121/161/201.}
\end{figure}

\begin{figure}[!htb]
    \centering
\includegraphics[width=0.24\linewidth]{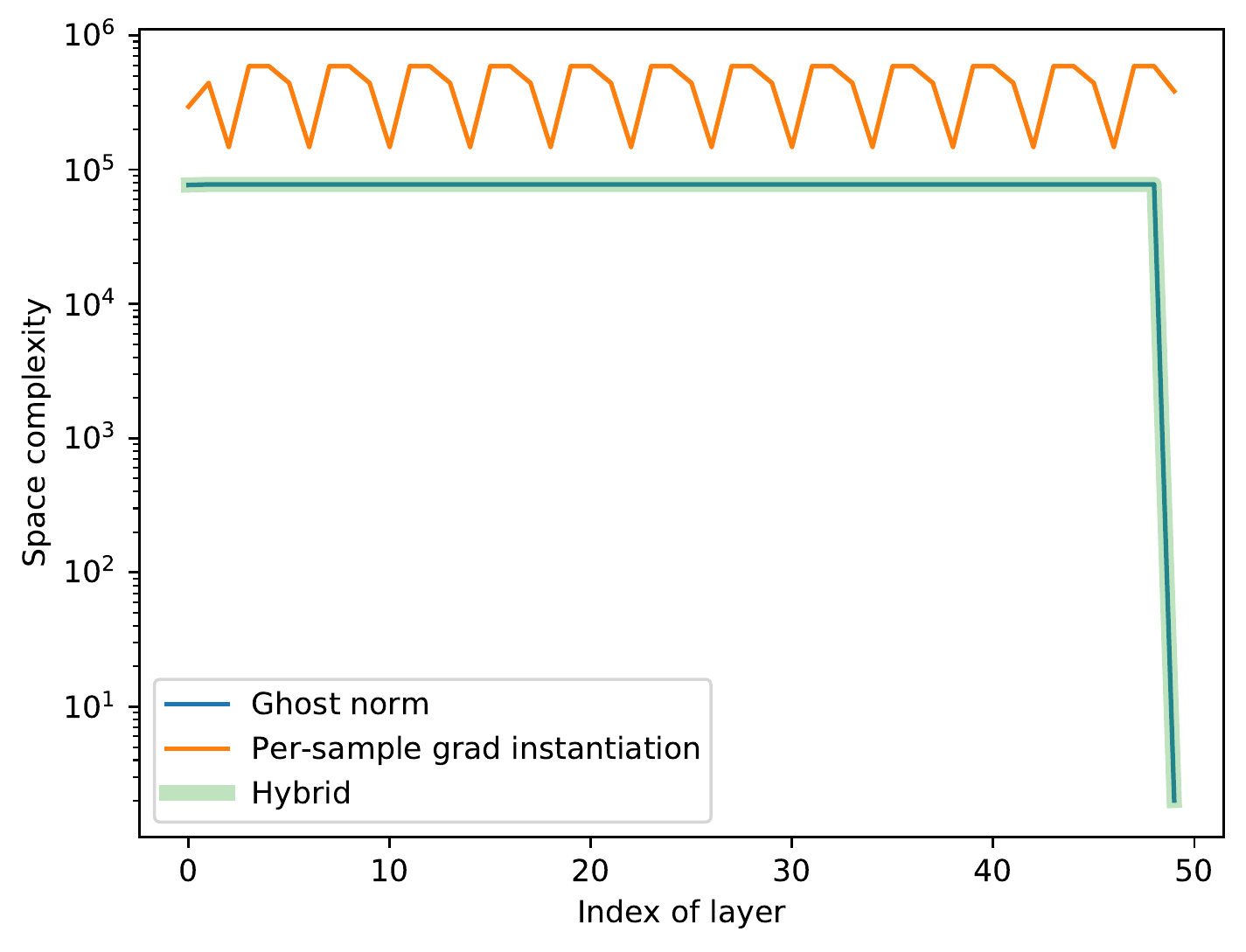}
\includegraphics[width=0.24\linewidth]{figs/vit_base_patch16_224_layer_complexity.pdf}
\includegraphics[width=0.24\linewidth]{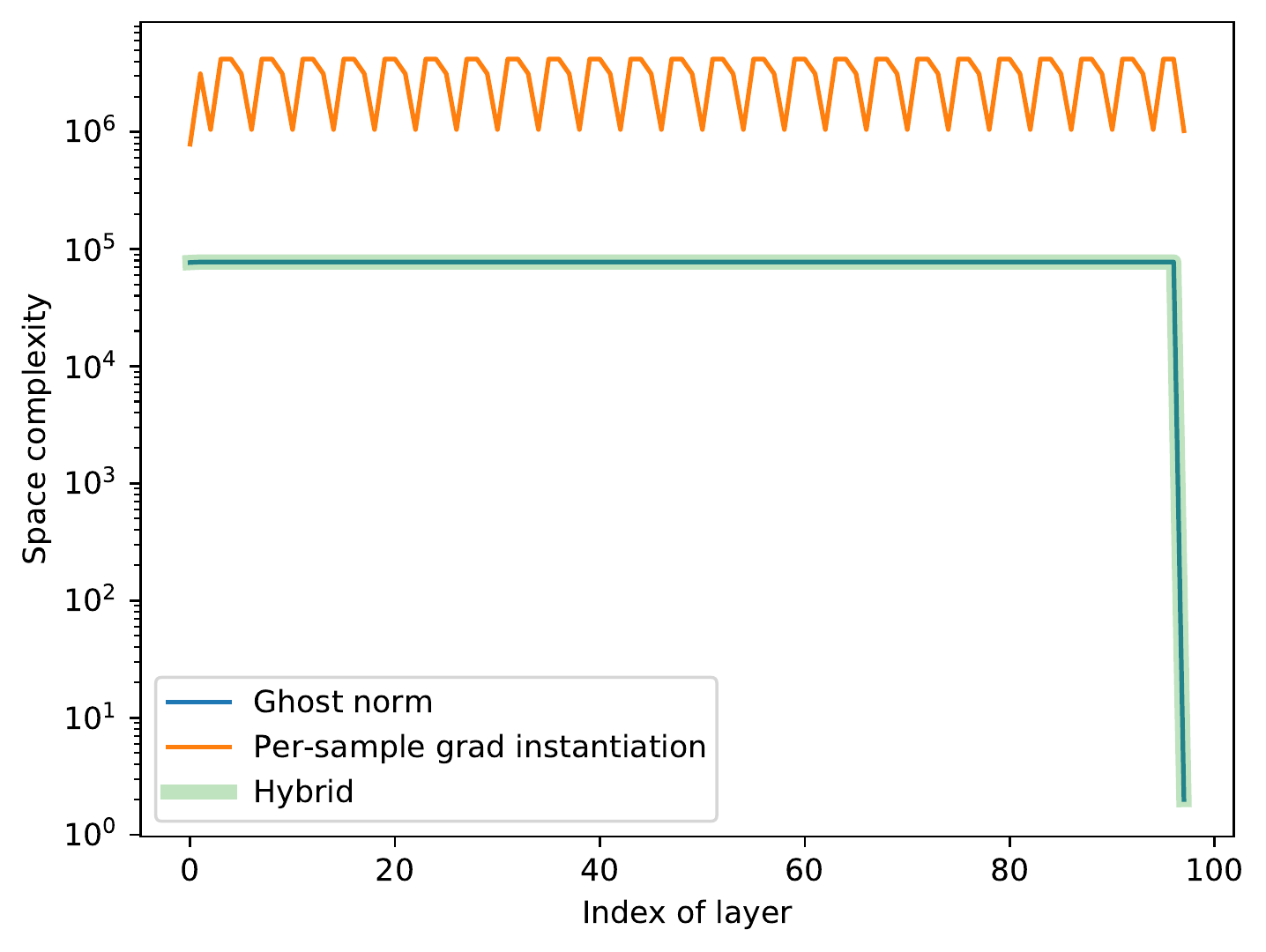}
\includegraphics[width=0.24\linewidth]{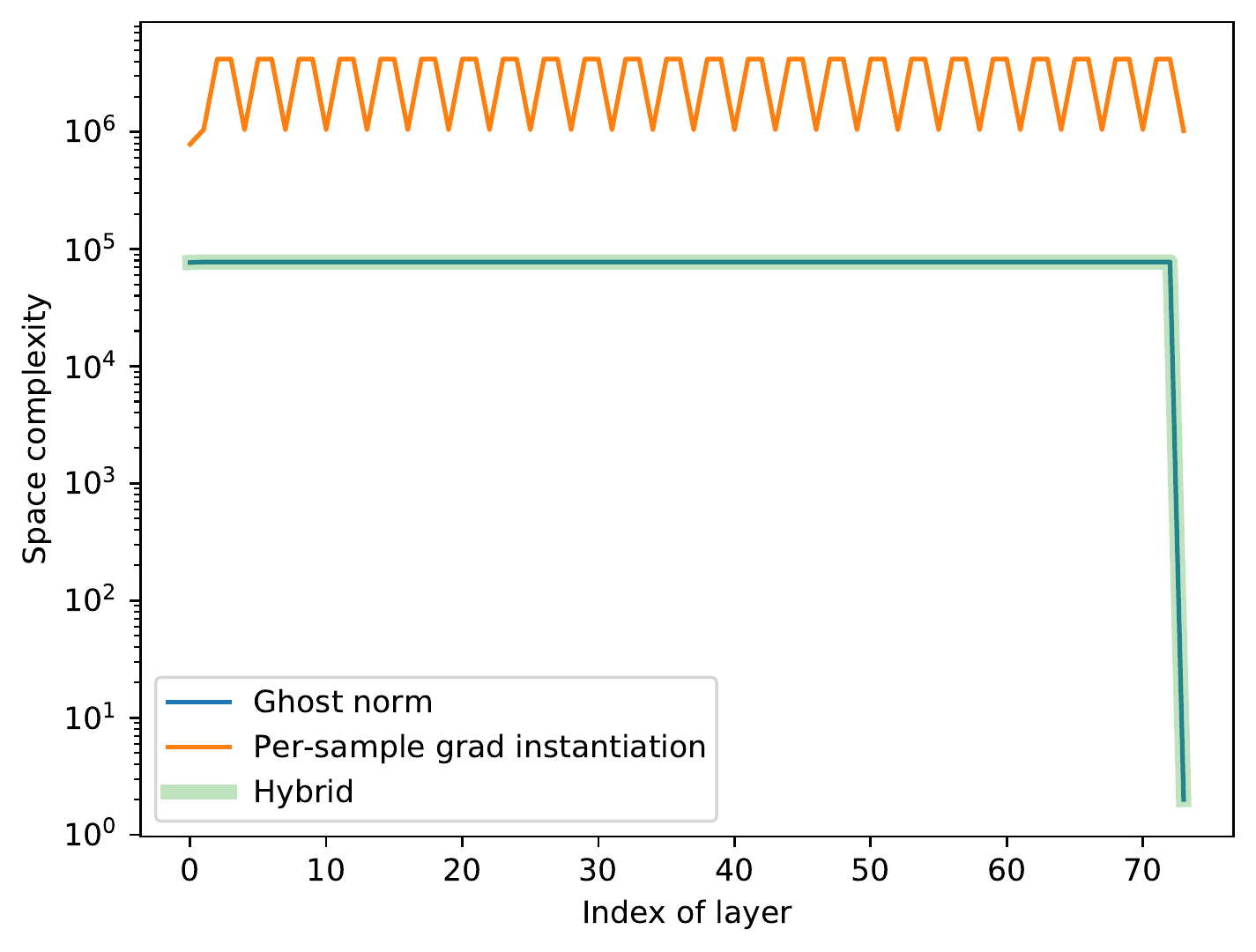}
\caption{Layerwise space complexity of computing the per-sample gradient norm. Left to right: ViT small/base/large, and BEiT-large.}
\end{figure}

\subsection{Effect by feature dimension ($T=32^2/224^2/512^2$)}
Generally speaking, higher feature dimension requires a deeper threshold, after which the per-sample gradient instantiation is not preferred. That is, high dimensional data does not prefer ghost norm. This pattern even holds for vision transformers, on which MixGhostClip/BK-MixGhostClip is equivalent to GhostClip/BK for low feature dimension.
\begin{figure}[!htb]
    \centering
\includegraphics[width=0.3\linewidth]{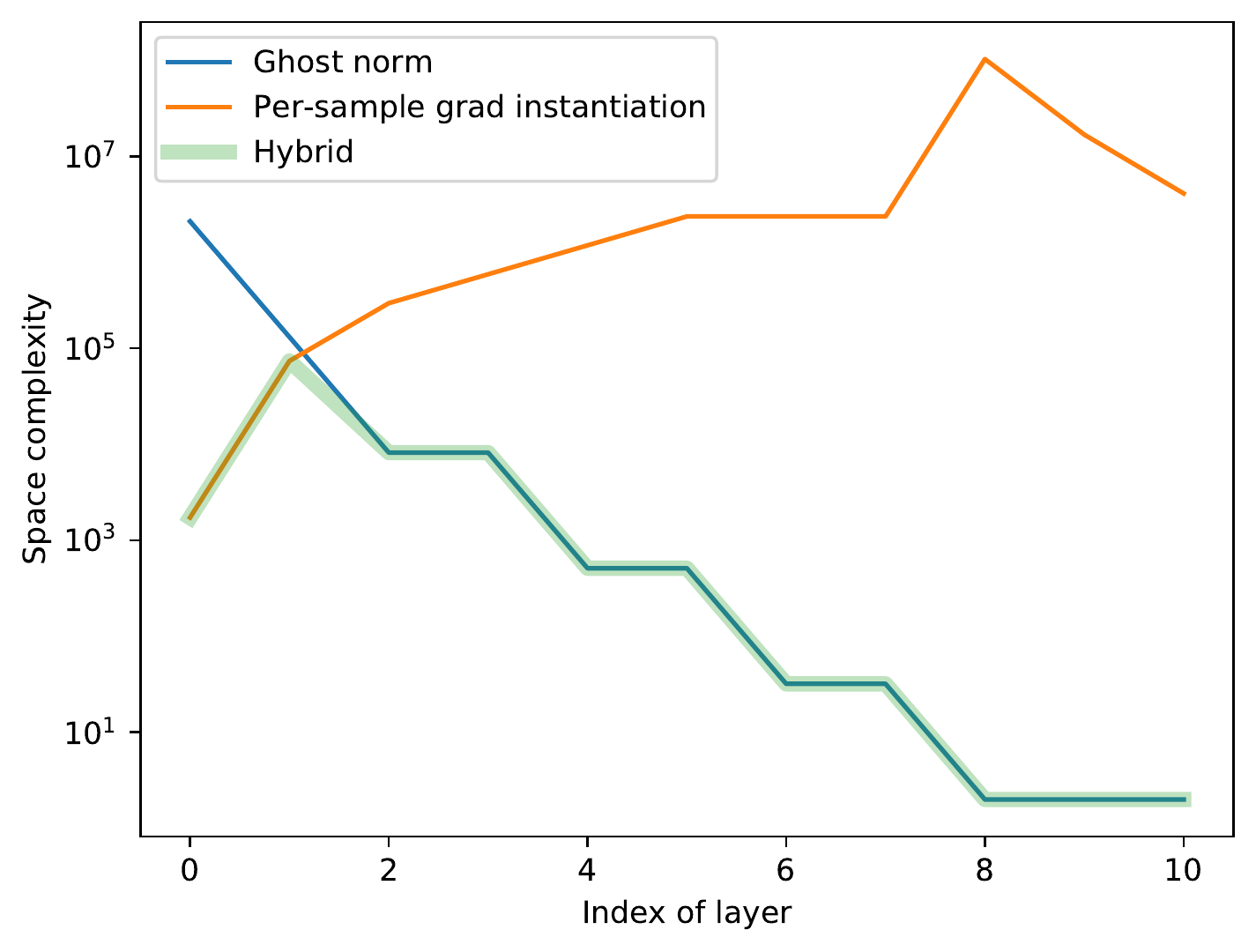}
\includegraphics[width=0.3\linewidth]{figs/vgg11_layer_complexity.pdf}
\includegraphics[width=0.3\linewidth]{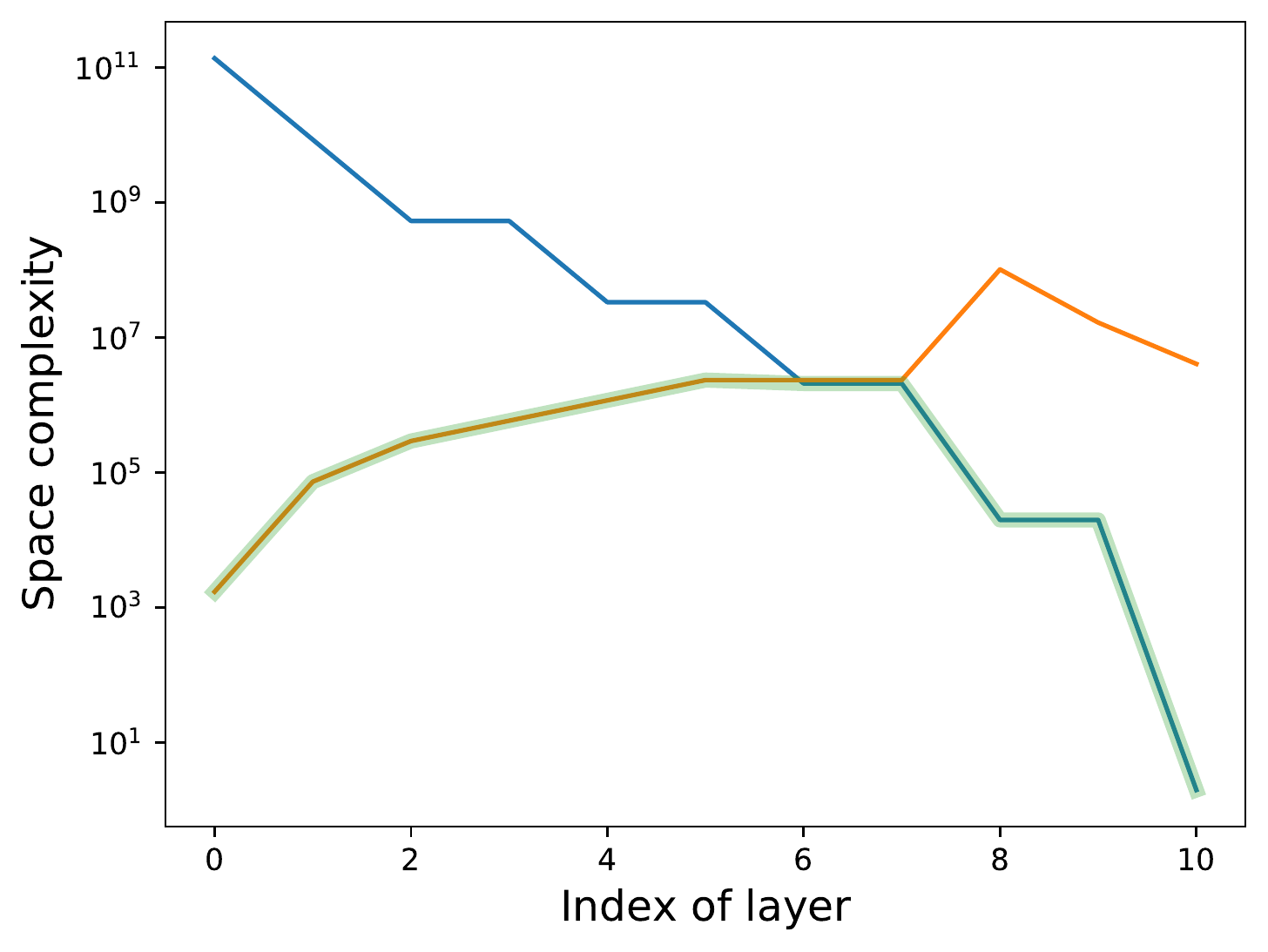}
\caption{Layerwise space complexity of computing the per-sample gradient norm in VGG11.}
\end{figure}

\begin{figure}[!htb]
    \centering
\includegraphics[width=0.3\linewidth]{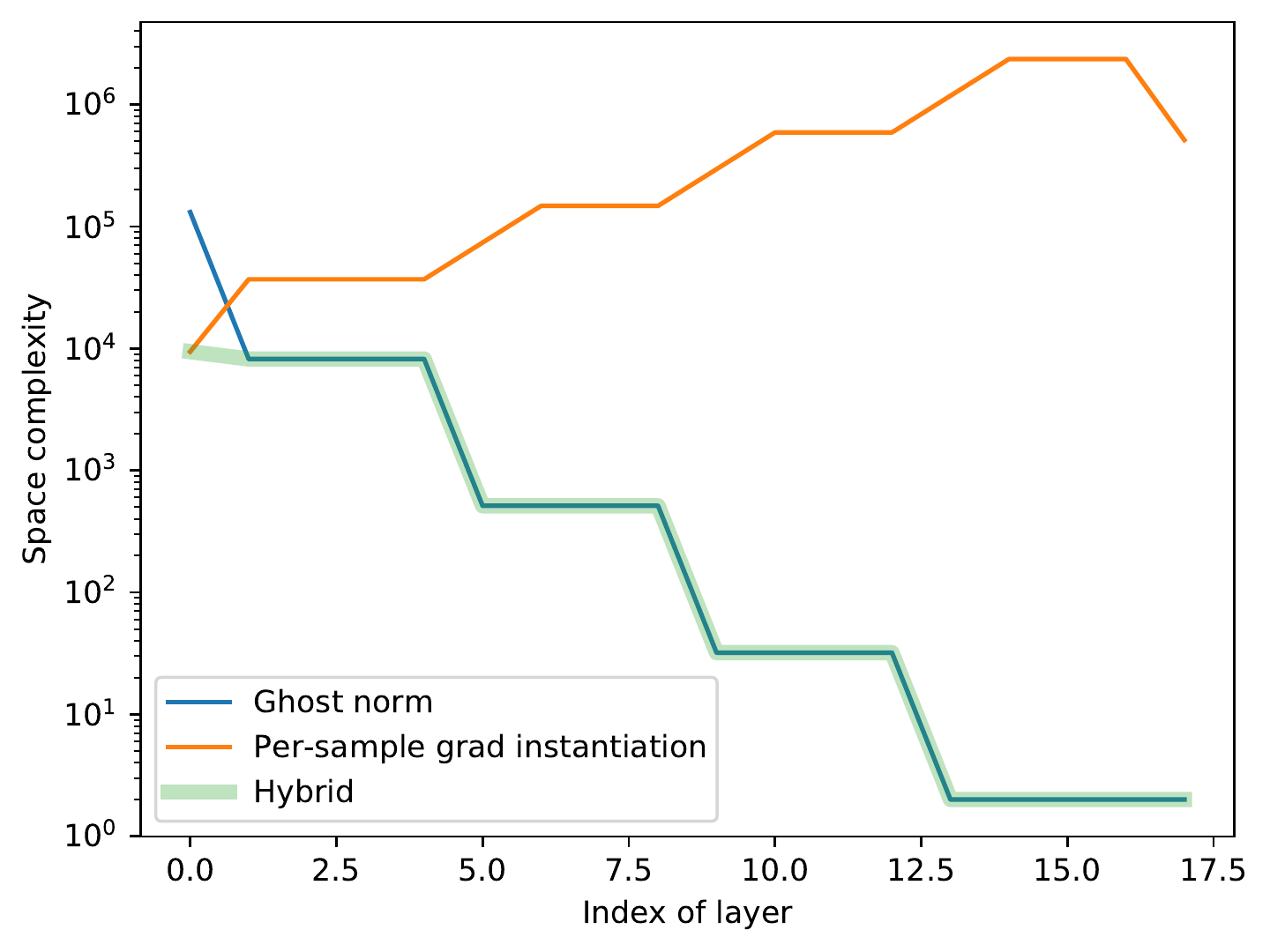}
\includegraphics[width=0.3\linewidth]{figs/resnet18_layer_complexity.pdf}
\includegraphics[width=0.3\linewidth]{figs/resnet18_layer_complexity_huge.pdf}
\caption{Layerwise space complexity of computing the per-sample gradient norm in ResNet18.}
\end{figure}

\begin{figure}[!htb]
    \centering
\includegraphics[width=0.3\linewidth]{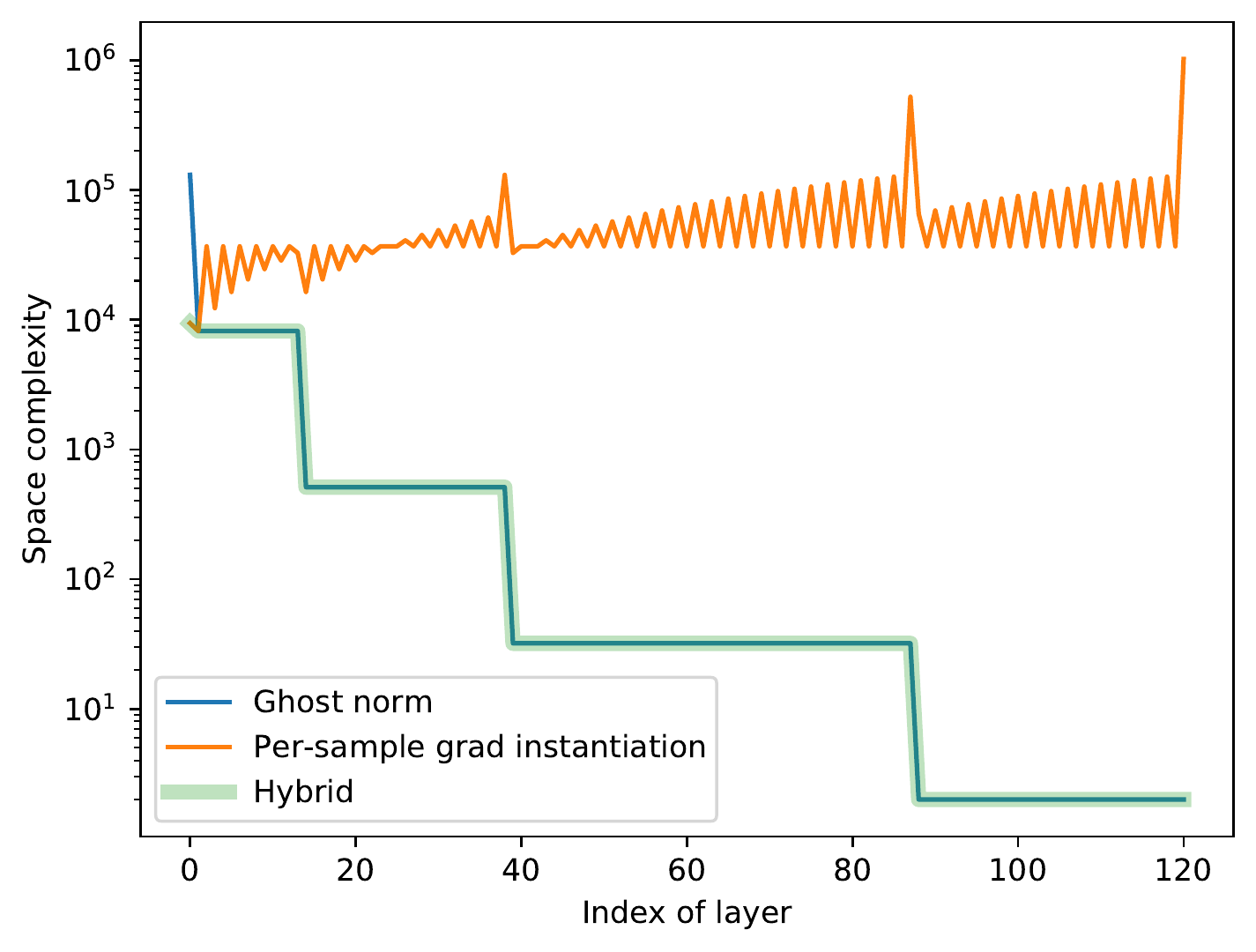}
\includegraphics[width=0.3\linewidth]{figs/densenet121_layer_complexity.pdf}
\includegraphics[width=0.3\linewidth]{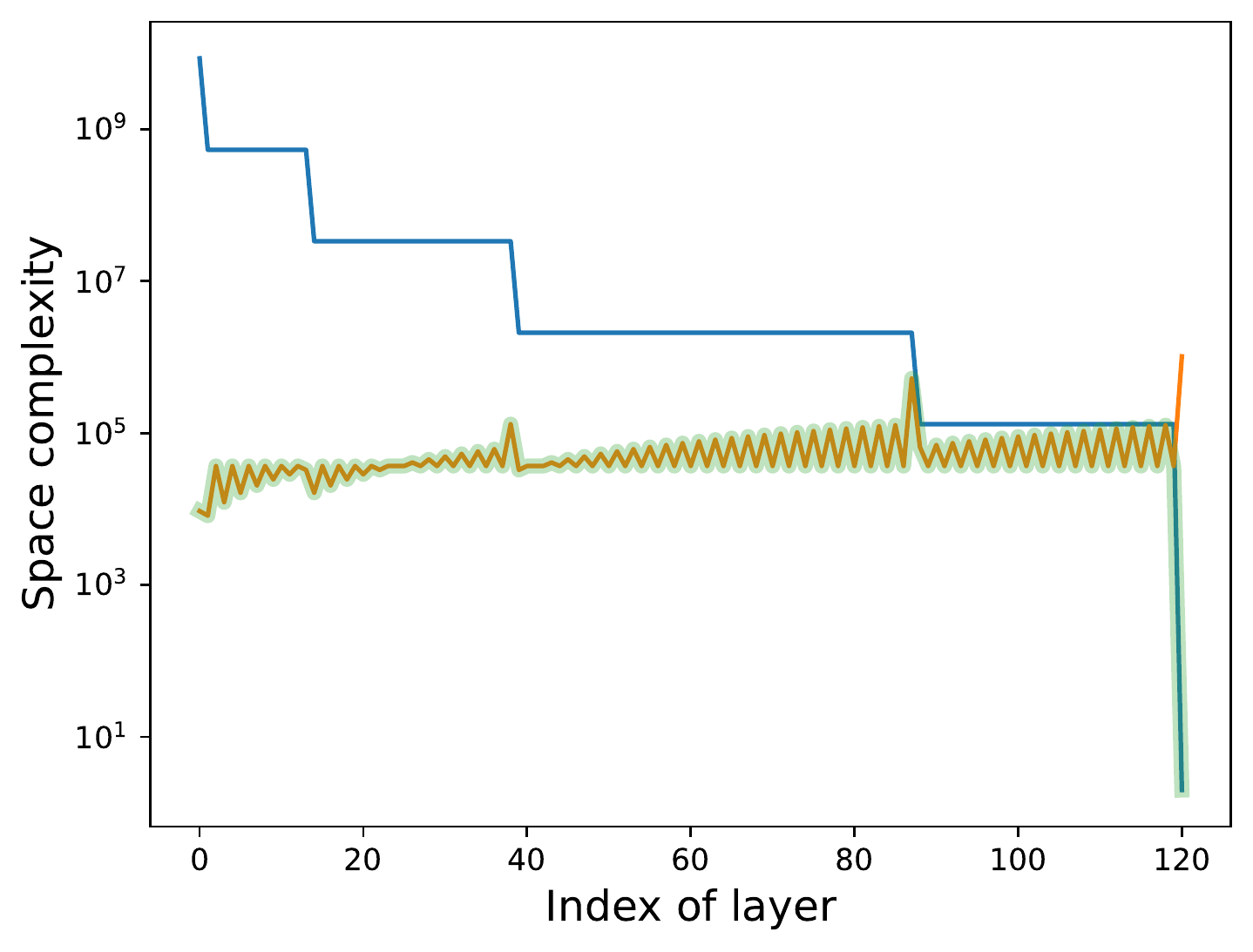}
\caption{Layerwise space complexity of computing the per-sample gradient norm in DenseNet121.}
\end{figure}

\begin{figure}[!htb]
    \centering
\includegraphics[width=0.3\linewidth]{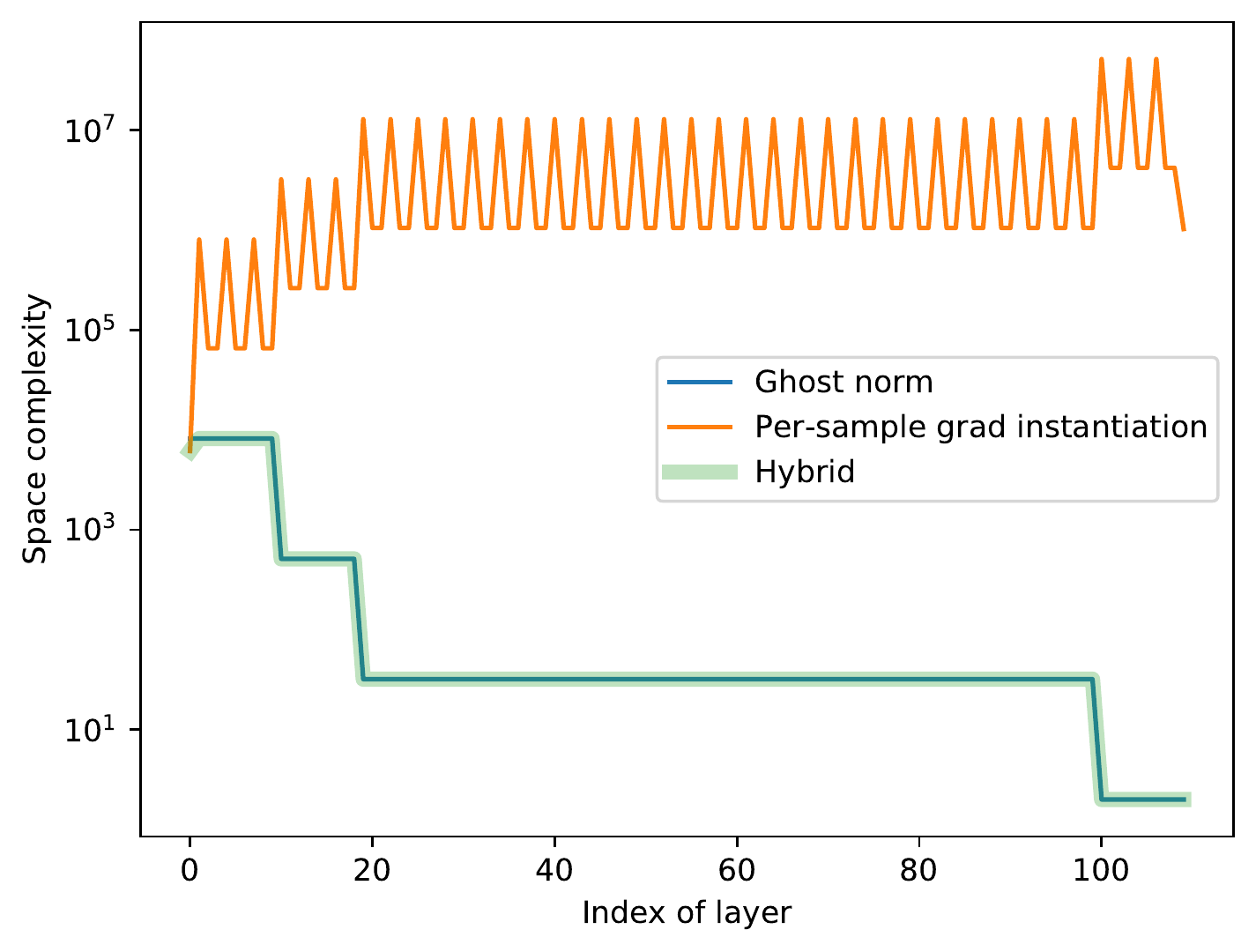}
\includegraphics[width=0.3\linewidth]{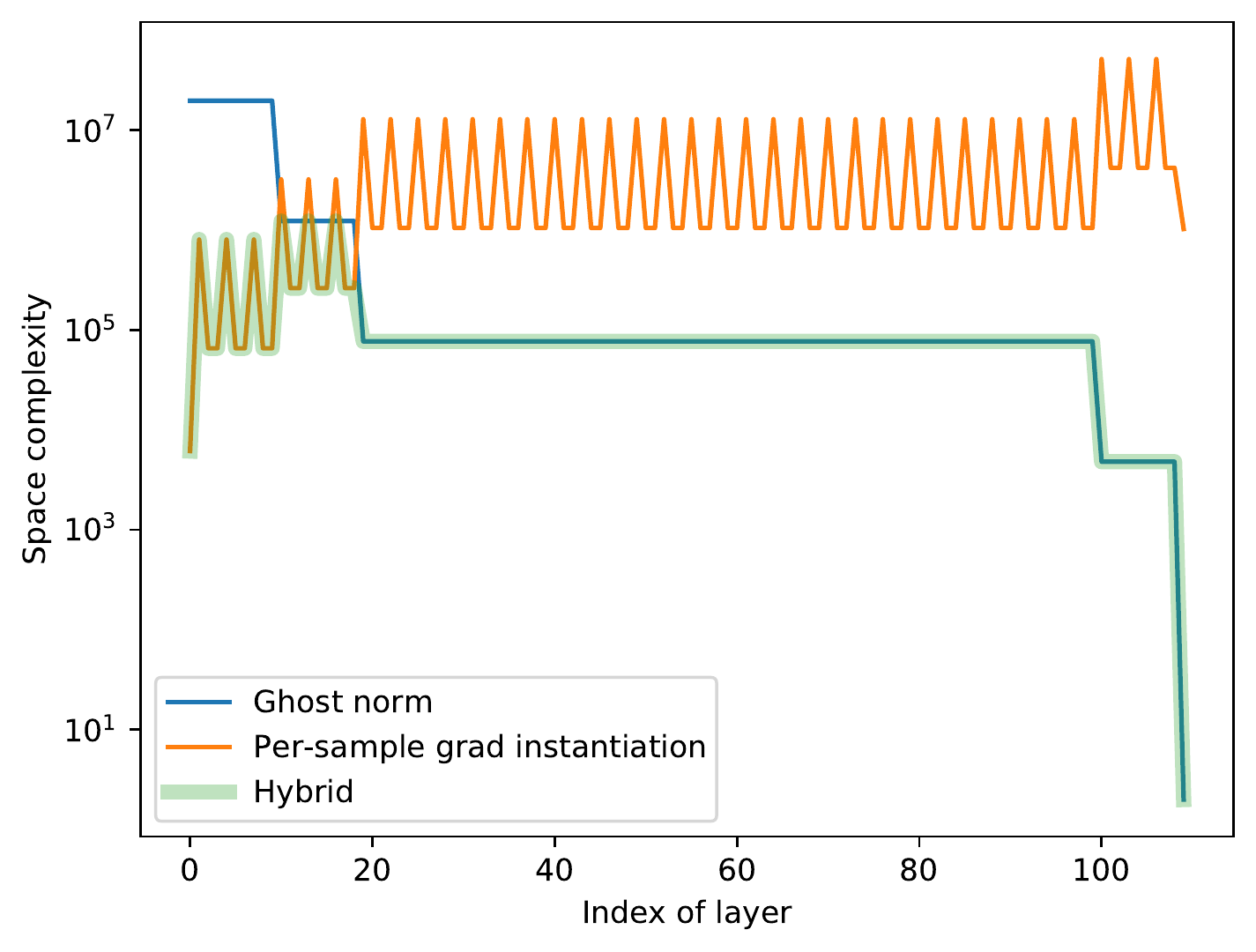}
\includegraphics[width=0.3\linewidth]{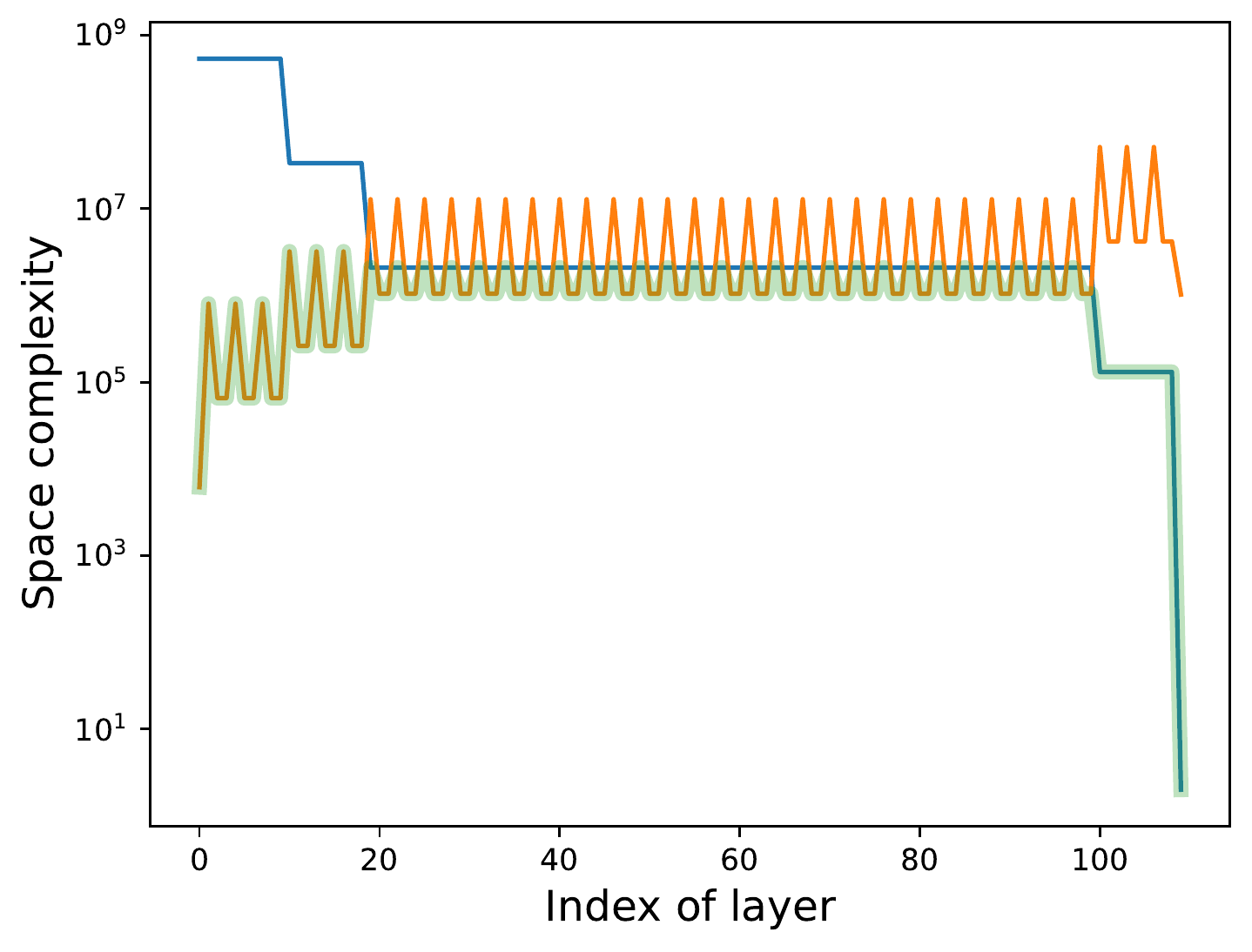}
\caption{Layerwise space complexity of computing the per-sample gradient norm in ConvNeXT.}
\end{figure}

\begin{figure}[!htb]
    \centering
\includegraphics[width=0.3\linewidth]{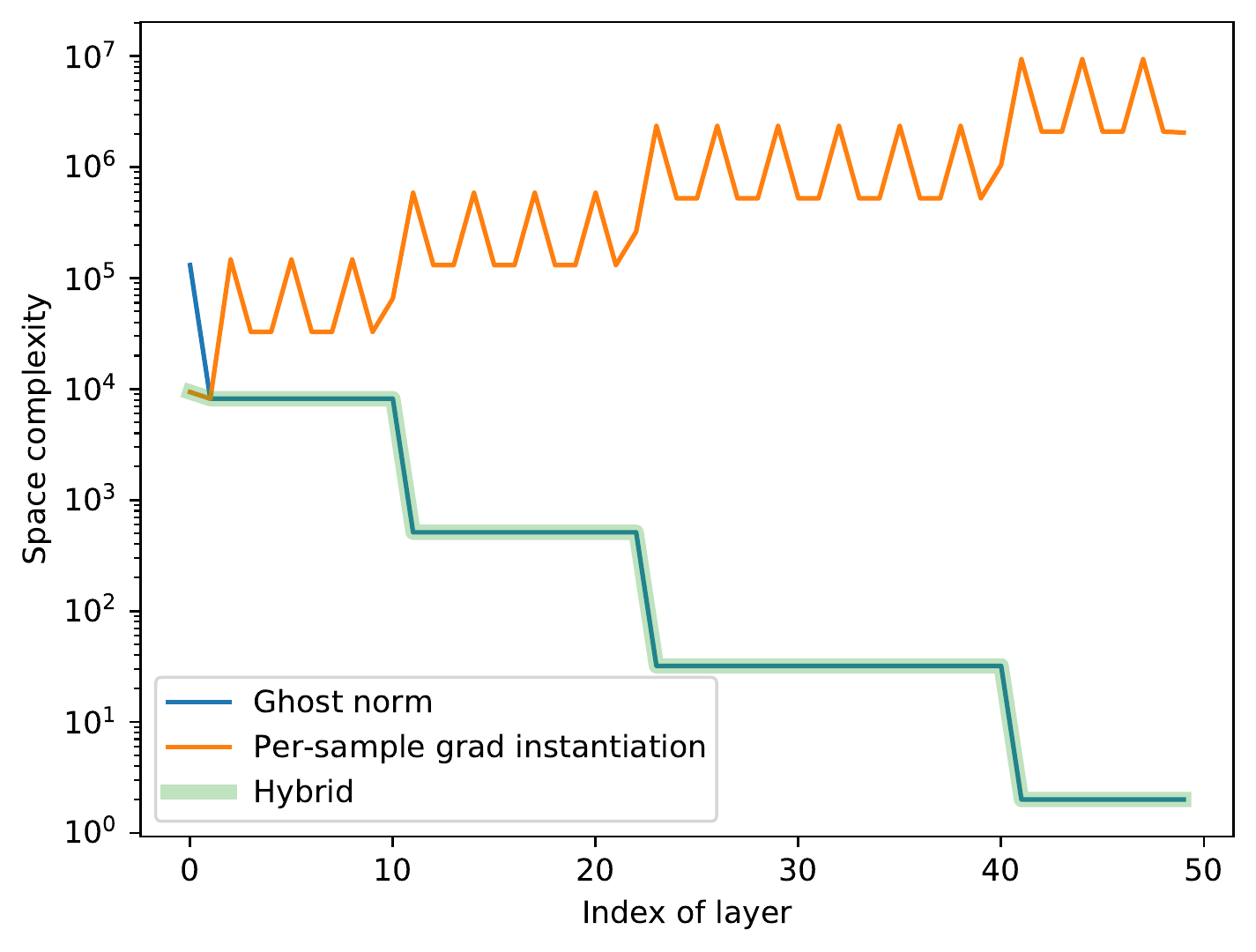}
\includegraphics[width=0.3\linewidth]{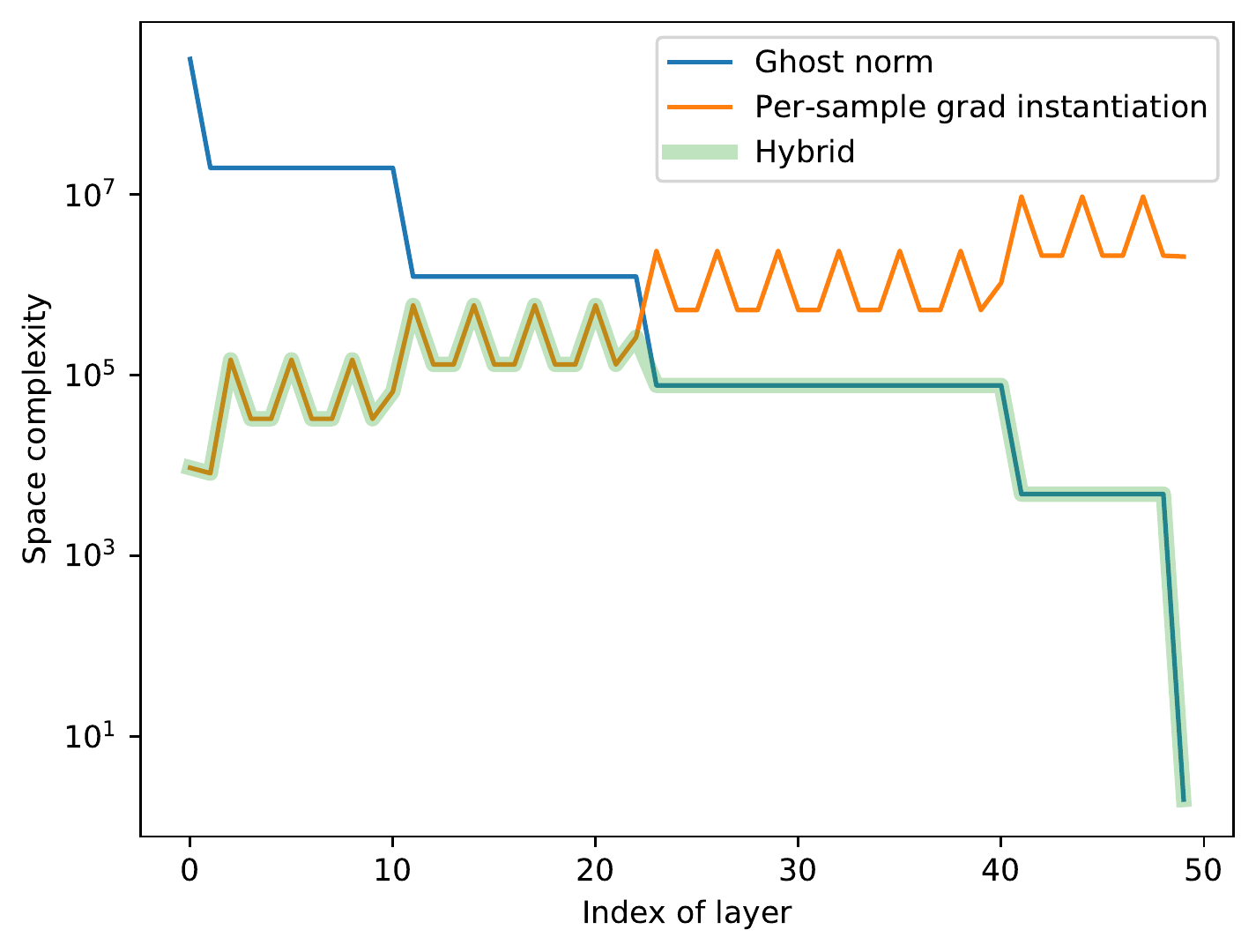}
\includegraphics[width=0.3\linewidth]{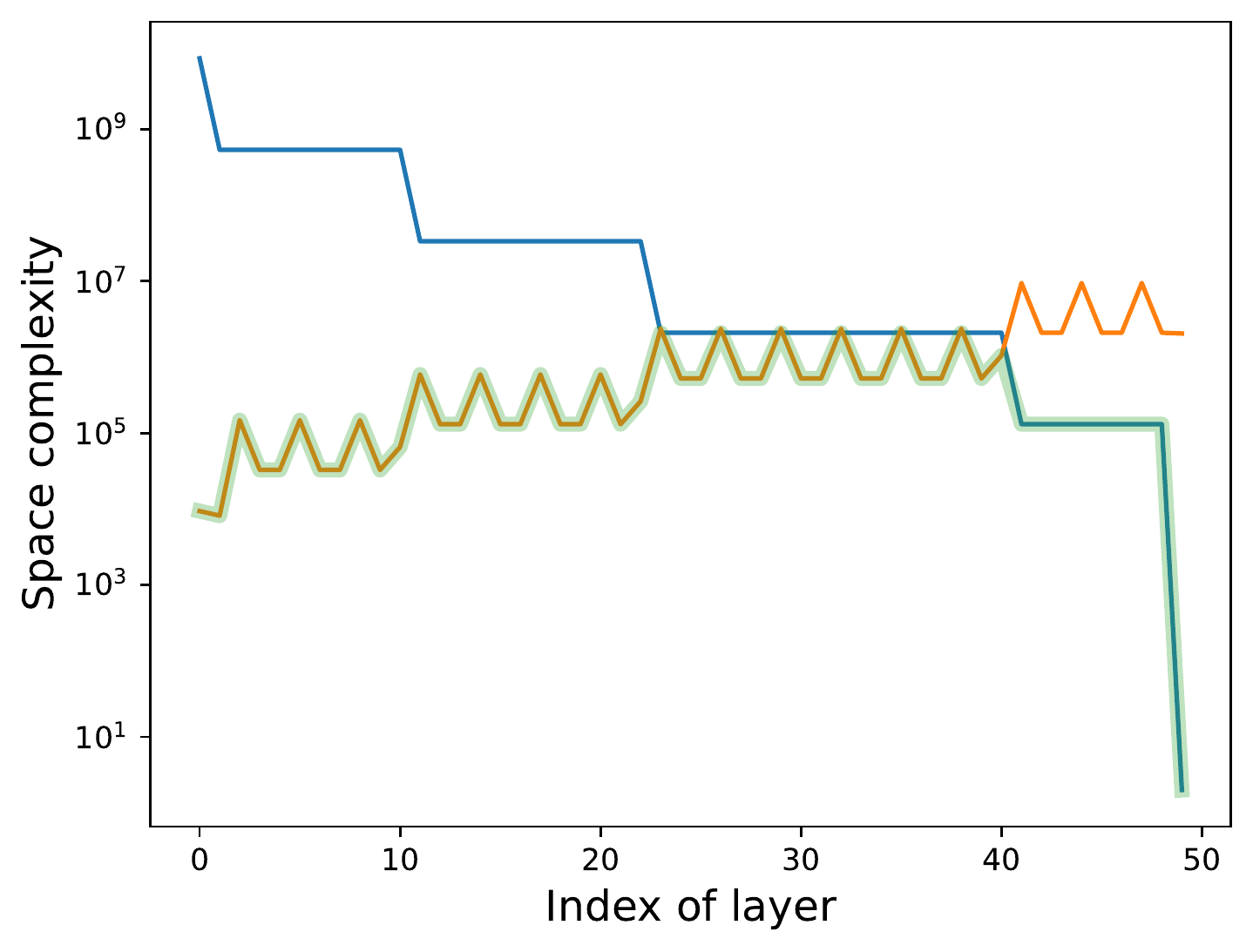}
\caption{Layerwise space complexity of computing the per-sample gradient norm in Wide ResNet50.}
\end{figure}

\begin{figure}[!htb]
    \centering
\includegraphics[width=0.3\linewidth]{figs/beit_large_patch16_224_layer_complexity.pdf}
\includegraphics[width=0.3\linewidth]{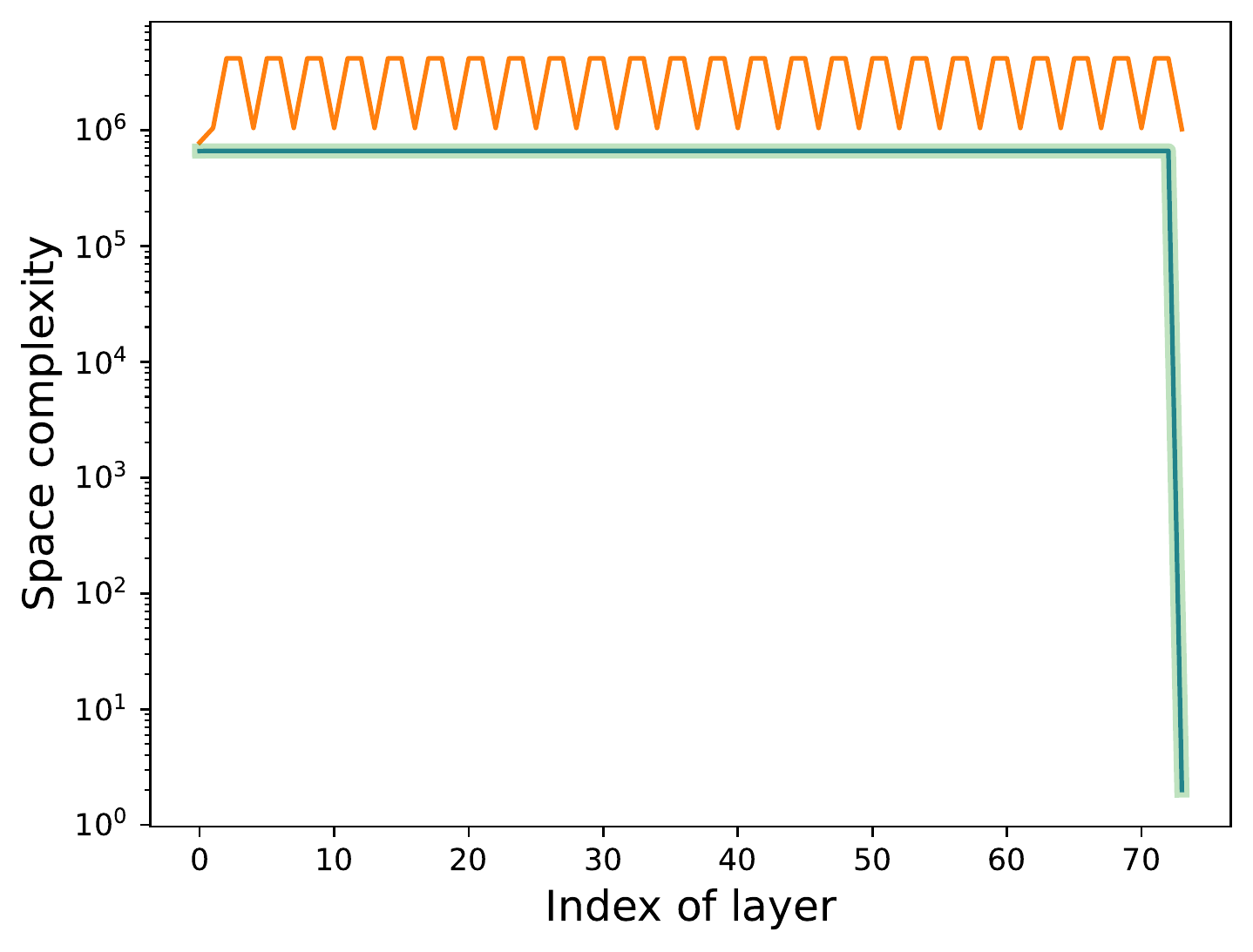}
\includegraphics[width=0.3\linewidth]{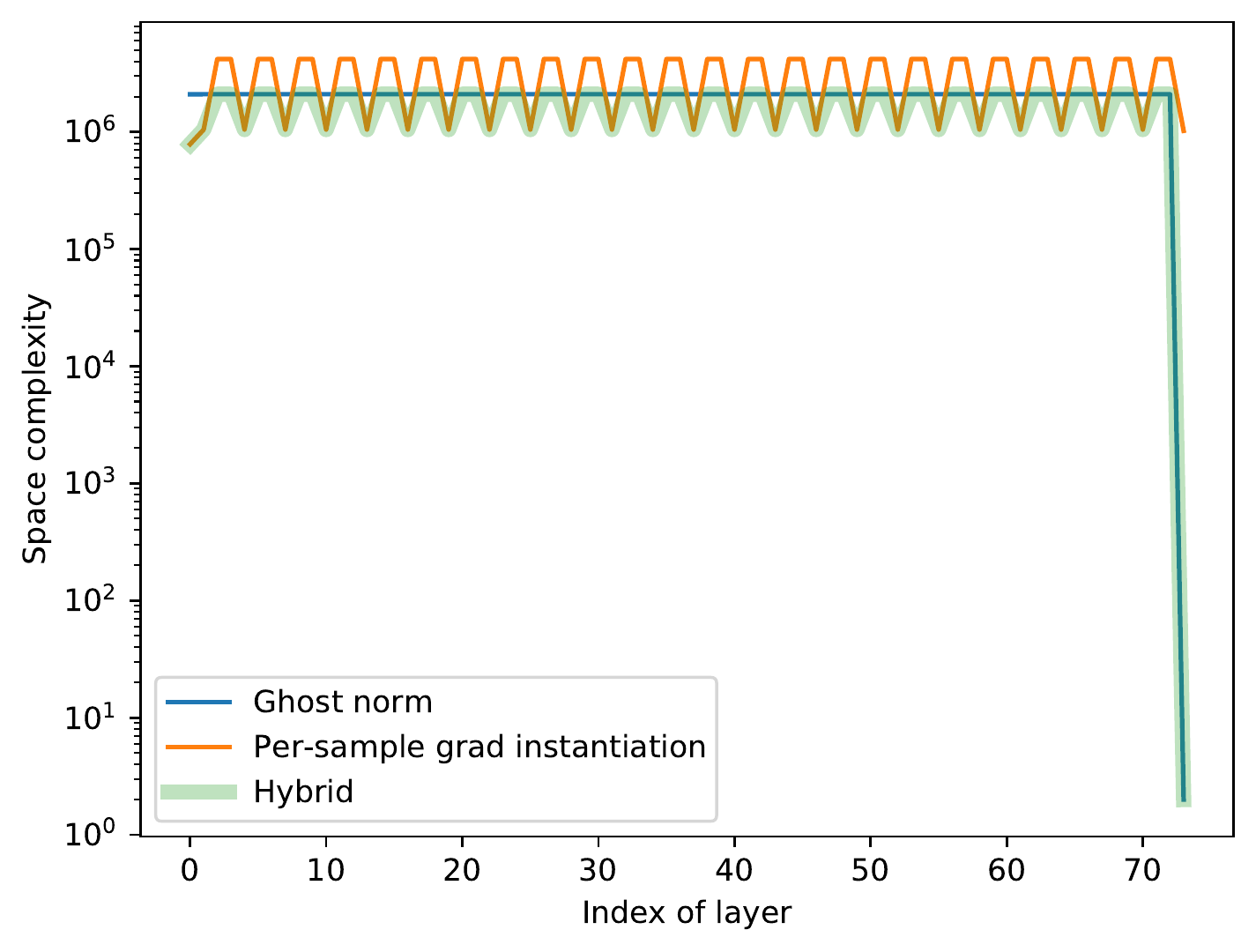}
\caption{Layerwise space complexity of computing the per-sample gradient norm in BEiT-large.}
\end{figure}

\end{document}